\documentclass[twocolumn,journal]{IEEEtran}
\usepackage[T1]{fontenc}
\usepackage[latin9]{inputenc}
\usepackage{array}
\usepackage{mathtools}
\usepackage{multirow}
\usepackage{amsmath}
\usepackage{amssymb}
\usepackage{stackrel}
\usepackage{graphicx}
\usepackage[unicode=true,
 bookmarks=true,bookmarksnumbered=true,bookmarksopen=true,bookmarksopenlevel=1,
 breaklinks=false,pdfborder={0 0 0},pdfborderstyle={},backref=false,colorlinks=false]
 {hyperref}
\hypersetup{pdftitle={Your Title},
 pdfauthor={Your Name},
 pdfpagelayout=OneColumn, pdfnewwindow=true, pdfstartview=XYZ, plainpages=false}
\usepackage{breakurl}

\makeatletter

\newcommand{\lyxmathsym}[1]{\ifmmode\begingroup\def\b@ld{bold}
  \text{\ifx\math@version\b@ld\bfseries\fi#1}\endgroup\else#1\fi}

\providecommand{\tabularnewline}{\\}

\usepackage[caption=false,font=footnotesize]{subfig}
\usepackage[linesnumbered, ruled]{algorithm2e}

\@ifundefined{showcaptionsetup}{}{%
 \PassOptionsToPackage{caption=false}{subfig}}
\usepackage{subfig}
\makeatother

\begin{document}

\title{\textbf{BS-Nets: An End-to-End Framework For Band Selection of Hyperspectral
Image}}

\author{Yaoming~Cai,~Xiaobo~Liu,~and Zhihua~Cai \thanks{This work was supported in part by the National Natural Science Foundation
of China under Grant 61773355 and Grant 61603355, in part by the Fundamental
Research Founds for National University, China University of Geosciences(Wuhan)
under Grant G1323541717, in part by the National Nature Science Foundation
of Hubei Province under Grant 2018CFB528, and in part by the Open
Research Project of Hubei Key Laboratory of Intelligent Geo-Information
Processing under Grant KLIGIP-2017B01. \emph{(Corresponding author:
X. Liu.)}}\thanks{Y. Cai and Z. Cai are with the School of Computer Science, China University
of Geosciences, Wuhan 430074, China (e-mail: \protect\href{http://caiyaom@cug.edu.cn}{caiyaom@cug.edu.cn};
\protect\href{http://zhcai@cug.edu.cn}{zhcai@cug.edu.cn}).}\thanks{X. Liu is with the School of Automation, China University of Geosciences,
Wuhan 430074, China (e-mail: \protect\href{http://xbliu@cug.edu.cn}{xbliu@cug.edu.cn}). }}
\maketitle
\begin{abstract}
Hyperspectral image (HSI) consists of hundreds of continuous narrow
bands with high spectral correlation, which would lead to the so-called
Hughes phenomenon and the high computational cost in processing. Band
selection has been proven effective in avoiding such problems by removing
the redundant bands. However, many of existing band selection methods
separately estimate the significance for every single band and cannot
fully consider the nonlinear and global interaction between spectral
bands. In this paper, by assuming that a complete HSI can be reconstructed
from its few informative bands, we propose a general band selection
framework, Band Selection Network (termed as BS-Net). The framework
consists of a band attention module (BAM), which aims to explicitly
model the nonlinear inter-dependencies between spectral bands, and
a reconstruction network (RecNet), which is used to restore the original
HSI cube from the learned informative bands, resulting in a flexible
architecture. The resulting framework is end-to-end trainable, making
it easier to train from scratch and to combine with existing networks.
We implement two BS-Nets respectively using fully connected networks
(BS-Net-FC) and convolutional neural networks (BS-Net-Conv), and compare
the results with many existing band selection approaches for three
real hyperspectral images, demonstrating that the proposed BS-Nets
can accurately select informative band subset with less redundancy
and achieve significantly better classification performance with an
acceptable time cost.
\end{abstract}


\section{Introduction}

\IEEEPARstart{H}{yperspectral} images (HSIs) acquired by remote sensors
consist of hundreds of narrow bands containing rich spectral and spatial
information, which provides an ability to accurately recognize the
region of interest. Over the past decade, HSIs have been widely applied
in various fields, ranging from agriculture \cite{HSI-Agriculture-Applications-2015-JSTARS-2015}
and land management \cite{HSI-spectral-unmixing-urban-environment-ReSensEnv-2017}
to medical imaging \cite{HSI-Medical-JBO-Lu-2014} and forensics \cite{HSI-forensic-traces-FCI-2012}. 

As the development of hyperspectral imaging techniques, the spectral
resolution has been improved greatly, resulting in difficulty of analyzing.
According to the characteristic of hyperspectral imaging, there is
a high correlation between adjacent spectral bands \cite{HSI_BS-EvoMultiObj-GongMG-TGRS-2016,Graph-Regu-Fast-RPCA-HSI_BS-SunW-TGRS-2018,HSI_BS-Laplacian-Regularized-LRSC-ZhaiH-TGRS-2018}.
The high-dimensional HSI data not only increases the time complexity
and space complexity but leads to the so-called Hughes phenomenon
or curse of dimensionality \cite{HSIC_SVM-MelganiF-TGRS-2004}. As
a result, redundancy reduction becomes particularly important for
HSI processing. 

Band Selection (BS) \cite{HSI_BS-Mutual-Information-Semi-Supervised-FengJ-TGRS-2015,HSI_BS-ISSC-SunWW-JSTARS-2015,HSI_BS-Dual-Clustering-YuanY-TGRS-2016},
also known as Feature Selection, is an effective redundancy reduction
scheme. Its basic idea is to select a significant band subset which
includes most information of the original band set. In contrast to
the feature extraction methods \cite{HSI-Lap-Collaborative-Discriminant-Jiangxw-RS-2019}
which reduces dimensionality based on the complex feature transformation,
BS keeps main physical property containing in HSIs \cite{HSI_BS-EvoMultiObj-GongMG-TGRS-2016},
which makes it easier to explain and apply in practice. 

BS methods basically can be classed as supervised and unsupervised
methods \cite{HSI_BS-multi-feature-info-maxi-clustering-ICIP-2017}
based on whether the prior knowledge is used. Owing to more robust
performance and higher application prospect, unsupervised BS method
has attracted a great deal of attention over the last few decades.
Unsupervised BS methods can be further divided into three categories:
searching-based, clustering-based, and ranking-based methods \cite{HSI_BS-Self-Representation-SunWW-JSTARS-2017}.
The searching-based BS methods treat band selection as a combinational
optimization problem and optimize it using a heuristic searching method,
such as multi-objective optimization based band selection (MOBS) \cite{HSI_BS-EvoMultiObj-GongMG-TGRS-2016,HSI_BS-Multi-objectiveOpt-ZhangMY-AppSoftCom-2018,HSI_BS-EMO-SparRepres-HuP-GRSL-2018}.
However, heuristic searching methods are generally time-consuming.
The clustering-based BS methods assume spectral bands are clusterable
\cite{HSI_BS-Optimal-Clustering-Framework-WangQ-TGRS-2018,HSI_BS-multi-feature-info-maxi-clustering-ICIP-2017}.
Since the similarity between spectral bands is made full consideration,
clustering-based methods have achieved great success in recent years,
for example, subspace clustering (ISSC) \cite{HSI_BS-ISSC-SunWW-JSTARS-2015,HSI_BS-Laplacian-Regularized-LRSC-ZhaiH-TGRS-2018}
and sparse non-negative matrix factorization clustering (SNMF) \cite{HSI_BS-SNMF-JiM-FITEE}.
The ranking-based BS methods endeavor to assign a rank or weight for
each spectral band by estimating the band significance, e.g., maximum-variance
principal component analysis (MVPCA) \cite{HSI_BS-MVPCA-ChangCI-TGRS-1999},
sparse representation (SpaBS)\cite{HSI_BS-SpaBS-TargetDectection-SunK-GRSL-2015,HSI_BS-Multitask-Sparsity-Pursuit-YuanY-TGRS-2015},
and geometry-based band selection (OPBS) \cite{HSI-Band_Selection-OPBS-Geometry-Based-ZhangW-TGRS-2018},
etc.

Nevertheless, many existing BS methods are basing on the linear transformation
of spectral bands, resulting in the lack of consideration of the inherent
nonlinear relationship between spectral bands. Furthermore, most of
the BS methods commonly view every single spectral band as a separate
image or point and evaluate its significance independently. For example,
clustering-based BS methods are essentially clustering spectral images
with single channel \cite{HSI_BS-ISSC-SunWW-JSTARS-2015,HSI_BS-Laplacian-Regularized-LRSC-ZhaiH-TGRS-2018,HSI_BS-SpaBS-TargetDectection-SunK-GRSL-2015}.
Therefore, these methods can not take the global spectral interrelationship
into account and are difficult to combine with various post-processing,
such as classification \cite{HSIC-Gaussian-Process-JiangXW-GRSL-2017}. 

In this paper, we treat HSI band selection as a spectral reconstruction
task assuming that spectral bands can be sparsely reconstructed using
a few informative bands. Unlike the existing BS methods, we aim to
take full consideration of the globally nonlinear spectral-spatial
relationship and allow to select significant bands from the complete
spectral band set, even the 3-D HSI cubes. To this end, we design
a band selection network (BS-Net) based on using deep neural networks
(DNNs) \cite{Deep-learning-LeCun-Nature-2015,CNN-overview-GuJX-PR-2018}
to explicitly model the nonlinear interdependencies between spectral
bands. Although DNNs have been widely used for HSI classification
\cite{CNN-HSI-FeaExtr-Clas-Chenyushi-TGRS-2016,Bi-Conv-LSTM-HSI-Liu-RS-2017,ResidualNN-HSI-Zhong-TGRS-2018}
and feature extraction \cite{BLDE-CNN-HSI-Zhao-TGRS-2016,CNN-HSI-FeaExtr-Clas-Chenyushi-TGRS-2016,Deep-FullyCNN-HSI-Jiao-TGRS-2017},
DNN-based band selection has not attracted much attention yet. 

The main contributions of this paper are as follows:

\begin{enumerate} 
\item We proposed an end-to-end band selection framework based on using deep neural networks to learn the nonlinear interdependencies between spectral bands. To the best of our knowledge, this is among the few deep learning based band selection methods.
\item We implemented two different BS-Nets according to the different application scenarios, i.e., spectral-based BS-Net-FC and spectral-spatial-based BS-Net-Conv.
\item We extensively evaluated the proposed BS-Nets framework from aspects of classification performance and quantitative evaluation, showing that BS-Nets can achieve state-of-the-art results.
\end{enumerate}

The rest of the paper is structured as follows. We first define the
notations and review the basic concepts of deep learning in Section
\ref{sec:Preliminary}. Second, we introduce the proposed BS-Net architecture
and its implementation in Section \ref{sec:BS-Nets}. Next, in Section
\ref{sec:Results}, we explain the experiments that performed to investigate
the performance of the proposed methods, compared with existing BS
methods, and discuss their results. Finally, we conclude with a summary
and final remarks in Section \ref{sec:Conclusions}.

\section{Preliminary \label{sec:Preliminary}}

\subsection{Definition and Notations}

We denote a 3-D HSI cube consisting of $b$ spectral bands and $N\times M$
pixels as $\boldsymbol{I}\in\mathbb{R}^{N\times M\times b}$. For
convenience, we regard \textbf{$\boldsymbol{I}$} as a set $\boldsymbol{B}=\left\{ \boldsymbol{B}_{i}\right\} _{i=1}^{b}$
which contains $b$ band images, where $\boldsymbol{B}_{i}$ indicates
$i$-th band image. HSI band selection can thus be formally defined
as a function $\psi:\varOmega=\psi\left(\boldsymbol{B}\right)$ that
takes all bands as input and produces a band subset with as less as
possible reduction of redundant information and satisfied $\varOmega\subseteq\boldsymbol{B}$,
$\left|\varOmega\right|=k<b$. 

In the following, unless as otherwise specified herein, we uniformly
use tensors to represent the inputs, outputs, and intermediate outputs
involved in the neural networks. For example, the input of a convolutional
layer is denoted as a 4-D tensor $\boldsymbol{x}\in\mathbb{R}^{n\times m\times c}$,
where $n\times m$ is the spatial size of the input feature maps,
and $c$ is the number of the channels. 

\begin{figure}[tbh]
\begin{centering}
\includegraphics[width=0.8\columnwidth]{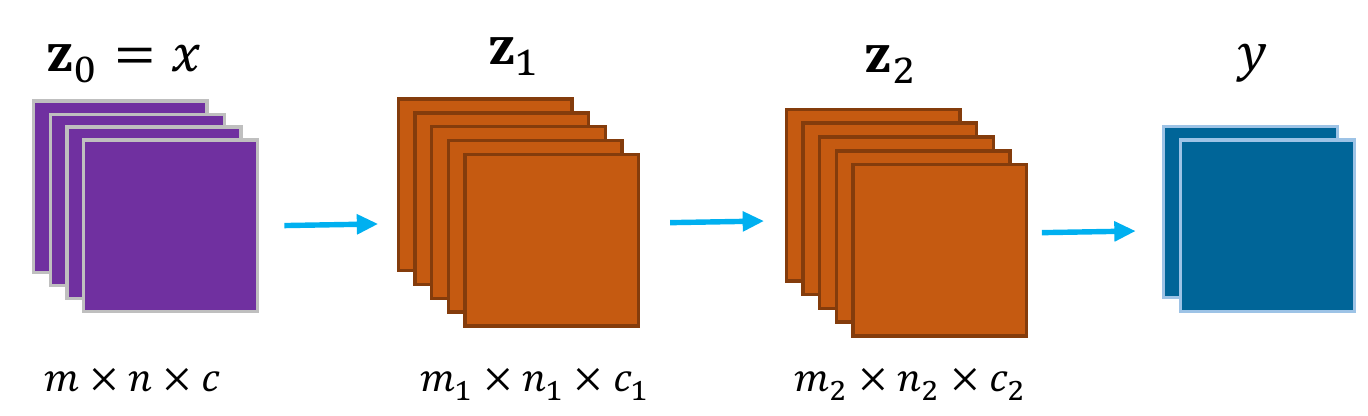}
\par\end{centering}
\caption{An example of classical CNN with input $\boldsymbol{x}$, output $y$,
and two convolutional layers $\boldsymbol{z}_{1}$ and $\boldsymbol{z}_{2}$
. \label{fig:An-example-of-cnn}}
\end{figure}

\subsection{Convolutional Neural Networks}

Deep learning has achieved great success in numerous applications
ranging from image recognition to natural language processing \cite{Deep-learning-LeCun-Nature-2015,DL-HSI-overview-ZhangLP-GRSM-2016,HE-ELM-CYM-PRL-2018}.
The collection of deep learning methods includes Convolutional Neural
Networks (CNN) \cite{CNN-overview-GuJX-PR-2018,Mixed-dense-CNN-PeltMD-PNAS-2018},
Generative Adversarial Networks (GAN) \cite{GAN-Goodfellow-NIPS-2014,GAN-Overview-Creswell-IEEEIPM-2018},
and Recurrent Neural Networks (RNN) \cite{HSIC-RNN-Mou-TGRS-2017},
to name a few. In this section, we take CNN as an example to introduce
the basic idea of DNNs, since it is the most popular deep learning
method in HSI processing. 

Convolutional neural networks (CNN) are inspired by the natural visual
perception mechanism of the living creatures \cite{CNN-overview-GuJX-PR-2018}.
The classical CNN consists of multiple layers of convolutional operations
with nonlinear activations, sometimes, followed by a regression layer.
A schematic representation of the basic CNN architecture is shown
in Fig. \ref{fig:An-example-of-cnn}. We define CNN as a function
that takes a tensor $\boldsymbol{x}\in\mathbb{R}^{m\times n\times c}$
as input and produces a certain output $\boldsymbol{y}$. The function
can be written as $\boldsymbol{y}=\boldsymbol{f}\left(\boldsymbol{x};\varTheta\right)$,
where $\varTheta$ is the trainable parameters consisting  of weights
and biases involved in CNN. 

The training of CNN includes two stages. The first stage is values
feedforward wherein each layer yields a dozen of feature maps $\boldsymbol{h}_{i}\in\mathbb{R}^{m_{i}\times n_{i}\times c_{i}}$.
Let $\hbar:\mathbb{R}^{m\times n\times c}\rightarrow\mathbb{R}^{m_{i}\times n_{i}\times c_{i}}$
be the convolutional operation and $\boldsymbol{\sigma}$ be a element-wise
nonlinear function such as Sigmoid and Rectified Linear Unit (ReLU).
The convolutional layer can be represented as 

\begin{equation}
\boldsymbol{h}_{i}=\boldsymbol{\sigma}\left(\hbar\left(\boldsymbol{x};\mathbf{W}\right)+\boldsymbol{b}\right)\label{eq:conv-maps}
\end{equation}
Here $\mathbf{W}$ and $\boldsymbol{b}$ indicate weights (aka convolutional
kernels or convolutional filters) and bias, respectively. 

The second stage is called error backpropagation which updates parameters
using the gradient descent method. The ultimate goal of CNN is to
find an appropriate group of filters to minimize the cost function,
e.g.,  Mean Square Error (MSE) function. The cost function can be
denoted as

\begin{equation}
\mathcal{J}\left(\varTheta\right)=Cost\left(\boldsymbol{y},\boldsymbol{f}\left(\boldsymbol{x};\varTheta\right)\right)
\end{equation}
The parameters updating is given by 

\begin{equation}
\varTheta\eqqcolon\varTheta-\eta\frac{\partial\mathcal{J}}{\partial\varTheta}
\end{equation}
Where $\eta$ is learning rate (or step size), and the partial derivatives
of the cost function w.r.t. the trainable parameters can be calculated
using the chain rule. 

\subsection{Attention Mechanism}

Attention is, to some extent, motivated by how human pay visual attention
to different regions of an image or correlate words in one sentence.
In \cite{Squeeze-and-Excitation-Hu-CVPR-2018}, attention was defined
as a method to bias the allocation of available processing resources
towards the most informative components of an input signal. Mathematically,
attention in deep learning can be broadly interpreted as a function
of importance weights, $\delta$. 

\begin{equation}
\boldsymbol{\omega}=\delta\left(\boldsymbol{x};\varTheta\right)\label{eq:attention-definition}
\end{equation}
Where $\boldsymbol{\omega}$ can be a matrix or vector that indicates
the importance of a certain input. The implementation of attention
is generally consisting of a gating function (e.g.,  Sigmoid or Softmax)
and combined with multiple layers of nonlinear feature transformation. 

Attention mechanism is widely applied across a range of tasks, including
image processing \cite{Squeeze-and-Excitation-Hu-CVPR-2018,Spatial-Transformer-Networks-attention-NIPS-2015,Residual-Attention-Wang-CVPR-2017}
and natural language processing \cite{ICML-Xu_2015-img-caption-attention-Xu-2015,Attention-Machine-Translation-Luong-EMNLP-2015,Attention-is-All-you-Need-NIPS-2017}.
In this paper, we focus mainly on the attention in visual systems.
According to the different concerns of attention methods, visual attention
basically can be divided into three categories. The first category
is spatial attention, which is used to learn the pixel-wise relationship
over the images, such as Spatial Transformer Networks \cite{Spatial-Transformer-Networks-attention-NIPS-2015}.
Similarly, the second category is focusing on learning the channel-wise
relationship, which is also called channel attention, e.g.,  Squeeze-and-Excitation
Networks \cite{Squeeze-and-Excitation-Hu-CVPR-2018}. The third category
is the combination of both channel attention and spatial attention.
Thus it is mixed attention, such as Convolutional Block Attention
Module \cite{SBAM-channel-spatial-attention-ECCV-2018}. 

For an HSI band selection task, our goal is to pay more attention
to those informative bands and moreover to avoid the influence of
the trivial bands. Therefore, our proposed BS-Nets are essentially
a variant of the channel attention based method and we refer to such
an attention used in HSI as Band Attention (BA).

\begin{figure}[t]
\begin{centering}
\includegraphics[width=1\columnwidth]{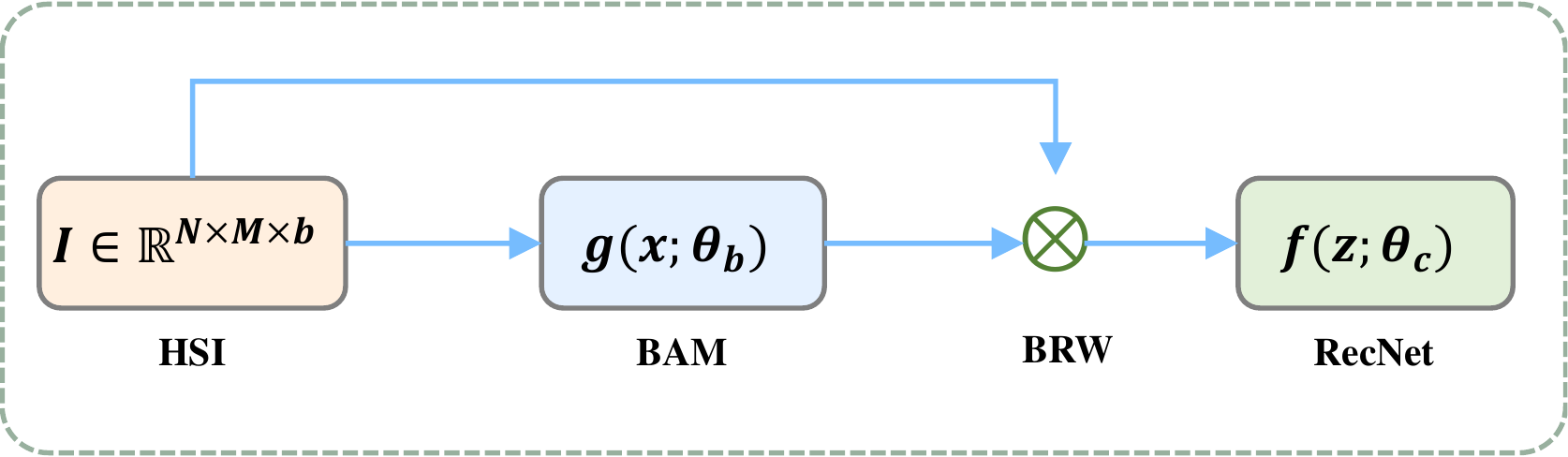}
\par\end{centering}
\caption{Overview of Band Selection Networks: A given HSI data is first passed
onto a Band Attention Module (BAM) to explicitly model the nonlinear
interdependencies between spectral bands. Then, the input HSI is re-weighted
band-wisely by a Band Re-weighting (BRW) operation. Finally, a Reconstruction
Network (RecNet) is conducted to restore the original spectral bands
from the re-weighted bands. \label{fig:Schematic-representation-of-BS-Net-overall}}
\end{figure}

\section{BS-Nets \label{sec:BS-Nets}}

\begin{figure*}[tbh]
\begin{centering}
\subfloat[BS-Net-FC.\label{fig:Schematic-representation-of-BS-Net-MLP}]{\begin{centering}
\includegraphics[width=1\columnwidth]{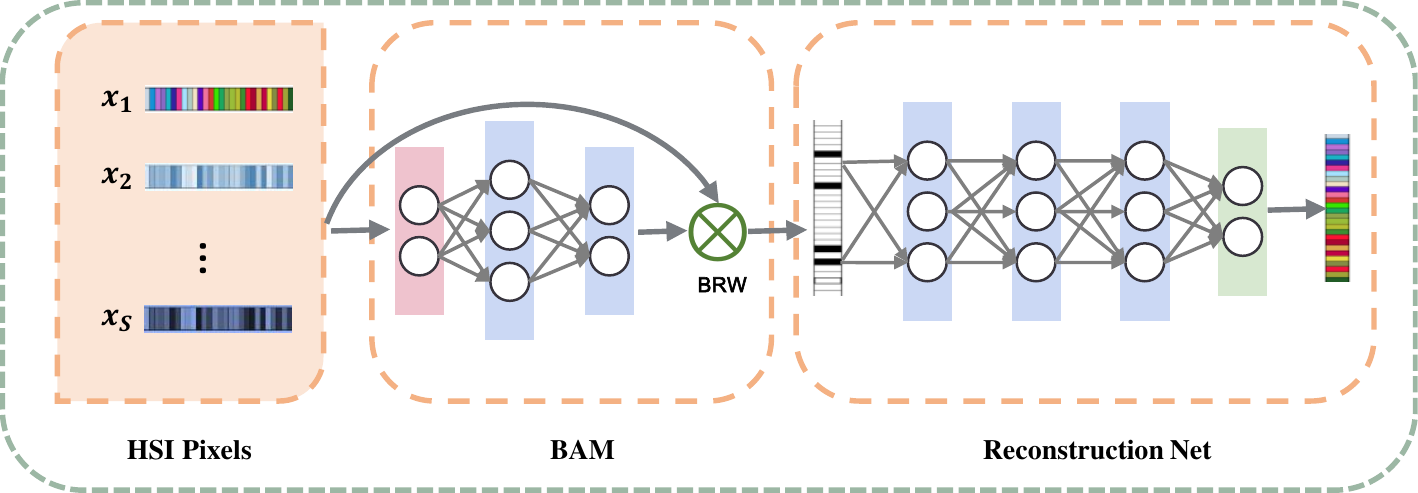}
\par\end{centering}
}\subfloat[BS-Net-Conv.\label{fig:Schematic-representation-of-BS-Net}]{\begin{centering}
\includegraphics[width=1\columnwidth]{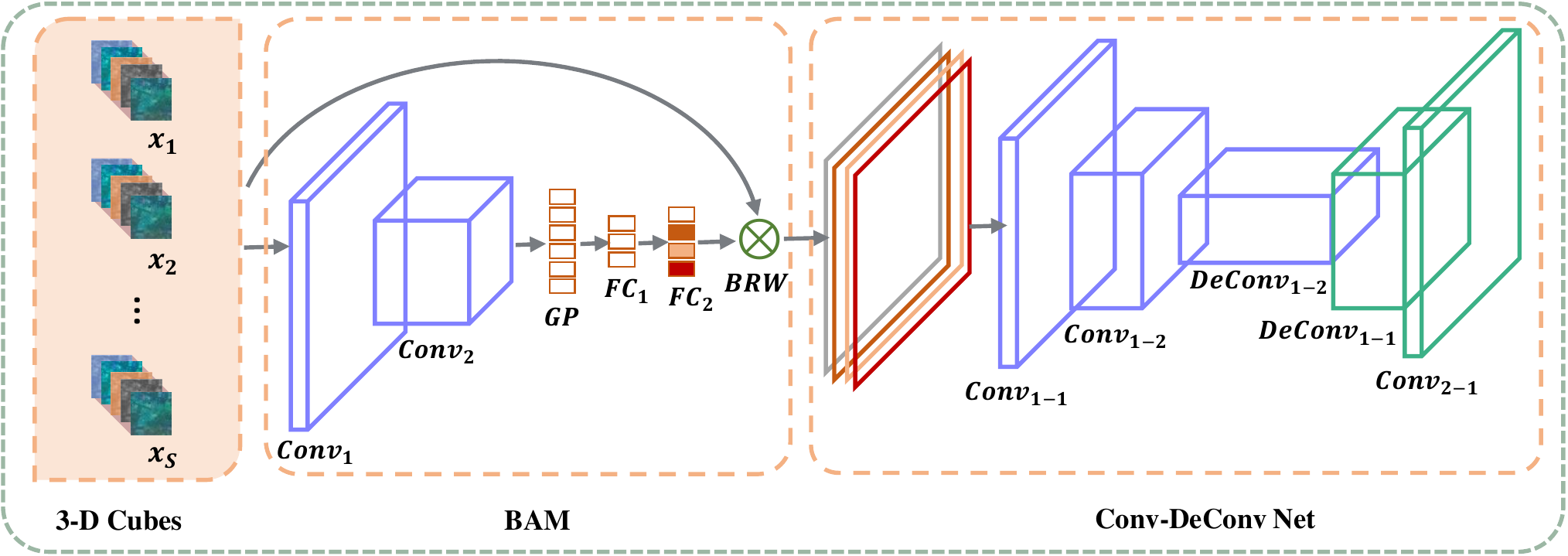}
\par\end{centering}
}
\par\end{centering}
\caption{Implementation details of BS-Nets based on different networks. (a)
BS-Net based on fully connected networks with spectral inputs. (b)
BS-Net based on convolutional neural networks with spectral-spatial
inputs. \label{fig:Implementation-examples-of-BS-Nets}}
\end{figure*}

In this section, we first introduce the main components included in
the BS-Nets general architecture. Then, we give two versions of implementations
of the BS-Nets based on fully connected networks and convolutional
neural networks, respectively. Finally, we show a discussion on the
BS-Nets. 

\subsection{Architecture of BS-Nets}

The key to the BS-Nets is to convert the band selection as a sparse
band reconstruction task, i.e., recover the complete spectral information
using a few informative bands. For a given spectral band, if it is
informative then it will be essential for a spectral reconstruction.
To this end, we design a deep neural network based on the attention
mechanism. In Fig. \ref{fig:Schematic-representation-of-BS-Net-overall},
we show the overall architecture of the proposed framework, which
consists of three components: band attention module (BAM), band re-weighting
(BRW), and reconstruction network (RecNet). The detailed introduction
is given as follows. 

The BAM is a branch network which we use to learn the band weights.
As shown in Fig. \ref{fig:Schematic-representation-of-BS-Net-overall},
BAM directly takes HSI as input and aims to fully extract the interdependencies
between spectral bands. We express BAM as a function $g$ that takes
a certain HSI cube $\boldsymbol{x}$ as input and produces a non-negative
band weights tensor, \textbf{$\boldsymbol{w}\in\mathbb{R}^{1\times1\times b}$}. 

\begin{equation}
\boldsymbol{w}=g\left(\boldsymbol{x};\boldsymbol{\varTheta}_{b}\right)\label{eq:BAM-1}
\end{equation}
Here $\boldsymbol{\varTheta}_{b}$ denotes the trainable parameters
involved in the BAM. To guarantee the non-negativity of the learned
weights, Sigmoid function is adopted as the activation of the output
layer in BAM, which is written as:

\begin{equation}
\phi\left(\boldsymbol{w}\right)=\frac{1}{1+e^{-\boldsymbol{w}}}\label{eq:sig}
\end{equation}

To create an interaction between the original inputs and their weights,
a band-wise multiplication operation is conducted. We refer to this
operation as BRW. It can be explicitly represented as follows. 

\begin{equation}
\mathbf{z}=\boldsymbol{x}\otimes\boldsymbol{w}\label{eq:BReW-1}
\end{equation}
Where $\otimes$ indicates the band-wise production between $\boldsymbol{x}$
and $\boldsymbol{w}$, and $\mathbf{z}$ is the re-weighted counterpart
of the input $\boldsymbol{x}$. 

In the next step, we employ the RecNet to recover the original spectral
band from the re-weighted counterpart. Similarly, we define the RecNet
as a function $f$ that takes a re-weighted tensor $\mathbf{z}$ as
input and outputs its prediction. 

\begin{equation}
\hat{\boldsymbol{x}}=f\left(\mathbf{z};\boldsymbol{\varTheta}_{c}\right)
\end{equation}
Where $\hat{\boldsymbol{x}}$ is the prediction output for the original
input $\boldsymbol{x}$, and $\boldsymbol{\varTheta}_{c}$ denotes
the trainable parameters involved in RecNet. 

In order to measure the reconstruction performance, we use the Mean-Square
Error (MSE) as the cost function, denoted as $\mathcal{L}$. We define
it as follows:

\begin{equation}
\mathcal{L}=\frac{1}{2S}\stackrel[i=1]{S}{\sum}\left\Vert \boldsymbol{x}_{i}-\hat{\boldsymbol{x}}_{i}\right\Vert _{2}^{2}\label{eq:loss-mse}
\end{equation}
Here $S$ is the number of training samples. Moreover, we desire to
keep the band weights as sparse as possible such that we can interpret
them more easily. For this purpose, we impose an $L_{1}$ norm constraint
on the band weights. The resulting loss function is given as follows: 

\begin{equation}
\mathcal{L}\left(\boldsymbol{\varTheta}_{b},\boldsymbol{\varTheta}_{c}\right)=\frac{1}{2S}\stackrel[i=1]{S}{\sum}\left\Vert \boldsymbol{x}_{i}-\hat{\boldsymbol{x}}_{i}\right\Vert _{2}^{2}+\lambda\stackrel[i=1]{S}{\sum}\left\Vert \boldsymbol{w}_{i}\right\Vert _{1}\label{eq:loss-final}
\end{equation}
Where $\lambda$ is a regularization coefficient which balances the
minimization between the reconstruction error and regularization term.
Eq. \eqref{eq:loss-final} can be optimized by using a gradient descent
method, such as Stochastic Gradient Descent (SGD) and Adaptive Moment
Estimation (Adam). 

According to the learned sparse band weights, we can determine the
informative bands by averaging the band weights for all the training
samples. The average weight of the $j$-th band is computed as: 

\begin{equation}
\overline{w}_{j}=\frac{1}{S}\stackrel[i=1]{S}{\sum}\boldsymbol{w}_{ij}\label{eq:mean-weights-1}
\end{equation}
Those bands which have larger average weights are considered to be
significant since they make more contributions to the reconstruction.
In practice, the top $k$ bands are selected as the significant band
subset. The pseudocode of BS-Nets is given in Algorithm 1. 

\begin{algorithm}[tbh]  
\caption{Pseudocode of BS-Nets} 
\label{alg:pseudocode}
\KwIn{HSI cube: $\boldsymbol{I}\in\mathbb{R}^{N\times M\times b}$; Band subset size: $k$; and BS-Nets hyper-parameters.}
\KwOut{Informative band subset.}
Preprocess HSI and generate training samples\;
Random initialize $\boldsymbol{\varTheta}_{b}$ and $\boldsymbol{\varTheta}_{c}$ according to the given network configure\;
\While{Model is convergent or maximum iteration is met}
{
	Sample a batch of training samples $\boldsymbol{x}$\;
	Calculate bands weights: $\boldsymbol{w}=g\left(\boldsymbol{x};\boldsymbol{\varTheta}_{b}\right)\label{eq:BAM-1}$\;
	Re-weight spectral bands: $\mathbf{z}=\boldsymbol{x}\otimes\boldsymbol{w}\label{eq:BReW-1}$\;
	Reconstruct spectral bands: $\hat{\boldsymbol{x}}=f\left(\mathbf{z};\boldsymbol{\varTheta}_{c}\right)$\;
	Update $\boldsymbol{\varTheta}_{b}$ and $\boldsymbol{\varTheta}_{c}$ by minimizing Eq.\eqref{eq:loss-final} using Adam algorithm\;
}
Calculate average band weights according to Eq. \eqref{eq:mean-weights-1}\;
Select  top $k$ bands\;
\end{algorithm}

\subsection{BS-Net Based on Fully Connected Networks (BS-Net-FC)}

In Fig. \ref{fig:Implementation-examples-of-BS-Nets} (a), we show
the first implementation of BS-Net based on fully modeling the nonlinear
relationship between the spectral information. In this case, both
of BAM and RecNet are implemented with fully connected networks, and
thus we refer to this BS-Net as BS-Net-FC. 

As illustrated in Fig. \ref{fig:Implementation-examples-of-BS-Nets}
(a), the BAM is designed as a bottleneck structure with multiple fully
connected layers, with ReLu activations for all the middle hidden
layers. According to the\textbf{ }information bottleneck theory\textbf{
}\cite{Deep-Learning-information-bottleneck-Tishby-ITM-2015}, bottleneck
structure would be favorable for the extraction of information, although
different structures are allowed in BS-Nets. 

In BS-Net-FC, we use spectral vectors (pixels) as the training samples.
For convenience, we denote the training set comprising $S$ samples
as a 4-D tensor $\mathbf{X}\in\mathbb{R}^{S\times1\times1\times b}$,
where $S=M\times N$. By rewriting the band weights in the tensor
form, represented as $\mathbf{W}\in\mathbb{R}^{S\times1\times1\times b}$,
the BRW is actually an element-wise production operation that can
be written as $\mathbf{Z}=\mathbf{X}\otimes\mathbf{W}$, where $\mathbf{Z}$
is the re-weighted spectral inputs. In RecNet, we use a simple multi-layer
perceptron model with the same number of hidden neurons with ReLu
activations to reconstruct spectral information. 

\subsection{BS-Net Based on Convolutional Networks (BS-Net-Conv)}

During the training in BS-Net-FC, only the spectral information is
taken into account. The lack of consideration for the spatial information
would result in low-efficiency use of the spectral-spatial information
containing in HSI. To enhance the BS-Net-FC, we implement the second
BS-Net by using convolutional networks, which is termed as BS-Net-Conv.
The schematic of the implementation is given in Fig. \ref{fig:Implementation-examples-of-BS-Nets}
(b). 

In the BAM, we first employ several 2-D convolutional layers to extract
spectral and spatial information simultaneously. Then, a global pooling
(GP) layer is used to reduce the spatial size of the resulting feature
maps. Finally, the final band weights $\mathbf{W}$ is generated by
a few fully connected layer and used to reweight the spectral bands.
BS-Net-Conv adopts a convolutional-deconvolutional network (Conv-DeConv
Net) to implement the RecNet. Similar to the classical auto-encoder,
Conv-DeConv Net includes a convolutional encoder which extracts deep
features and a deconvolutional decoder which up-samples feature maps. 

Instead of using single pixels, BS-Net-Conv takes 3-D HSI patches
which includes spectral and spatial information as the training samples.
To generate enough training samples, we use a rectangular window of
size $a\times a$ to slides across the given HSI with stride $t$.
The generated training samples can be denoted as $\mathbf{X}\in\mathbb{R}^{S\times a\times a\times b}$,
where $S=\frac{M-a}{t}\times\frac{N-a}{t}+1$. Notice that the number
of training samples in BS-Net-Conv is less than that in BS-Net-FC. 

\subsection{Remarks on BS-Net framework}

The key to our proposed BS-Net framework is to use deep neural networks
to explicitly learn spectral bands weights. Compared with the existing
band selection methods, the framework has the following advantages.
The first is the framework is end-to-end trainable, making it easy
to combine with specific tasks and existed neural networks, such as
deep learning based HSI classification. The second is the framework
is capable of adaptively exacting spectral and spatial information,
which avoids hand-designed features and reduces the noise effect.
The third is the framework is nonlinear, enabling it to make full
exploration of the nonlinear relationship between bands. The fourth
is the framework is flexible to be implemented with diverse networks. 
\begin{center}
\begin{table}[tbh]
\caption{Summary of Indian Pines, Pavia University, and Salinas data sets.\label{tab:Data-sets-descriptions} }
\centering{}%
\begin{tabular}{|c|c|c|c|}
\hline 
Data sets & Indina Pines & Pavia University & Salinas\tabularnewline
\hline 
\hline 
Pixels & 145$\times$145 & 610$\times$340 & 512$\times$217\tabularnewline
\hline 
Channels & 200 & 103 & 204\tabularnewline
\hline 
Classes & 16 & 9 & 16\tabularnewline
\hline 
Labeled pixels & 10249 & 42776 & 54129\tabularnewline
\hline 
Sensor & AVIRIS & ROSIS & AVIRIS\tabularnewline
\hline 
\end{tabular}
\end{table}
\par\end{center}

\begin{center}
\begin{table}[tbh]
\caption{Hyper-parameters settings for different BS methods.\label{tab:Hyperparameters-settings-for-baseline}}
\centering{}%
\begin{tabular}{|c|c|}
\hline 
Baselines & Hyper-parameters\tabularnewline
\hline 
\hline 
ISSC & $\lambda=1e5$\tabularnewline
\hline 
SpaBS & $\lambda=1e2$\tabularnewline
\hline 
MVPCA & \textendash{}\tabularnewline
\hline 
SNMF & $maxiter=100$\tabularnewline
\hline 
MOBS & $maxiter=100$,$NP=100$\tabularnewline
\hline 
OPBS & \textendash{}\tabularnewline
\hline 
BS-Net-FC & $\lambda=1e-2$,$\eta=2e-3$,$maxiter=100$\tabularnewline
\hline 
BS-Net-Conv & $\lambda=1e-2$,$\eta=2e-3$,$maxiter=100$\tabularnewline
\hline 
\end{tabular}
\end{table}
\par\end{center}

\begin{center}
\begin{table}[tbh]
\caption{Configuration of BS-Net-FC for Indian Pines data set.\label{tab:BS-Net-architecture-details-FC} }
\centering{}%
\begin{tabular}{|c|c|c|c|}
\hline 
Branch & Layer & Hidden Neurons & Activation\tabularnewline
\hline 
\hline 
\multirow{4}{*}{BAM} & Input & 200 & \textendash{}\tabularnewline
\cline{2-4} 
 & FC1-1 & 64 & ReLU\tabularnewline
\cline{2-4} 
 & FC1-2 & 128 & ReLU\tabularnewline
\cline{2-4} 
 & FC1-3 & 200 & Sigmoid\tabularnewline
\hline 
\multirow{4}{*}{RecNet} & FC2-1 & 64 & ReLU\tabularnewline
\cline{2-4} 
 & FC2-2 & 128 & ReLU\tabularnewline
\cline{2-4} 
 & FC2-3 & 256 & ReLU\tabularnewline
\cline{2-4} 
 & FC2-4 & 200 & Sigmoid\tabularnewline
\hline 
\end{tabular}
\end{table}
\par\end{center}

\begin{center}
\begin{table}[tbh]
\caption{Configuration of BS-Net-Conv for Indian Pines data set.\label{tab:BS-Net-architecture-details-Conv} }
\centering{}%
\begin{tabular}{|c|c||c|c|}
\hline 
Branch & Layer & Kernel & Activation\tabularnewline
\hline 
\hline 
\multirow{4}{*}{BAM} & Conv1 & $3\times3\times64$ & ReLU\tabularnewline
\cline{2-4} 
 & GP & \textendash{} & \textendash{}\tabularnewline
\cline{2-4} 
 & FC1 & 128 & ReLU\tabularnewline
\cline{2-4} 
 & FC2 & 200 & Sigmoid\tabularnewline
\hline 
\multirow{5}{*}{RecNet} & Conv1-1 & $3\times3\times128$ & ReLU\tabularnewline
\cline{2-4} 
 & Conv1-2 & $3\times3\times64$ & ReLU\tabularnewline
\cline{2-4} 
 & DeConv1-2 & $3\times3\times64$ & ReLU\tabularnewline
\cline{2-4} 
 & DeConv1-1 & $3\times3\times128$ & ReLU\tabularnewline
\cline{2-4} 
 & Conv2-1 & $1\times1\times200$ & Sigmoid\tabularnewline
\hline 
\end{tabular}
\end{table}
\par\end{center}

\section{Results \label{sec:Results}}

\subsection{Setup }

In this section, we will widely evaluate the performance of the proposed
BS-Nets on three real HSI data sets: Indian Pines, Pavia University,
and Salinas. The summary of the three data sets is shown in Table
\ref{tab:Data-sets-descriptions}. For each data set, we first investigate
the model convergence by analyzing the training loss, classification
accuracy, and band weights. Next, we compare the classification performance
for BS-Nets with many existing band selection methods, i.e., ISSC
\cite{HSI_BS-ISSC-SunWW-JSTARS-2015}, SpaBS \cite{HSI_BS-SpaBS-TargetDectection-SunK-GRSL-2015},
MVPCA \cite{HSI_BS-MVPCA-ChangCI-TGRS-1999}, SNMF \cite{HSI_BS-SNMF-JiM-FITEE},
MOBS \cite{HSI_BS-EvoMultiObj-GongMG-TGRS-2016}, and OPBS \cite{HSI-Band_Selection-OPBS-Geometry-Based-ZhangW-TGRS-2018}.
The hyper-parameters settings of all the methods are listed in Table
\ref{tab:Hyperparameters-settings-for-baseline}. To better demonstrate
the performance, we also compare with all bands. Finally, we make
a deep analysis on the selected band subsets from aspects of visualization
and quantification. 

Similar to the evaluation strategy adopted in \cite{HSI_BS-EvoMultiObj-GongMG-TGRS-2016,HSI-Band_Selection-OPBS-Geometry-Based-ZhangW-TGRS-2018},
we use Support Vector Machine (SVM) with radial basis function kernel
as the classifier to evaluate the classification performance of the
selected band subsets. For the sake of fairness, we randomly select
$5\%$ of labeled samples from each data set for training set, and
the rest for testing set. Three popular quantitative indices, i.e.,
 Overall Accuracy (OA), Average Accuracy (AA), and Kappa coefficient
(Kappa) are calculated by evaluating each BS method for $20$ independent
runs. 

To quantitatively analyze the selected band subsets, the entropy and
mean spectral divergence (MSD) \cite{HSI-Band_Selection-Volume-Gradient-Based-GengX-TGRS-2014,HSI_BS-EvoMultiObj-GongMG-TGRS-2016}
of band subsets are calculated. For a single band $\boldsymbol{B}_{i}$,
its entropy is defined as follows: 

\begin{equation}
H\left(\boldsymbol{B}_{i}\right)=-\underset{y\in\Psi}{\sum}p\left(y\right)\log\left(p\left(y\right)\right)
\end{equation}
Here $y$ denotes a gray level of the histogram of the $i$-th band
$\boldsymbol{B}_{i}$, and $p\left(y\right)=\frac{n\left(y\right)}{N\times M}$
indicates the ratio (probability) of the number of $y$ to that all
pixels. According to the characteristic of entropy, the larger the
entropy is, the more image details the band contains \cite{HSI_BS-EvoMultiObj-GongMG-TGRS-2016}.
The MSD is an average measurement index for a band subset $\boldsymbol{B}$,
which is expressed as:

\begin{equation}
M\left(\boldsymbol{B}\right)=\frac{2}{k\left(k-1\right)}\stackrel[i=1]{k}{\sum}\stackrel[j=1]{k}{\sum}D_{SKL}\left(\boldsymbol{B}_{i}\parallel\boldsymbol{B}_{j}\right)\label{eq:msd}
\end{equation}
Where $D_{SKL}$ is the symmetrical Kullback\textendash Leibler divergence
which measures the dissimilarity between $\boldsymbol{B}_{i}$ and
$\boldsymbol{B}_{j}$. Specifically, $D_{SKL}$ is defined as follows: 

\begin{equation}
D_{SKL}\left(\boldsymbol{B}_{i}\parallel\boldsymbol{B}_{j}\right)=D_{KL}\left(\boldsymbol{B}_{i}\parallel\boldsymbol{B}_{j}\right)+D_{KL}\left(\boldsymbol{B}_{j}\parallel\boldsymbol{B}_{i}\right)
\end{equation}
Here $D_{KL}\left(\boldsymbol{B}_{i}\parallel\boldsymbol{B}_{j}\right)$
can be computed from the gray histogram information. From Eq. \eqref{eq:msd},
MSD evaluates the redundancy among the selected bands, that is, the
larger the value of the MSD is, the less redundancy is contained among
the selected bands.

The configuration of BS-Nets implemented in our experiments are shown
in Table \ref{tab:BS-Net-architecture-details-FC} and Table \ref{tab:BS-Net-architecture-details-Conv}.
In reprocessing, we scale all the HSI pixel values to the range $\left[0,1\right]$.
All the baseline methods are evaluated with Python 3.5 running on
an Intel Xeon E5-2620 2.10 GHz CPU with 32 GB RAM. In addition, we
implement BS-Nets with TensorFlow-GPU 1.6 \footnote{https://tensorflow.google.cn}
and accelerate them on a NVIDIA TITAN Xp GPU with 11 GB graphic memory.
One may refer to \href{https://github.com/AngryCai}{https://github.com/AngryCai}
for the source codes and trained models. 

\subsection{Results on Indian Pines Data Set }

\subsubsection{Data Set}

This scene was gathered by AVIRIS sensor over the Indian Pines test
site in North-western Indiana and consists of $145\times145$ pixels
and $224$ spectral reflectance bands in the wavelength range $0.4\lyxmathsym{\textendash}2.5$
$\left(\times10^{-6}\right)$ meters. The scene contains two-thirds
agriculture, and one-third forest or other natural perennial vegetation.
There are two major dual lane highways, a rail line, as well as some
low-density housing, other built structures, and smaller roads. Since
the scene is taken in June some of the crops presents, corn, soybeans,
are in early stages of growth with less than $5\%$ coverage. The
ground-truth available is designated into sixteen classes and is not
all mutually exclusive. We have also reduced the number of bands to
$200$ by removing bands covering the region of water absorption:
$[104-108]$, $[150-163]$, $220$.

\begin{figure}[tbh]
\subfloat[BS-Net-FC]{\begin{centering}
\includegraphics[width=0.48\columnwidth]{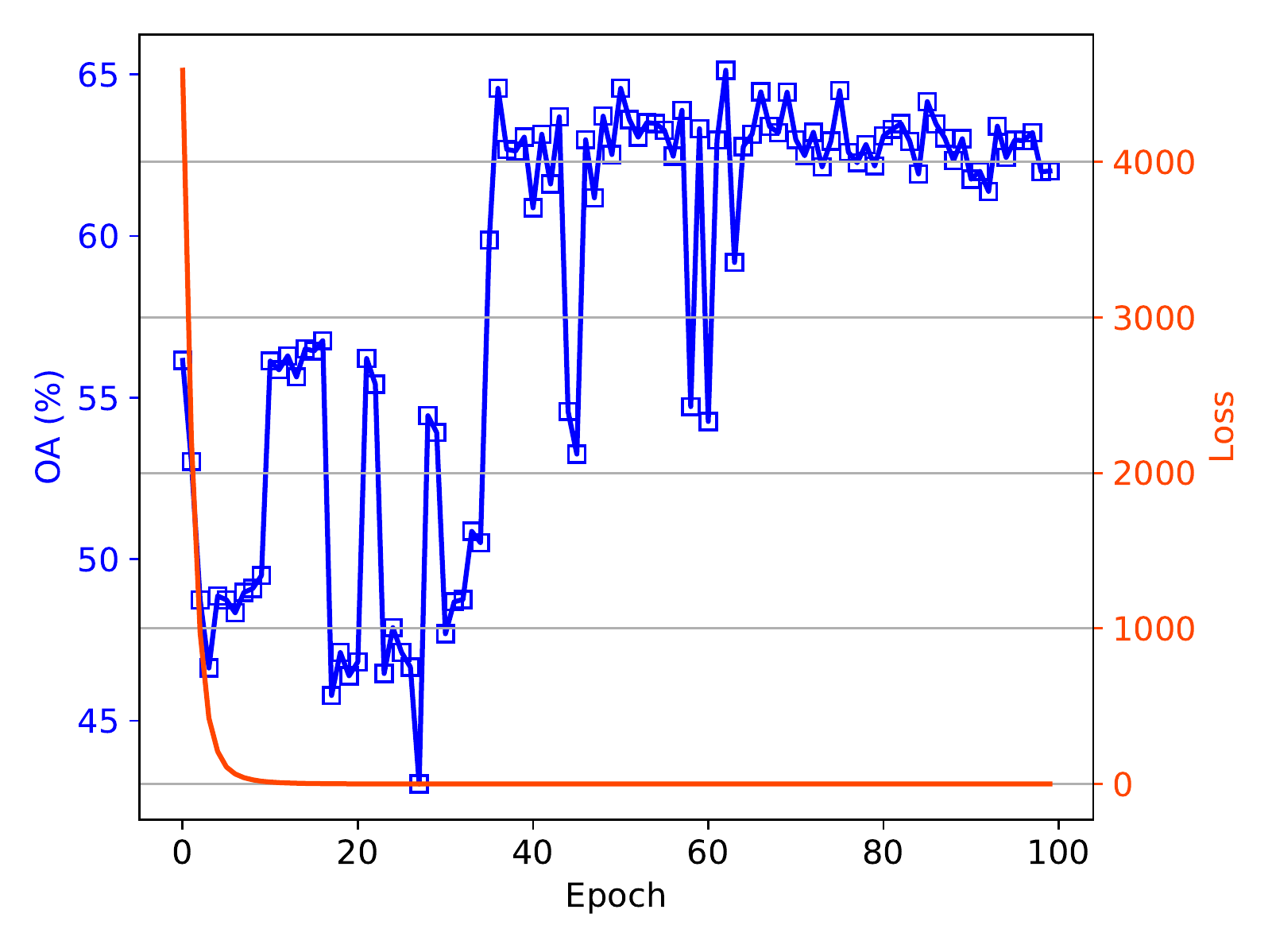}
\par\end{centering}
}\subfloat[BS-Net-Conv]{\begin{centering}
\includegraphics[width=0.48\columnwidth]{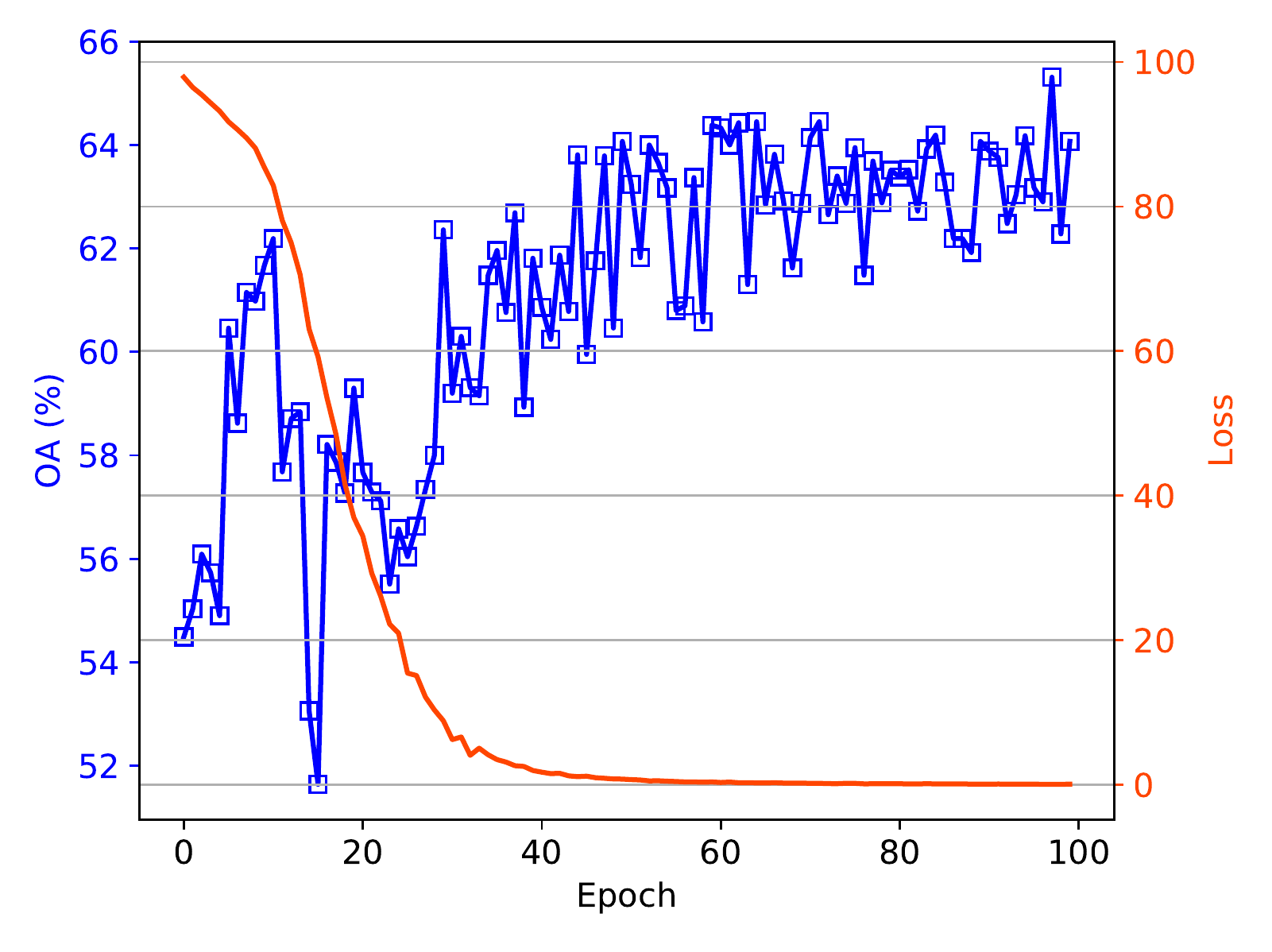}
\par\end{centering}
}

\subfloat[BS-Net-FC]{\begin{centering}
\includegraphics[width=0.48\columnwidth]{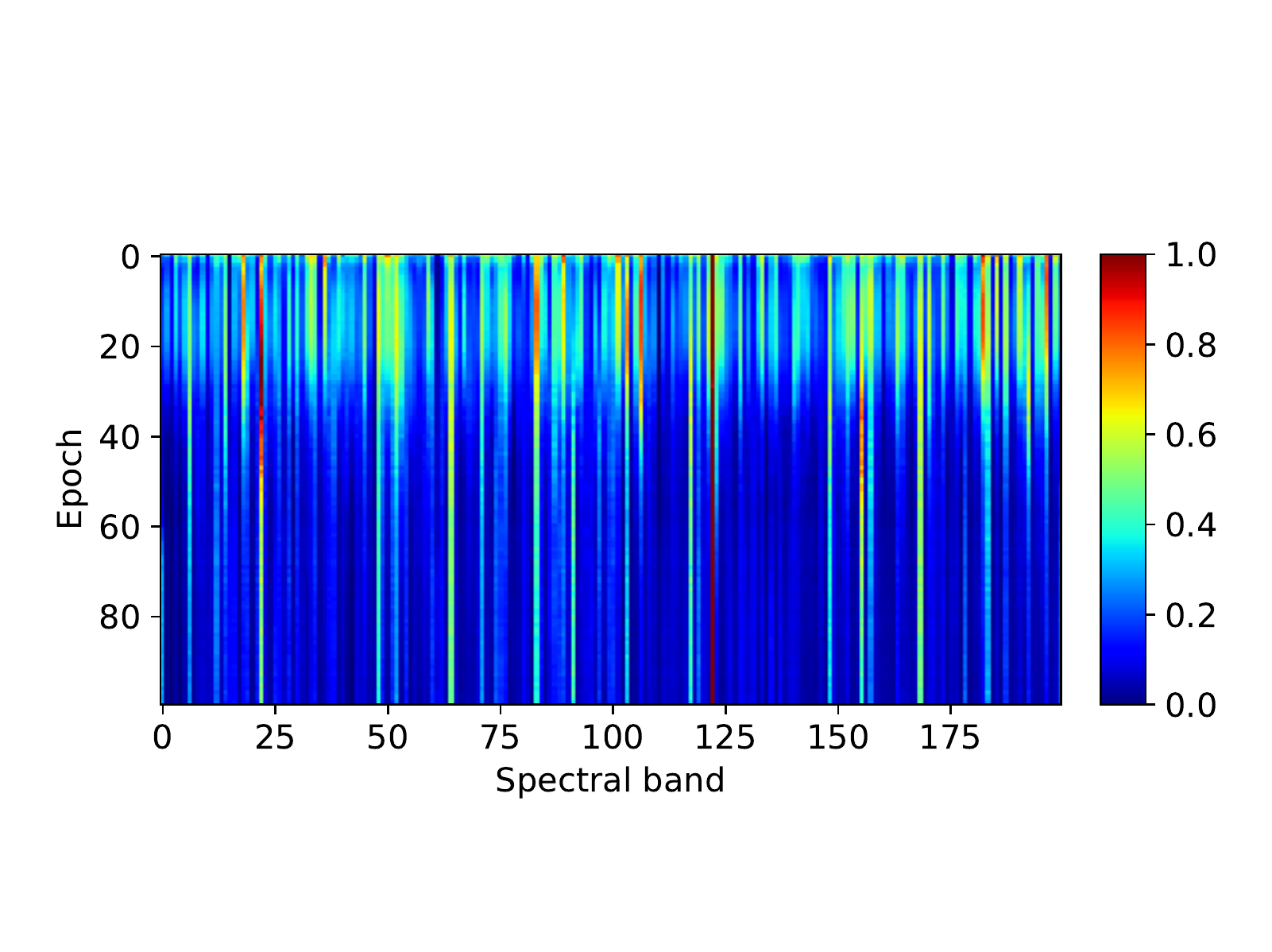}
\par\end{centering}
}\subfloat[BS-Net-Conv]{\begin{centering}
\includegraphics[width=0.48\columnwidth]{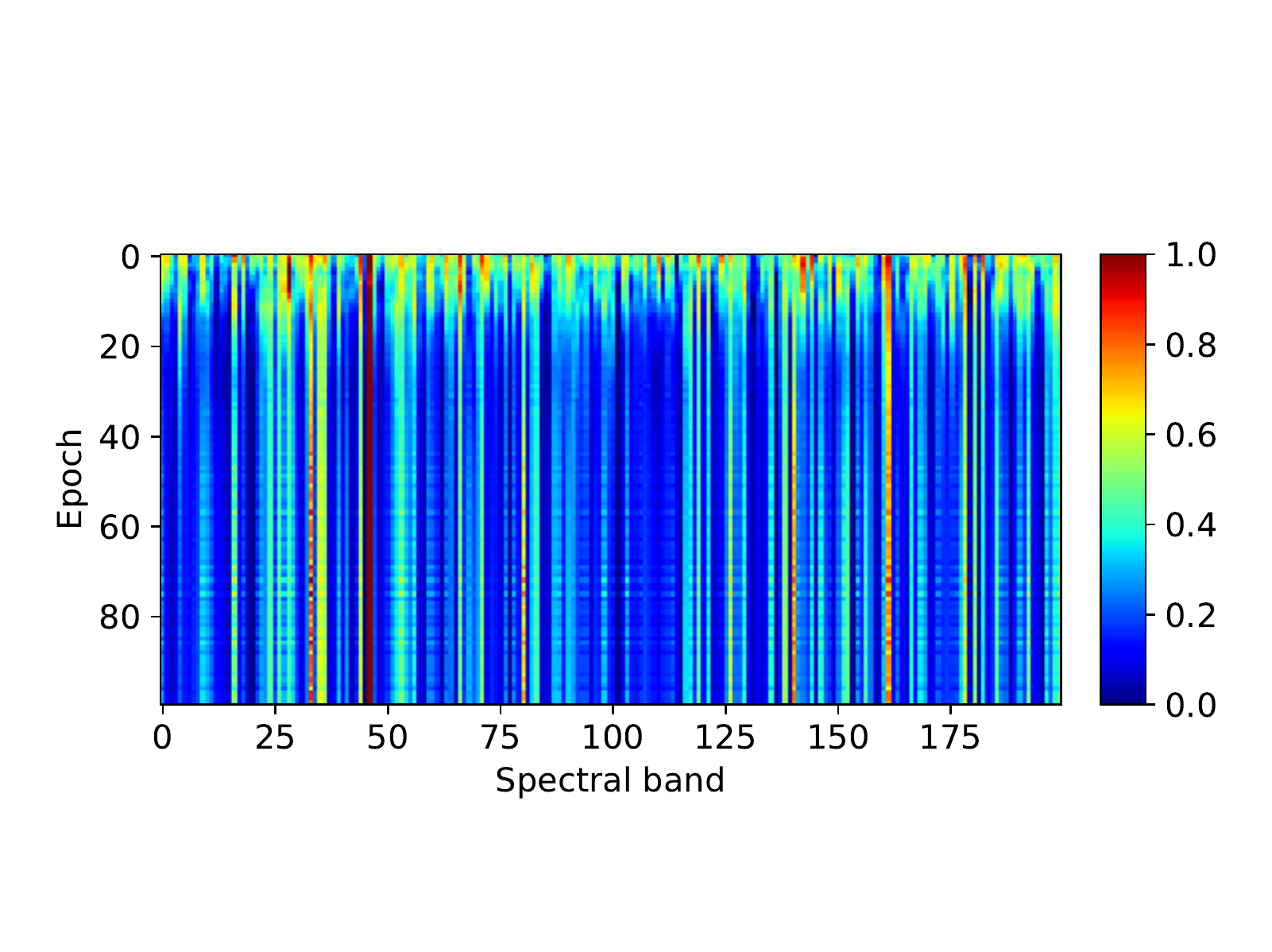}
\par\end{centering}
}

\caption{Analysis of the convergence of BS-Nets on Indian Pines data set. Visualization
of loss versus accuracy under different iterations for (a) BS-Net-FC
and (b) BS-Net-Conv. Visualization of average band weights under varying
iterations for (c) BS-Net-FC and (d) BS-Net-Conv. \label{fig:Analysis-of-band-Indian}}
\end{figure}

\subsubsection{Analysis of Convergence of BS-Nets }

To analyze the convergence of BS-Nets, we train BS-Nets for 100 iterations
and plot their training loss curves and classification accuracy using
the best $5$ significant bands. The results of BS-Net-FC and BS-Net-Conv
are shown in Fig. \ref{fig:Analysis-of-band-Indian} (a) and (b),
respectively. Observing from Fig. \ref{fig:Analysis-of-band-Indian}
(a)-(b), the reconstruction errors decrease with iterations, and at
the same time, the classification accuracies increase. The loss values
of BS-Net-FC is very close to zero after 20 iterations and the classification
accuracy finally stabilizes around $63\%$ after 40 interactions.
The similar trend can be found from BS-Net-Conv, showing that the
proposed BS-Nets are easy to train with high convergency speed. Furthermore,
we can find that the classification accuracy is increased from $42\%$
to $63\%$ for BS-Net-FC and $52\%$ to $64\%$ for BS-Net-Conv. Therefore,
the well-optimized BS-Nets  can respectively achieve $21\%$ and $12\%$
improvement in terms of classification accuracy, which demonstrates
the effectiveness of the proposed BS-Nets. 

We further visualize the average band weights obtained by BS-Net-FC
and BS-Net-Conv with different iterations in Fig. \ref{fig:Analysis-of-band-Indian}
(c) and (d). To better show the change trend of band weights, we scale
these weights into range $\left[0,1\right]$. The horizontal and vertical
axis represent the band number and iterations, respectively, where
each column depicts the change of one band's weights. As we can see,
the the band weights distribution becomes gradually sparser and sparser.
Meanwhile, the informative bands become easier to be distinguished
due to the trivial bands will finally be assigned with very small
weights. 

\subsubsection{Performance Comparison}

\begin{figure*}[tbh]
\begin{centering}
\subfloat[OA]{\includegraphics[width=0.65\columnwidth]{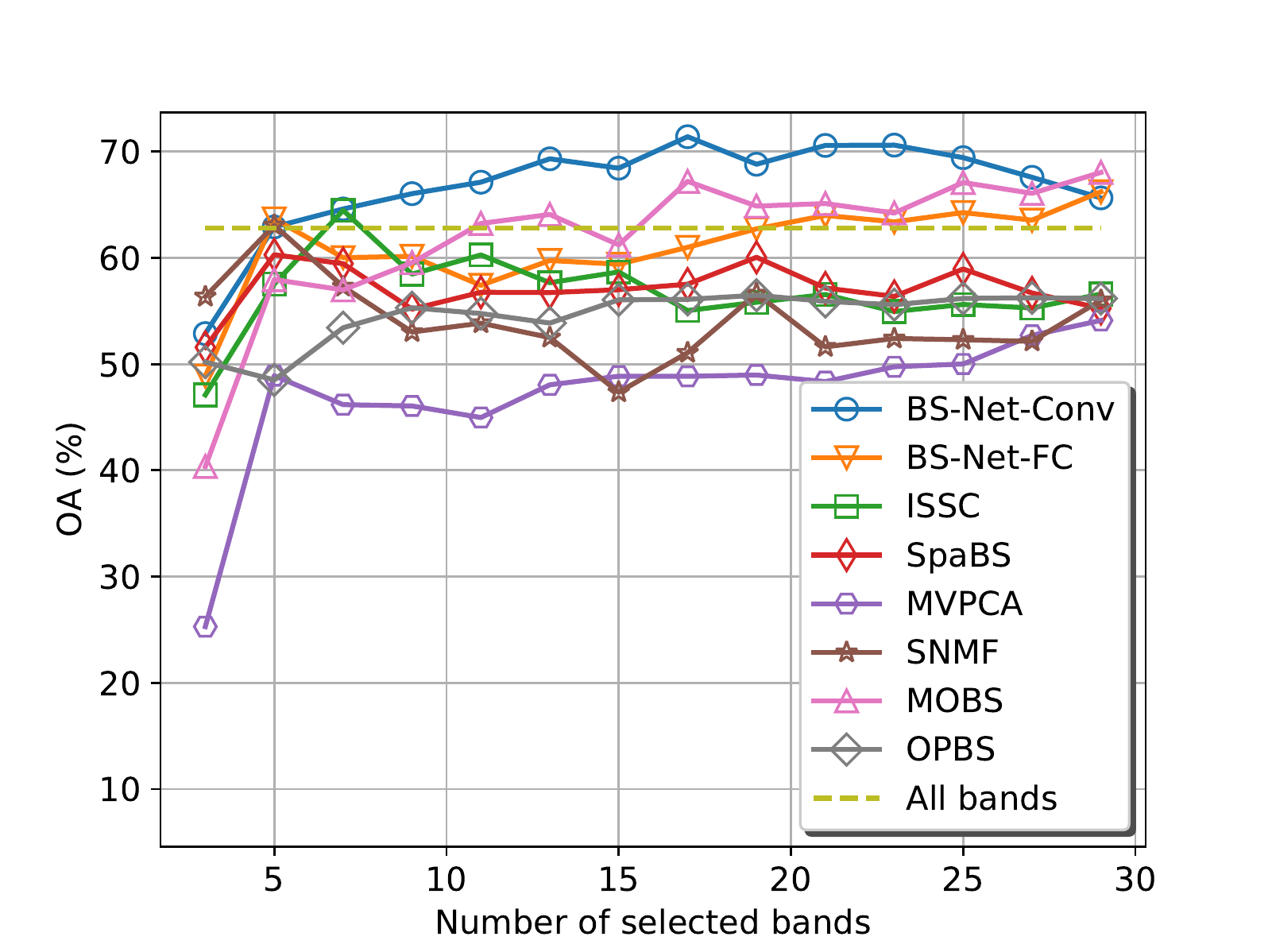}

}\subfloat[AA ]{\includegraphics[width=0.65\columnwidth]{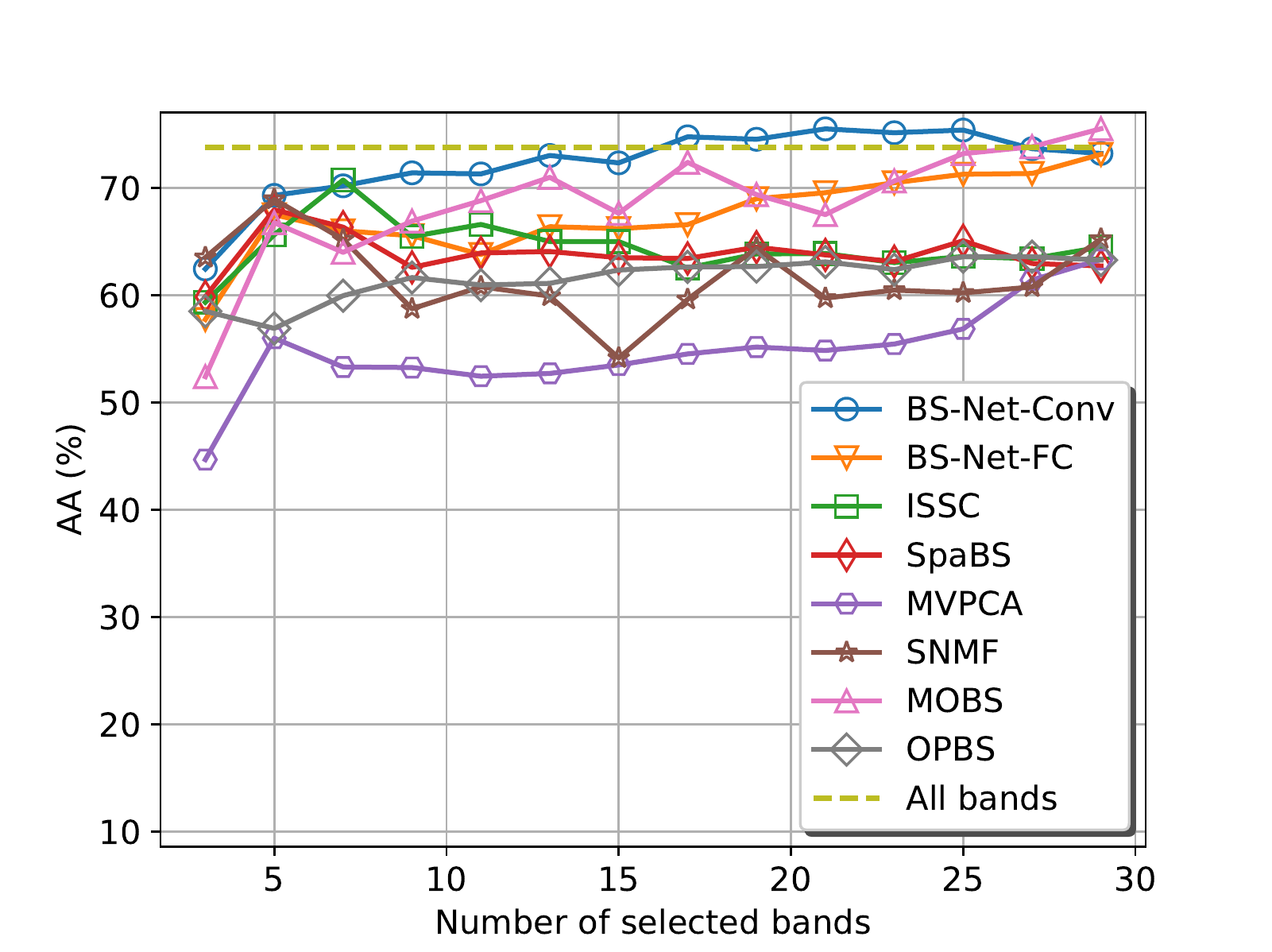}

}\subfloat[Kappa]{\includegraphics[width=0.65\columnwidth]{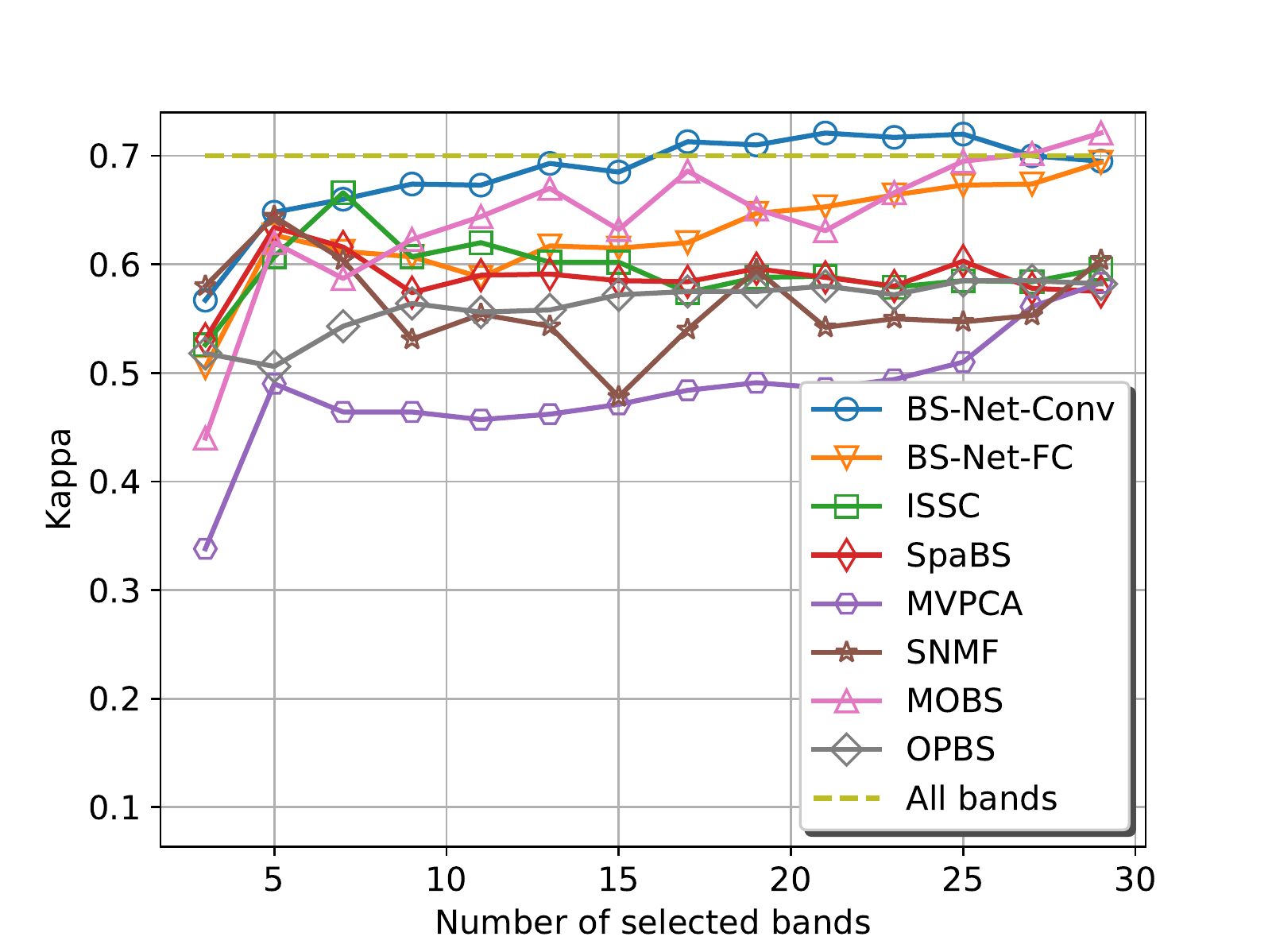}

}
\par\end{centering}
\caption{Performance comparison of different BS methods with different band
subset sizes on Indian Pines data set. (a) OA; (b) AA; (c) Kappa.
\label{fig:Performance-comparison-of-indan}}
\end{figure*}

To demonstrate the effectiveness of both BS-Nets, we compare the classification
performance of different BS methods under different band subset sizes.
Three indices, i.e.,  OA, AA, and Kappa, are computed under band subset
sizes ranging from 3 to 30 with 2-band interval. All bands performance
is also compared as an important reference. To ensure a reliable classification
result, we conduct each method for 20 times and randomly select the
training set and test set during each time. Fig. \ref{fig:Performance-comparison-of-indan}
(a)-(c) shows the average comparison results of OA, AA, and Kappa,
respectively. From Fig. \ref{fig:Performance-comparison-of-indan}
(a)-(c), BS-Net-Conv achieves the best OA, AA, and Kappa when the
band subset size is larger than $5$, while BS-Net-FC achieves comparable
performance with MOBS but is superior to the other 5 competitors.
BS-Net-Conv respectively requires $5$, $17$, and $17$ bands to
achieve better OA, AA, and Kappa than all bands, while other methods
either worst than all bands or have to use more bands. Notice that
a counter-intuitive phenomenon that the classification performance
is not always increased by selecting more bands can be found from
the results. For example, ISSC and SNMF display significant decreasing
trend when the band subset size is over $7$. The phenomenon can be
explained as the so-called Hughes phenomenon \cite{Hughes-Phenomenon-Explanation-TPAMI-1979},
i.e., the classification accuracy increases first and then decreases
with the selected bands. However, even if so, BS-Net occurs the phenomenon
obviously later than other methods, which means our methods are able
to select more effective band subset.

\begin{table*}[tbh]
\caption{Classification performance of different methods using 15 bands on
Indian Pines data set. \label{tab:Performance-comparison-of-indian}}
\centering{}{\scriptsize{}}%
\begin{tabular}{|c|c|c||c|c|c|c|c|c|c|c|}
\hline 
{\scriptsize{}No.} & {\scriptsize{}\#Train} & {\scriptsize{}\#Test} & {\scriptsize{}ISSC} & {\scriptsize{}SpaBS} & {\scriptsize{}MVPCA} & {\scriptsize{}SNMF} & {\scriptsize{}MOBS} & {\scriptsize{}OPBS} & {\scriptsize{}BS-Net-FC} & {\scriptsize{}BS-Net-Conv}\tabularnewline
\hline 
\hline 
{\scriptsize{}1} & {\scriptsize{}2} & {\scriptsize{}44} & {\scriptsize{}31.14$\pm$21.19} & {\scriptsize{}27.24$\pm$27.72 } & {\scriptsize{}19.43$\pm$13.84} & {\scriptsize{}14.29$\pm$16.18} & {\scriptsize{}55.03$\pm$20.82} & {\scriptsize{}34.97$\pm$24.82} & {\scriptsize{}38.98$\pm$22.78} & \textbf{\scriptsize{}66.41$\pm$24.59}\tabularnewline
\hline 
{\scriptsize{}2} & {\scriptsize{}71} & {\scriptsize{}1357} & {\scriptsize{}64.13$\pm$3.21} & {\scriptsize{}60.35$\pm$4.34} & {\scriptsize{}48.13$\pm$4.04} & {\scriptsize{}57.36$\pm$4.90} & {\scriptsize{}65.56$\pm$3.12} & {\scriptsize{}59.46$\pm$4.19} & {\scriptsize{}67.49$\pm$4.37} & \textbf{\scriptsize{}75.67$\pm$2.91}\tabularnewline
\hline 
{\scriptsize{}3} & {\scriptsize{}41} & {\scriptsize{}789} & {\scriptsize{}52.17$\pm$5.64} & {\scriptsize{}53.62$\pm$3.35} & {\scriptsize{}35.68$\pm$3.80} & {\scriptsize{}47.56$\pm$4.82} & {\scriptsize{}58.19$\pm$4.96} & {\scriptsize{}48.60$\pm$5.08} & {\scriptsize{}59.62$\pm$3.79} & \textbf{\scriptsize{}64.39$\pm$5.30}\tabularnewline
\hline 
{\scriptsize{}4} & {\scriptsize{}12} & {\scriptsize{}225} & {\scriptsize{}36.78$\pm$8.34} & {\scriptsize{}39.28$\pm$10.40} & {\scriptsize{}12.74$\pm$5.34} & {\scriptsize{}35.64$\pm$7.87} & {\scriptsize{}46.60$\pm$11.09} & {\scriptsize{}31.24$\pm$10.50} & {\scriptsize{}40.08$\pm$8.26} & \textbf{\scriptsize{}61.41$\pm$12.18}\tabularnewline
\hline 
{\scriptsize{}5} & {\scriptsize{}24} & {\scriptsize{}459} & {\scriptsize{}73.98$\pm$8.39} & {\scriptsize{}81.16$\pm$3.46} & {\scriptsize{}62.32$\pm$8.33} & {\scriptsize{}70.39$\pm$8.17} & {\scriptsize{}79.39$\pm$5.52} & {\scriptsize{}79.79$\pm$5.17} & {\scriptsize{}80.90$\pm$5.41} & \textbf{\scriptsize{}85.78$\pm$4.22}\tabularnewline
\hline 
{\scriptsize{}6} & {\scriptsize{}37} & {\scriptsize{}693} & {\scriptsize{}83.49$\pm$3.94} & {\scriptsize{}82.21$\pm$7.18} & {\scriptsize{}83.08$\pm$4.36} & {\scriptsize{}75.41$\pm$6.43} & {\scriptsize{}88.57$\pm$4.69} & {\scriptsize{}89.02$\pm$3.78} & {\scriptsize{}90.93$\pm$4.66} & \textbf{\scriptsize{}93.63$\pm$2.53}\tabularnewline
\hline 
{\scriptsize{}7} & {\scriptsize{}1} & {\scriptsize{}27} & {\scriptsize{}29.55$\pm$28.04} & {\scriptsize{}37.97$\pm$22.80} & {\scriptsize{}19.79$\pm$18.58 } & {\scriptsize{}12.99$\pm$15.46} & {\scriptsize{}39.16$\pm$32.62} & {\scriptsize{}29.11$\pm$23.12} & \textbf{\scriptsize{}52.44$\pm$28.16} & {\scriptsize{}45.30$\pm$39.45}\tabularnewline
\hline 
{\scriptsize{}8} & {\scriptsize{}24} & {\scriptsize{}454} & {\scriptsize{}88.33$\pm$4.63} & {\scriptsize{}86.75$\pm$10.38} & {\scriptsize{}85.15$\pm$5.65} & {\scriptsize{}88.08$\pm$6.45} & {\scriptsize{}92.04$\pm$5.84} & {\scriptsize{}90.68$\pm$5.47} & {\scriptsize{}93.77$\pm$4.85} & \textbf{\scriptsize{}96.86$\pm$2.82}\tabularnewline
\hline 
{\scriptsize{}9} & {\scriptsize{}1} & {\scriptsize{} 19} & {\scriptsize{}10.91$\pm$12.80} & {\scriptsize{}6.88$\pm$9.32} & {\scriptsize{}7.55$\pm$10.33} & {\scriptsize{}7.31$\pm$10.62} & \textbf{\scriptsize{}48.27$\pm$31.66} & {\scriptsize{}7.46$\pm$10.87} & {\scriptsize{}24.26$\pm$22.49} & {\scriptsize{}28.07$\pm$35.09}\tabularnewline
\hline 
{\scriptsize{}10} & {\scriptsize{}49} & {\scriptsize{}923} & {\scriptsize{}52.69$\pm$3.87} & {\scriptsize{}53.21$\pm$5.10} & {\scriptsize{}40.71$\pm$4.74 } & {\scriptsize{}50.07$\pm$5.73} & {\scriptsize{}52.34$\pm$5.33} & {\scriptsize{}52.27$\pm$5.09} & {\scriptsize{}60.72$\pm$4.19} & \textbf{\scriptsize{}66.45$\pm$5.52}\tabularnewline
\hline 
{\scriptsize{}11} & {\scriptsize{}123} & {\scriptsize{}2332} & {\scriptsize{}62.14$\pm$2.39} & {\scriptsize{}57.87$\pm$2.42} & {\scriptsize{}50.58$\pm$3.07} & {\scriptsize{}56.52$\pm$2.91} & {\scriptsize{}62.10$\pm$3.81} & {\scriptsize{}61.37$\pm$3.68} & {\scriptsize{}67.04$\pm$3.69} & \textbf{\scriptsize{}71.69$\pm$2.91}\tabularnewline
\hline 
{\scriptsize{}12} & {\scriptsize{}30} & {\scriptsize{}563} & {\scriptsize{}35.02$\pm$5.38} & {\scriptsize{}43.15$\pm$7.78} & {\scriptsize{}34.44$\pm$6.57} & {\scriptsize{}31.09$\pm$5.28} & {\scriptsize{}57.20$\pm$8.34} & {\scriptsize{}30.52$\pm$5.29} & {\scriptsize{}42.95$\pm$6.48} & \textbf{\scriptsize{}65.30$\pm$5.76}\tabularnewline
\hline 
{\scriptsize{}13} & {\scriptsize{}10} & {\scriptsize{}195} & {\scriptsize{}86.78$\pm$8.03} & {\scriptsize{}91.18$\pm$7.31} & {\scriptsize{}82.45$\pm$8.59} & {\scriptsize{}81.06$\pm$10.54} & {\scriptsize{}91.73$\pm$5.11} & {\scriptsize{}80.96$\pm$12.79 } & {\scriptsize{}93.75$\pm$4.92} & \textbf{\scriptsize{}94.35$\pm$6.33}\tabularnewline
\hline 
{\scriptsize{}14} & {\scriptsize{}63} & {\scriptsize{}1202} & {\scriptsize{}85.76$\pm$3.50} & {\scriptsize{}89.83$\pm$2.32} & {\scriptsize{}85.94$\pm$3.48} & {\scriptsize{}86.17$\pm$3.64} & {\scriptsize{}86.99$\pm$3.57} & {\scriptsize{}86.44$\pm$5.21} & {\scriptsize{}89.08$\pm$3.49} & \textbf{\scriptsize{}89.46$\pm$3.07}\tabularnewline
\hline 
{\scriptsize{}15} & {\scriptsize{}19} & {\scriptsize{}367} & {\scriptsize{}31.21$\pm$7.32} & {\scriptsize{}32.80 $\pm$7.44} & {\scriptsize{}29.4$\pm$4.76} & {\scriptsize{}29.08$\pm$7.53} & {\scriptsize{}36.90$\pm$8.13} & {\scriptsize{}31.73$\pm$8.20} & {\scriptsize{}39.91$\pm$8.07} & \textbf{\scriptsize{}47.53$\pm$10.18}\tabularnewline
\hline 
{\scriptsize{}16} & {\scriptsize{}5} & {\scriptsize{} 88} & {\scriptsize{}80.76$\pm$18.76} & {\scriptsize{}71.1$\pm$21.56} & {\scriptsize{}76.28$\pm$8.33} & \textbf{\scriptsize{}82.98$\pm$5.47} & {\scriptsize{}81.76$\pm$5.3} & {\scriptsize{}80.86$\pm$7.46} & {\scriptsize{}81.90$\pm$6.22} & {\scriptsize{}76.98$\pm$21.11}\tabularnewline
\hline 
\hline 
\multicolumn{3}{|c||}{{\scriptsize{}OA (\%)}} & {\scriptsize{}56.55$\pm$2.47} & {\scriptsize{}57.16$\pm$2.25} & {\scriptsize{}48.35$\pm$1.92} & {\scriptsize{}51.62$\pm$2.02} & {\scriptsize{}65.11$\pm$2.76} & {\scriptsize{}55.90$\pm$2.39} & {\scriptsize{}63.99$\pm$2.51} & \textbf{\scriptsize{}70.58$\pm$2.87}\tabularnewline
\hline 
\multicolumn{3}{|c||}{{\scriptsize{}AA (\%)}} & {\scriptsize{}63.91$\pm$0.73} & {\scriptsize{}63.78$\pm$0.99} & {\scriptsize{}54.87$\pm$1.59} & {\scriptsize{}59.75$\pm$1.15} & {\scriptsize{}67.54$\pm$0.83} & {\scriptsize{}63.13$\pm$1.09} & {\scriptsize{}69.58$\pm$1.00} & \textbf{\scriptsize{}75.53$\pm$0.69}\tabularnewline
\hline 
\multicolumn{3}{|c||}{{\scriptsize{}Kappa}} & {\scriptsize{}0.589$\pm$0.008} & {\scriptsize{}0.588$\pm$0.011} & {\scriptsize{}0.486$\pm$0.018} & {\scriptsize{}0.542$\pm$0.013} & {\scriptsize{}0.631$\pm$0.009} & {\scriptsize{}0.580$\pm$0.012} & {\scriptsize{}0.653$\pm$0.011} & \textbf{\scriptsize{}0.721$\pm$0.008}\tabularnewline
\hline 
\end{tabular}{\scriptsize \par}
\end{table*}

In Table \ref{tab:Performance-comparison-of-indian}, we show the
detailed classification performance of selecting best $19$ bands
for different methods. As we can see, BS-Net-FC and BS-Net-Conv outperform
all the competitors in terms of OA, AA, Kappa. For some classes which
contain limited training samples, such as No. 7 and No. 9 class, BS-Net-FC
and BS-Net-Conv can still yield much better or comparable accuracy
compared with the other methods. Compared with BS-Net-FC, BS-Net-Conv
achieves $6.59\%$ improvement in terms of OA, showing that spectral-spatial
information is more effective for band selection than using only spectral
information. 

\subsubsection{Analysis of the Selected Bands}

\begin{table}[tbh]
\caption{The best 15 bands of Indian Pines data set selected by different BS
methods.\label{tab:The-best-20-IndianP}}
\centering{}{\scriptsize{}}%
\begin{tabular}{|c|c|}
\hline 
{\scriptsize{}Methods} & {\scriptsize{}Selected Bands}\tabularnewline
\hline 
\hline 
{\scriptsize{}BS-Net-FC} & {\scriptsize{}{[}165, 38, 51, 65, 12, 100, 0, 71, 5, 60, 88, 26, 164,
75, 74{]}}\tabularnewline
\hline 
{\scriptsize{}BS-Net-Conv} & {\scriptsize{}{[}46,33,140,161,80,35,178,44,126,36,138,71,180,66,192{]}}\tabularnewline
\hline 
{\scriptsize{}ISSC} & {\scriptsize{}{[}171,130,67,85,182,183,47,143,138,90,139,141,25,142,21{]}}\tabularnewline
\hline 
{\scriptsize{}SpaBS} & {\scriptsize{}{[}7, 96, 52, 171, 53, 3, 76, 75, 74, 95, 77, 73, 78,
54, 81{]}}\tabularnewline
\hline 
{\scriptsize{}MVPCA} & {\scriptsize{}{[}167,74,168,0,147,165,161,162,152,19,160,119,164,159,157{]}}\tabularnewline
\hline 
{\scriptsize{}SNMF} & {\scriptsize{}{[}23,197,198,94,76,2,87,105,143,145,11,84,132,108,28{]}}\tabularnewline
\hline 
{\scriptsize{}MOBS} & {\scriptsize{}{[}5,6,19,24,45,48,105,114,129,142,144,160,168,172,181{]}}\tabularnewline
\hline 
{\scriptsize{}OPBS} & {\scriptsize{}{[}28, 41, 60, 0, 74, 34, 88, 19, 17, 33, 56, 87, 22,
31, 73{]}}\tabularnewline
\hline 
\end{tabular}{\scriptsize \par}
\end{table}

\begin{figure*}[tbh]
\begin{centering}
\includegraphics[width=1.8\columnwidth]{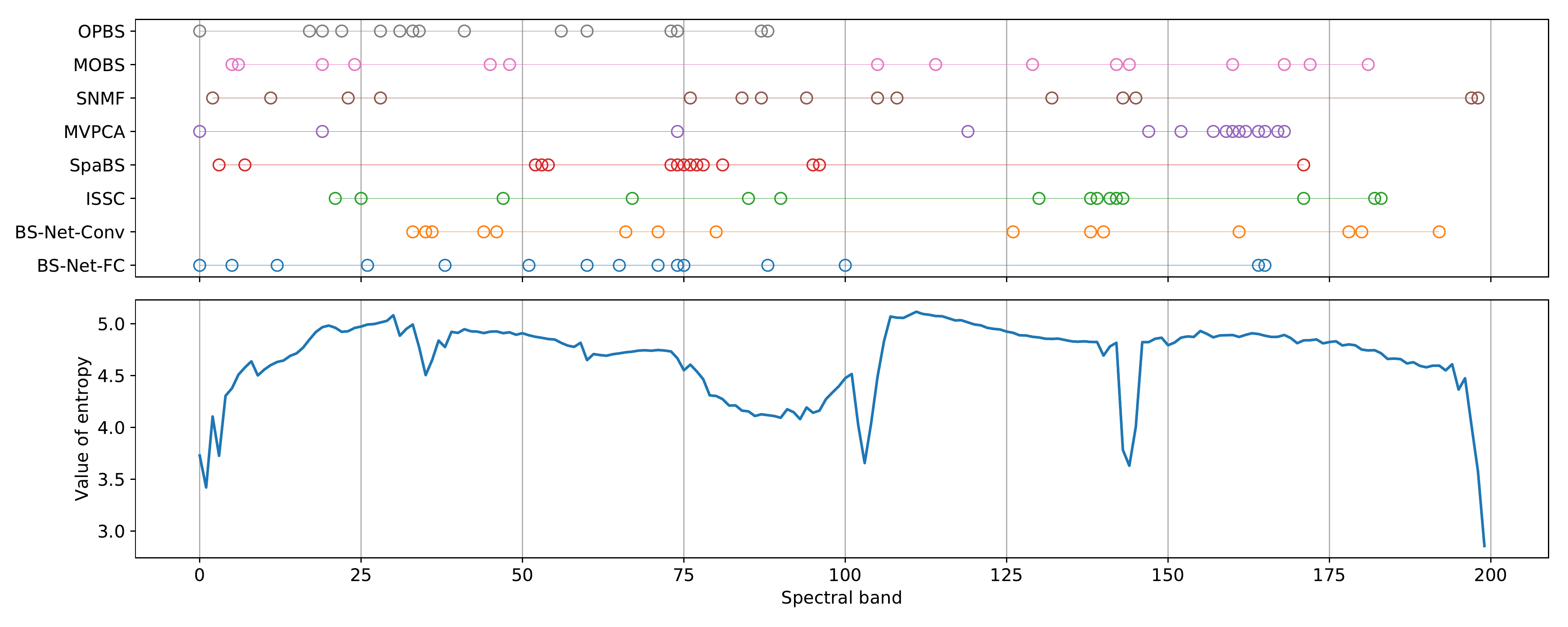}
\par\end{centering}
\caption{The best 15 bands of Indian Pines data set selected by different BS
methods (above) and the entropy value of each band (below). \label{fig:The-best-15band-distribution}}
\end{figure*}

We extensively analyze the selected bands in this section. Table \ref{tab:The-best-20-IndianP}
gives the best $15$ bands of Indian Pines data set selected by different
methods. To better show the band distribution, we indicate the locations
of these bands on the spectrum in Fig. \ref{fig:The-best-15band-distribution}
(above). Each row represents a BS  method with its corresponding locations
of the selected bands. As we can see, the results obtained by both
BS-Nets contain less continuous bands with a relatively uniform distribution.
Basing on the fact that the adjacent bands generally include higher
correlation and thus less redundancy containing among the band subsets
selected by both BS-Nets. Furthermore, we analyze these bands from
the perspective of the information entropy which we show in Fig. \ref{fig:The-best-15band-distribution}
(below). Those bands with extremely low entropy compared with their
adjacent bands can be regarded as noisy bands with little information,
i.e., $\left[104,105\right]$, $\left[144,145\right]$, $\left[198,199,200\right]$.
It can be seen that both BS-Nets avoid these regions with low entropy
since these noisy bands make no contribution to the spectral reconstruction.
Instead, both BS-Nets select informative bands from relatively smooth
regions with high entropy, thereby reducing the redundancy. 

\begin{figure}[tbh]
\begin{centering}
\includegraphics[width=0.8\columnwidth]{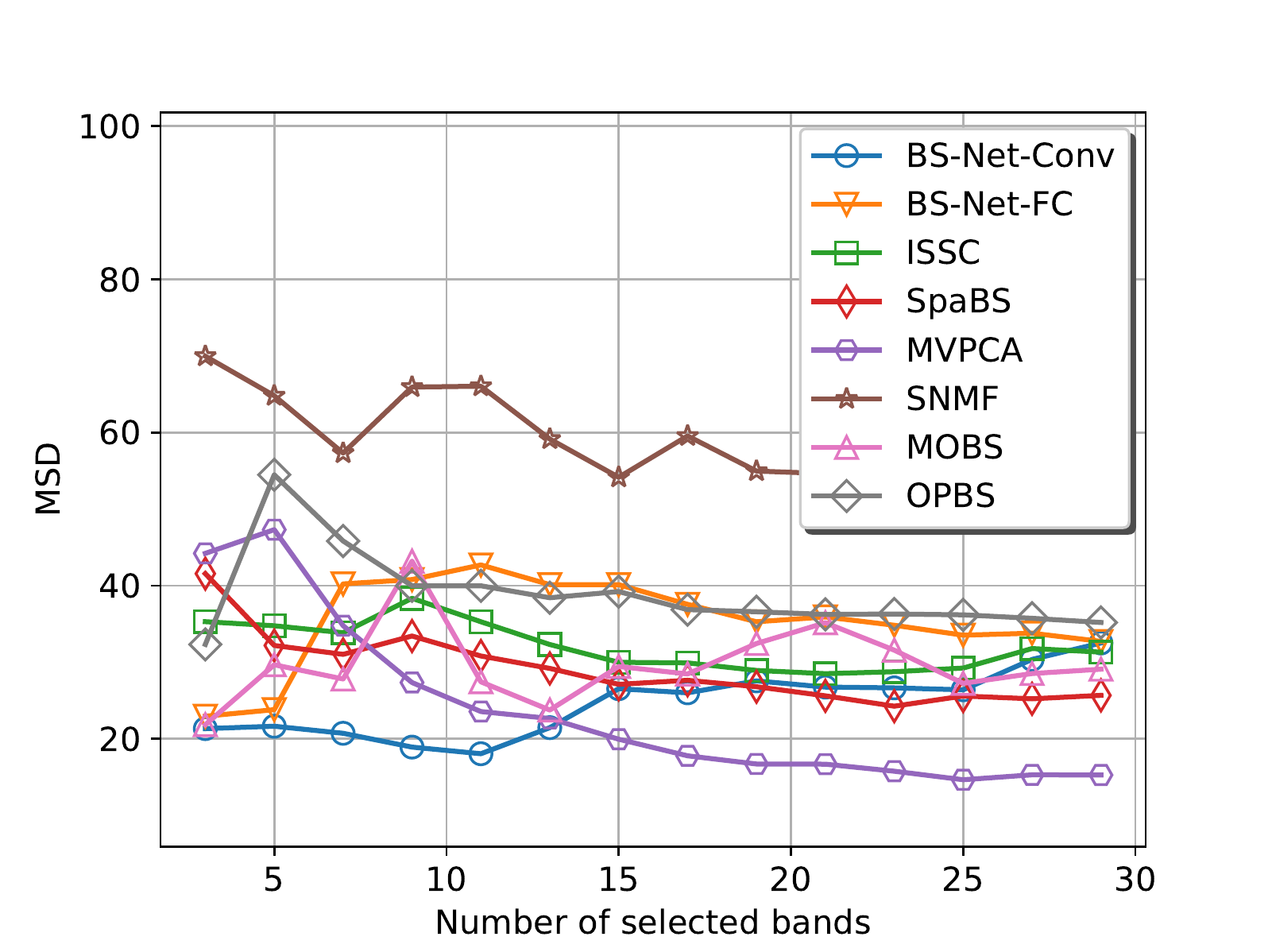}
\par\end{centering}
\caption{Mean Spectral Divergence values of different BS methods on Indian
Pines data set. \label{fig:Mean-Spectral-Divergence-IndianP}}
\end{figure}

In Fig. \ref{fig:Mean-Spectral-Divergence-IndianP}, we show the MSDs
of different BS methods under different band subset size. From Fig.
\ref{fig:Mean-Spectral-Divergence-IndianP}, BS-Net-FC has comparable
MSD values with OPBS but is better than most of the competitors, i.e.,
ISSC, SpaBS, MVPCA, and MPBS. Although BS-Net-Conv achieves the best
classification performance, it does not achieve the best MSD. As analyzed
in \cite{HSI_BS-EvoMultiObj-GongMG-TGRS-2016}, the reason for this
phenomenon is that the MSD will also increase if noisy bands are selected,
which can be concluded from Eq. \eqref{eq:msd}. For instance, the
MSDs of band subsets $\left[104,144\right]$ and $\left[104,25\right]$
are $106.64$ and $51.49$, respectively. It is obvious that $\left[104,144\right]$
shows much better MSD value than $\left[104,25\right]$, however,
$\left[104,144\right]$ contains two completely noisy bands which
makes less sense to the classification.

\subsection{Results on Pavia University Data Set }

\subsubsection{Data Set}

Pavia University data set was acquired by the ROSIS sensor during
a flight campaign over Pavia, northern Italy. This scene is a $103$
spectral bands $610\times610$ pixels image, but some of the samples
in the image contain no information and have to be discarded before
the analysis. The geometric resolution is 1.3 meters. The ground-truth
differentiates $9$ classes. 

\begin{figure}[tbh]
\subfloat[BS-Net-FC]{\begin{centering}
\includegraphics[width=0.48\columnwidth]{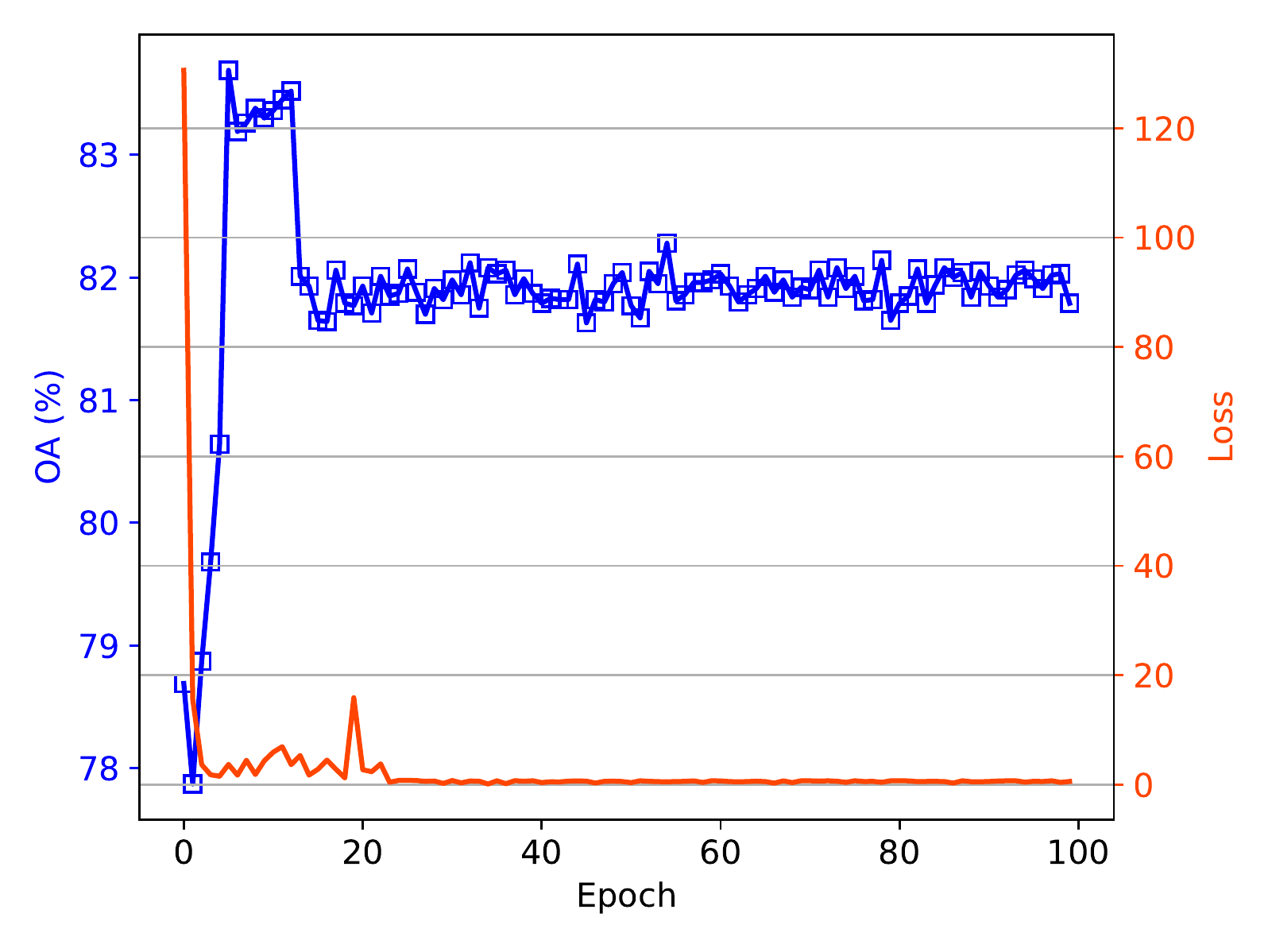}
\par\end{centering}
}\subfloat[BS-Net-Conv]{\begin{centering}
\includegraphics[width=0.48\columnwidth]{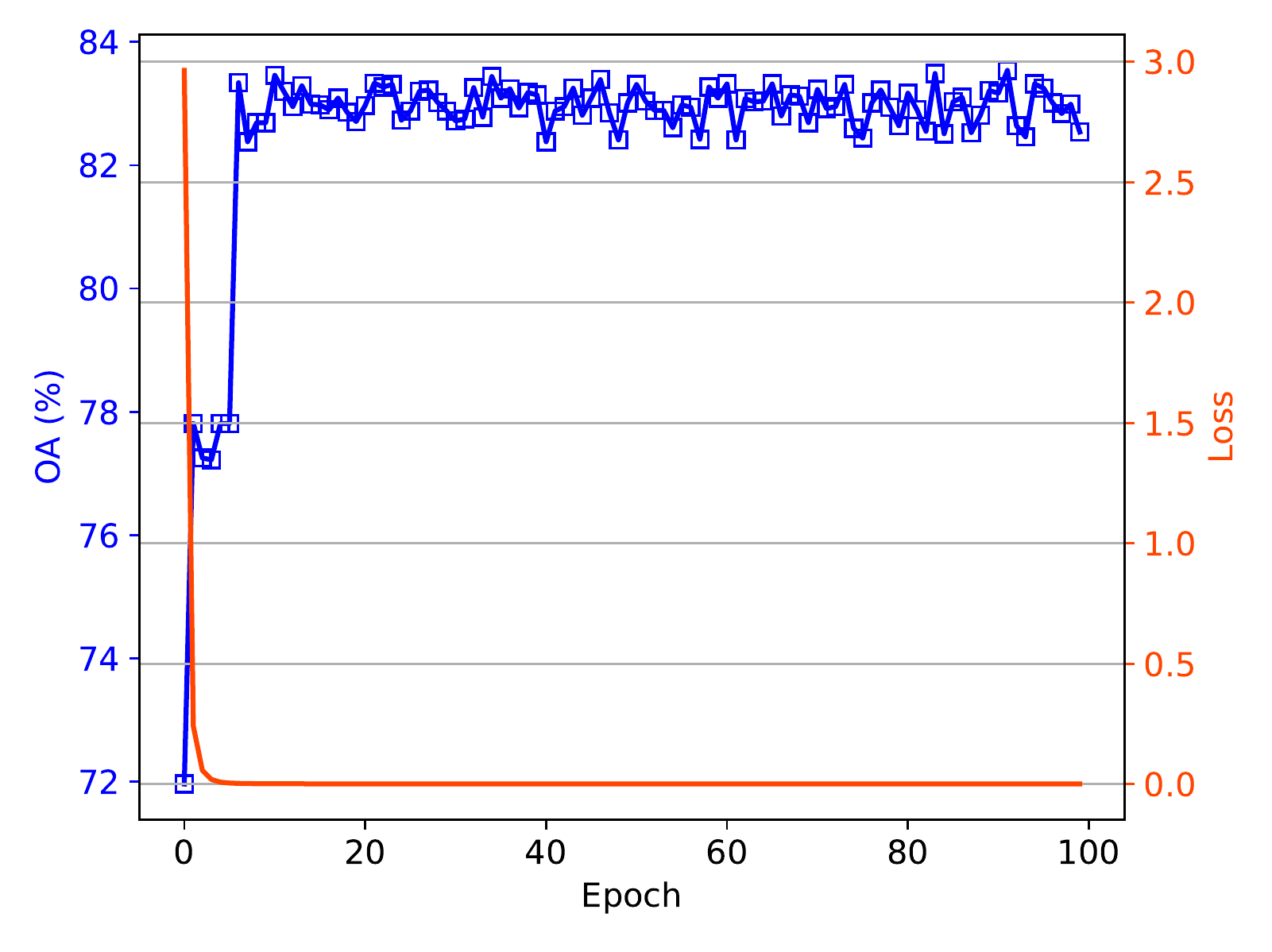}
\par\end{centering}
}

\subfloat[BS-Net-FC]{\begin{centering}
\includegraphics[width=0.48\columnwidth]{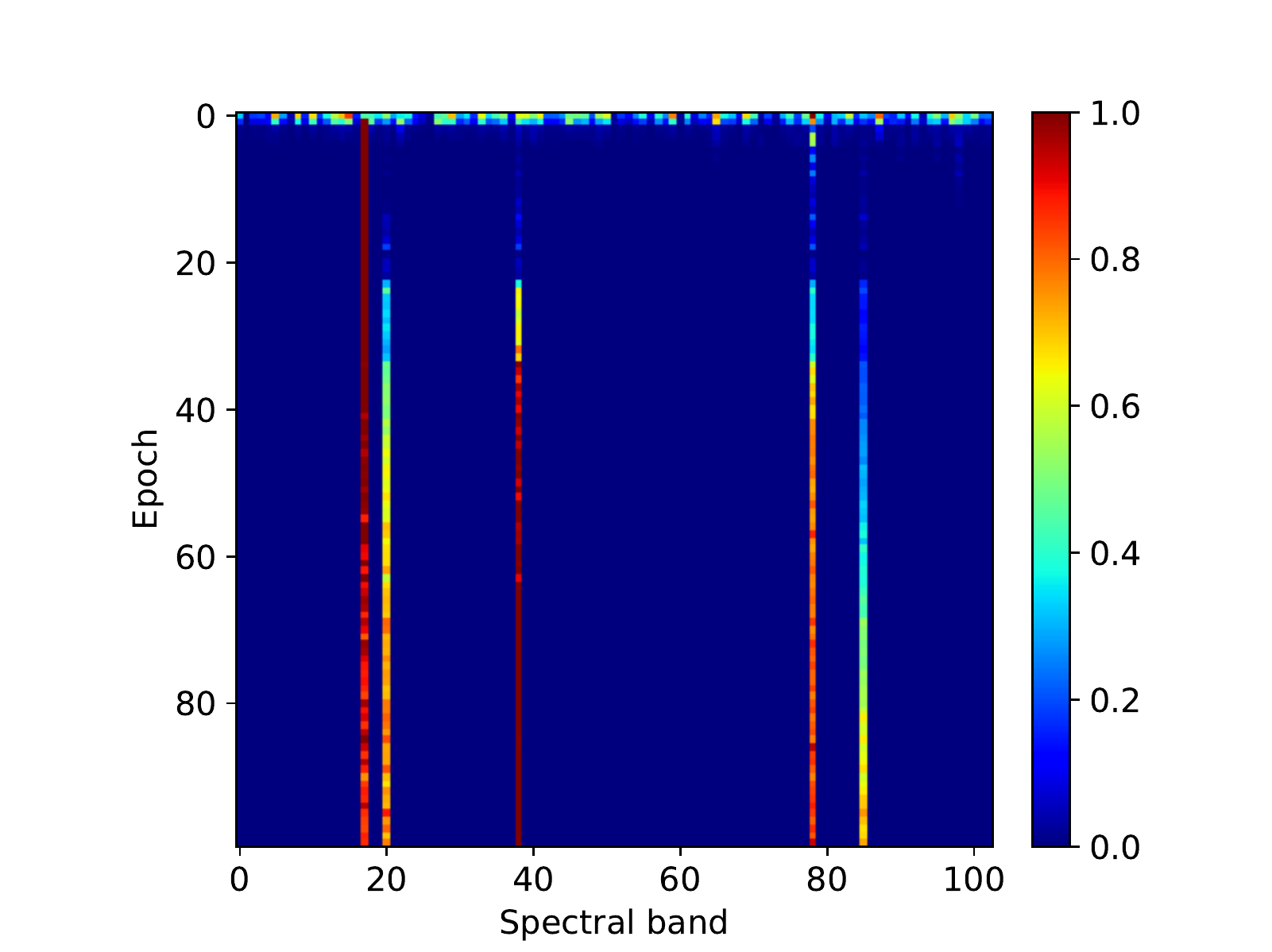}
\par\end{centering}
}\subfloat[BS-Net-Conv]{\begin{centering}
\includegraphics[width=0.48\columnwidth]{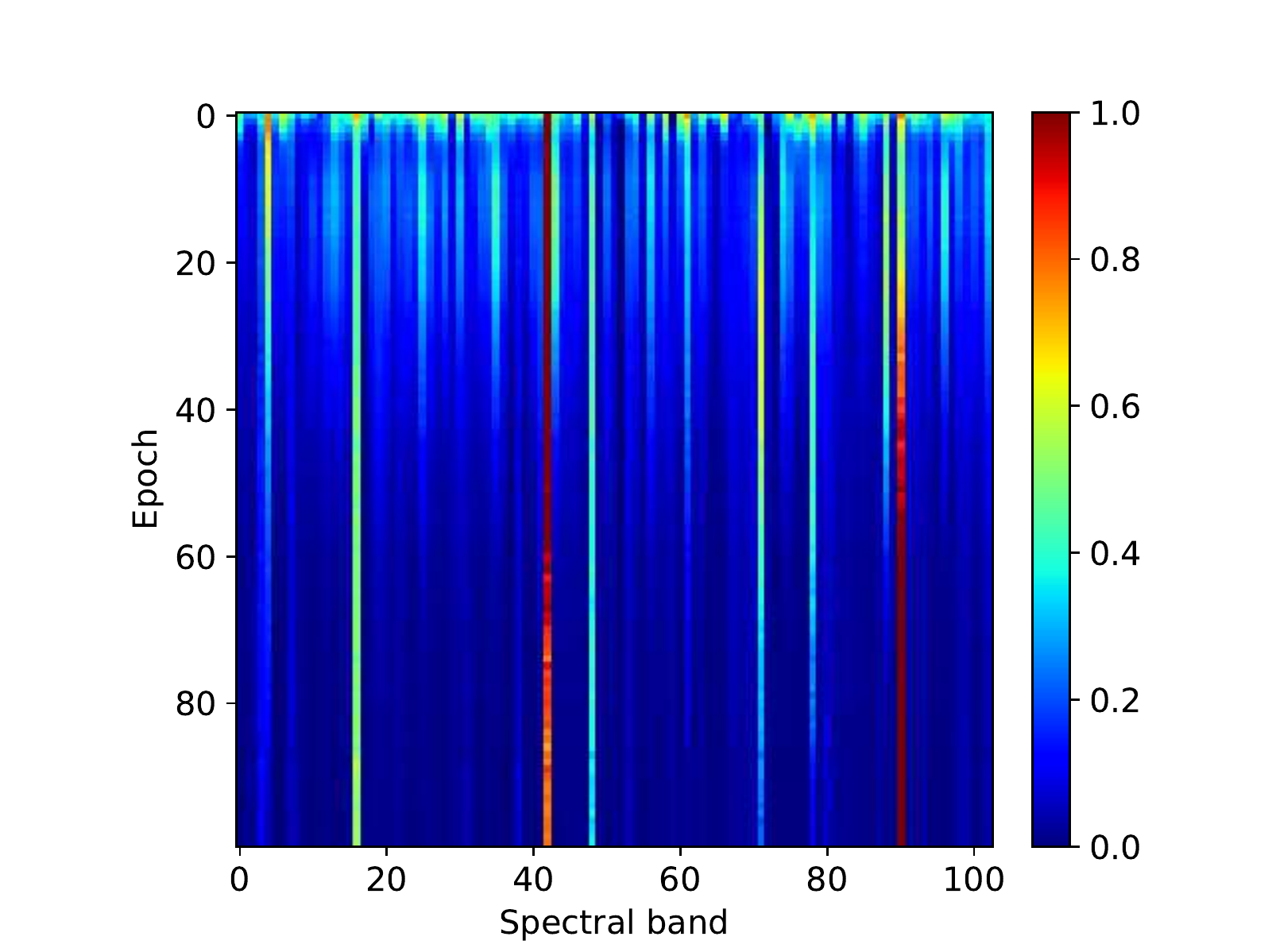}
\par\end{centering}
}

\caption{Analysis of the convergence of BS-Nets on Pavia University data set.
Visualization of loss versus accuracy under different iterations for
(a) BS-Net-FC and (b) BS-Net-Conv. Visualization of normalized average
band weights under varying iterations for (c) BS-Net-FC and (d) BS-Net-Conv.
\label{fig:Analysis-of-the-convergence-PaviaU}}
\end{figure}

\subsubsection{Analysis of Convergence of BS-Nets }

We show the convergence curves and the change trend of band weights
in Fig. \ref{fig:Analysis-of-the-convergence-PaviaU} (a)-(d). From
the results, the loss values tend to be zero after several iterations
showing that both BS-Nets converge well. Meanwhile, as the increase
of iteration, the classification accuracies of the best five bands
are increased from $78\%$ to $83\%$ and $72\%$  to $83\%$ for
BS-Net-FC and BS-Net-Conv, respectively. 

As shown in Fig. \ref{fig:Analysis-of-the-convergence-PaviaU} (c)-(d),
the average band weights become very sparse and easy to distinguish
when iteration increases, especially in BS-Net-FC. Finally, only a
few significant bands, which are useful to the spectral reconstruction,
are highlighted. For example, one can obviously determine that the
significant bands of BS-Net-FC are $\left[38,78,17,20,85\right]$
from Fig. \ref{fig:Analysis-of-the-convergence-PaviaU} (c). Similarly,
BS-Net-Conv's best bands are $\left[90,42,16,48,71\right]$. 

\begin{figure*}[tbh]
\begin{centering}
\subfloat[OA]{\includegraphics[width=0.65\columnwidth]{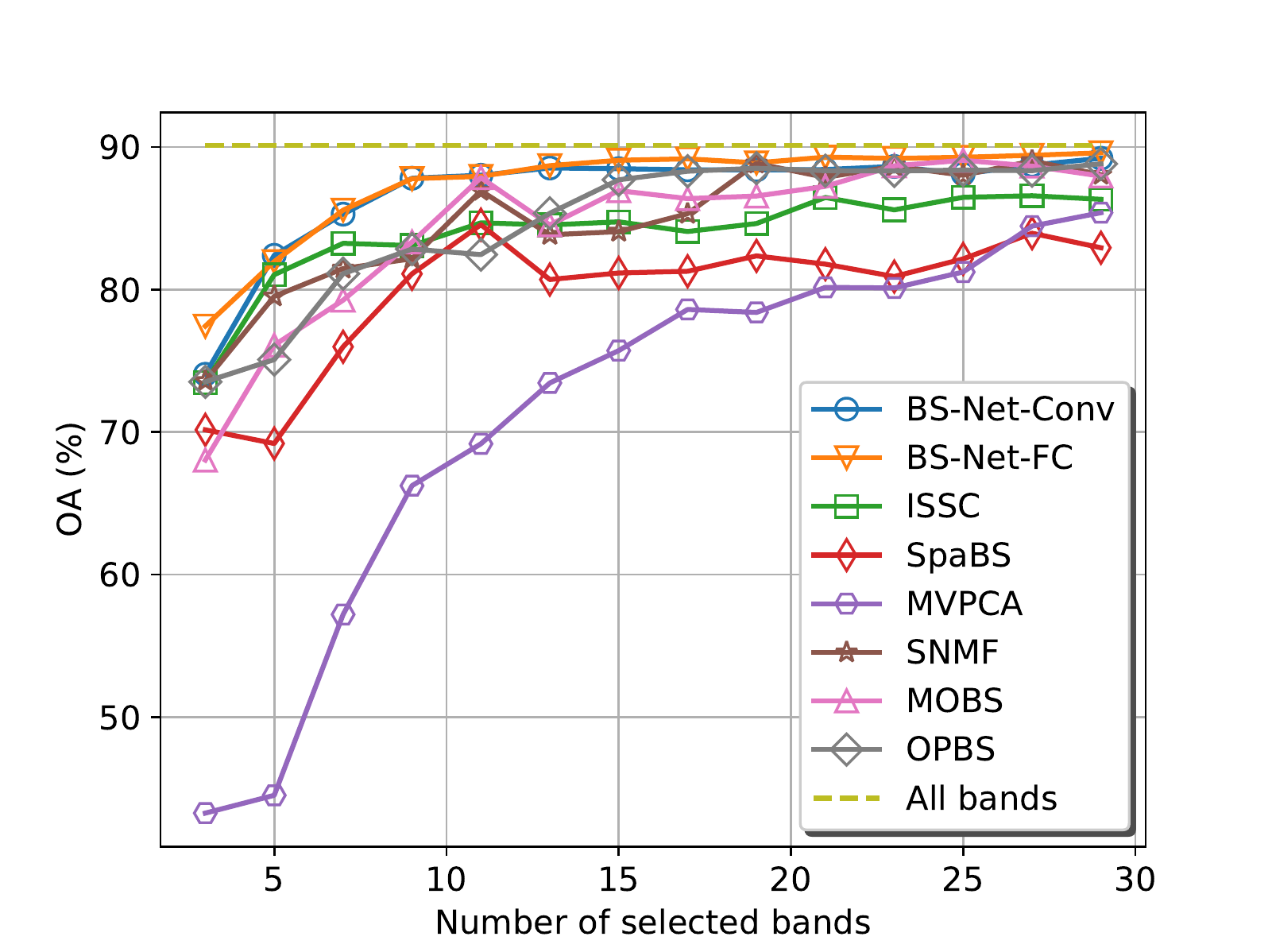}

}\subfloat[AA ]{\includegraphics[width=0.65\columnwidth]{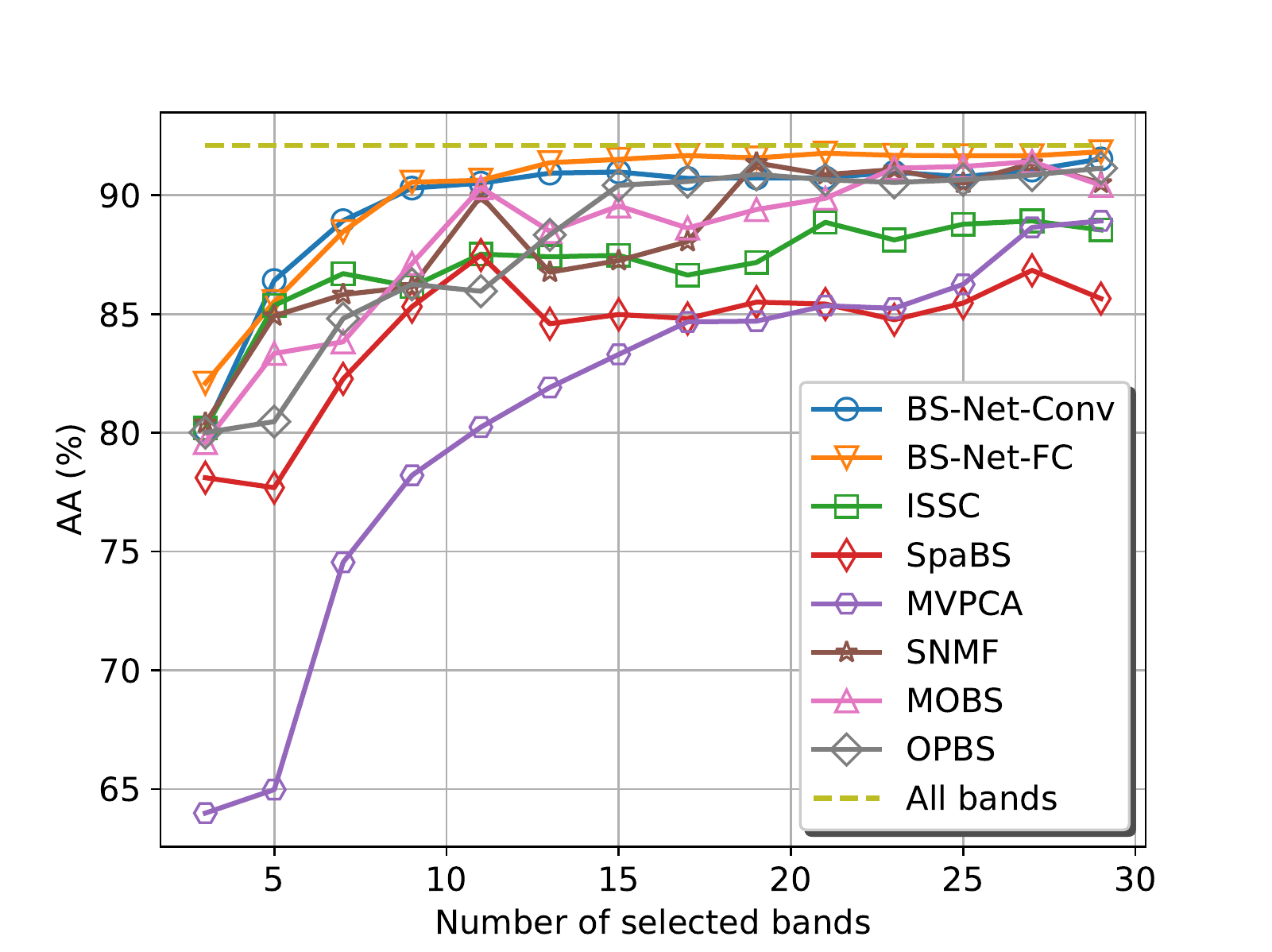}

}\subfloat[Kappa]{\includegraphics[width=0.65\columnwidth]{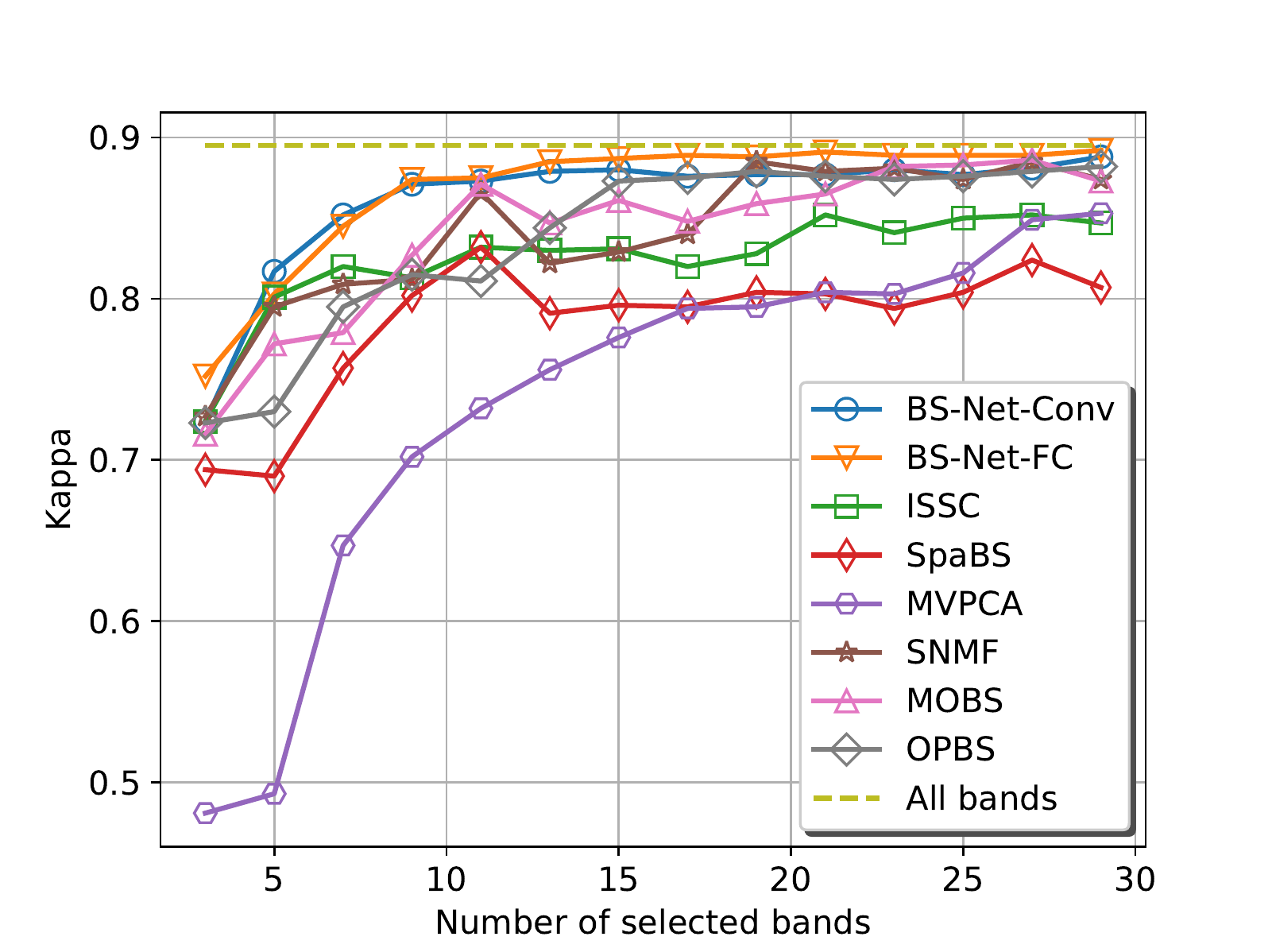}

}
\par\end{centering}
\caption{Performance comparison of different BS methods with different band
subset sizes on Pavia University data set. (a) OA; (b) AA; (c) Kappa.
\label{fig:Performance-comparison-of-paviaU}}
\end{figure*}

\begin{table*}[tbh]
\caption{Performance comparison of different methods using 15 bands on Pavia
University data set. \label{tab:Performance-comparison-of-PaviaU}}
\centering{}{\scriptsize{}}%
\begin{tabular}{|c|c|c||c|c|c|c|c|c|c|c|}
\hline 
{\scriptsize{}No.} & {\scriptsize{}\#Train} & {\scriptsize{}\#Test} & {\scriptsize{}ISSC} & {\scriptsize{}SpaBS} & {\scriptsize{}MVPCA} & {\scriptsize{}SNMF} & {\scriptsize{}MOBS} & {\scriptsize{}OPBS} & {\scriptsize{}BS-Net-FC} & {\scriptsize{}BS-Net-Conv}\tabularnewline
\hline 
\hline 
{\scriptsize{}1} & {\scriptsize{}373} & {\scriptsize{}6258} & {\scriptsize{}90.82$\pm$1.23} & {\scriptsize{}88.79$\pm$1.37} & {\scriptsize{}86.91$\pm$1.18} & {\scriptsize{}90.21$\pm$1.11} & {\scriptsize{}91.17$\pm$1.33} & {\scriptsize{}90.72$\pm$1.32} & \textbf{\scriptsize{}91.53$\pm$0.80 } & {\scriptsize{}91.15$\pm$0.88}\tabularnewline
\hline 
{\scriptsize{}2} & {\scriptsize{}920} & {\scriptsize{}17729} & {\scriptsize{}95.07$\pm$0.36} & {\scriptsize{}95.88$\pm$0.63} & {\scriptsize{}94.49$\pm$0.41} & {\scriptsize{}96.36$\pm$0.37} & {\scriptsize{}95.68$\pm$0.45} & {\scriptsize{}95.61$\pm$0.55} & \textbf{\scriptsize{}96.64$\pm$0.35} & {\scriptsize{}95.72$\pm$0.37}\tabularnewline
\hline 
{\scriptsize{}3} & {\scriptsize{}112} & {\scriptsize{}1987} & {\scriptsize{}69.65$\pm$3.30} & {\scriptsize{}62.98$\pm$3.05} & {\scriptsize{}57.06$\pm$4.59} & {\scriptsize{}72.77$\pm$3.00} & {\scriptsize{}70.97$\pm$2.88} & \textbf{\scriptsize{}73.60$\pm$3.13} & {\scriptsize{}72.30$\pm$2.94} & {\scriptsize{}71.20$\pm$3.01}\tabularnewline
\hline 
{\scriptsize{}4} & {\scriptsize{}134} & {\scriptsize{}2930} & {\scriptsize{}89.29$\pm$2.07} & {\scriptsize{}85.10$\pm$1.66} & {\scriptsize{}79.86$\pm$2.38} & {\scriptsize{}89.37$\pm$1.49} & {\scriptsize{}89.00$\pm$1.68} & {\scriptsize{}89.72$\pm$1.97} & \textbf{\scriptsize{}91.47$\pm$1.78} & {\scriptsize{}90.60$\pm$1.78}\tabularnewline
\hline 
{\scriptsize{}5} & {\scriptsize{}69} & {\scriptsize{}1276} & {\scriptsize{}98.89$\pm$0.35} & {\scriptsize{}99.10$\pm$0.27} & {\scriptsize{}98.39$\pm$0.67} & {\scriptsize{}99.18$\pm$0.39} & \textbf{\scriptsize{}99.44$\pm$0.20} & {\scriptsize{}99.18$\pm$0.50} & {\scriptsize{}99.31$\pm$0.30 } & {\scriptsize{}99.36$\pm$0.28}\tabularnewline
\hline 
{\scriptsize{}6} & {\scriptsize{}234} & {\scriptsize{}4795} & {\scriptsize{}74.49$\pm$1.64} & {\scriptsize{}52.0$\pm$3.01} & {\scriptsize{}72.76$\pm$1.87} & {\scriptsize{}85.53$\pm$1.44} & {\scriptsize{}78.74$\pm$1.69} & {\scriptsize{}84.64$\pm$1.49} & \textbf{\scriptsize{}86.55$\pm$1.65} & {\scriptsize{}84.31$\pm$1.54}\tabularnewline
\hline 
{\scriptsize{}7} & {\scriptsize{}71} & {\scriptsize{}1259} & {\scriptsize{}78.89$\pm$4.01} & {\scriptsize{}70.63$\pm$6.14} & {\scriptsize{}55.44$\pm$6.72} & {\scriptsize{}73.47$\pm$4.23} & {\scriptsize{}76.62$\pm$4.08} & {\scriptsize{}78.90$\pm$3.16} & \textbf{\scriptsize{}80.66$\pm$2.80} & {\scriptsize{}80.01$\pm$3.21}\tabularnewline
\hline 
{\scriptsize{}8} & {\scriptsize{}179} & {\scriptsize{}3503} & {\scriptsize{}81.28$\pm$1.83} & {\scriptsize{}81.87$\pm$2.45} & {\scriptsize{}76.49$\pm$3.61} & {\scriptsize{}84.11$\pm$2.55} & {\scriptsize{}83.56$\pm$2.18} & {\scriptsize{}82.92$\pm$1.60} & \textbf{\scriptsize{}85.16$\pm$2.30} & {\scriptsize{}83.37$\pm$1.96}\tabularnewline
\hline 
{\scriptsize{}9} & {\scriptsize{}46} & {\scriptsize{}901} & {\scriptsize{}99.77$\pm$0.16} & {\scriptsize{}99.71$\pm$0.14} & {\scriptsize{}99.96$\pm$0.15} & {\scriptsize{} 99.81$\pm$0.17} & {\scriptsize{}99.97$\pm$0.07} & {\scriptsize{}99.81$\pm$0.14} & \textbf{\scriptsize{} 99.97$\pm$0.05} & {\scriptsize{} 99.92$\pm$0.08}\tabularnewline
\hline 
\hline 
\multicolumn{3}{|c||}{{\scriptsize{}OA (\%)}} & {\scriptsize{}86.46$\pm$0.54} & {\scriptsize{}81.78$\pm$0.89} & {\scriptsize{}80.15$\pm$0.68} & {\scriptsize{}87.87$\pm$0.58} & {\scriptsize{}87.24$\pm$0.59} & {\scriptsize{}88.34$\pm$0.62} & \textbf{\scriptsize{}89.29$\pm$0.47} & {\scriptsize{}88.40$\pm$0.51}\tabularnewline
\hline 
\multicolumn{3}{|c||}{{\scriptsize{}AA (\%)}} & {\scriptsize{}88.86$\pm$0.28} & {\scriptsize{}85.42$\pm$0.33} & {\scriptsize{}85.35$\pm$0.22} & {\scriptsize{}90.76$\pm$0.26} & {\scriptsize{}89.87$\pm$0.27} & {\scriptsize{}90.65$\pm$0.37} & \textbf{\scriptsize{}91.77$\pm$0.30} & {\scriptsize{}90.87$\pm$0.28}\tabularnewline
\hline 
\multicolumn{3}{|c||}{{\scriptsize{}Kappa}} & {\scriptsize{}0.852$\pm$0.004} & {\scriptsize{}0.803$\pm$0.005} & {\scriptsize{}0.804$\pm$0.003} & {\scriptsize{}0.877$\pm$0.003} & {\scriptsize{}0.865$\pm$0.004} & {\scriptsize{}0.876$\pm$0.005} & \textbf{\scriptsize{}0.891$\pm$0.004} & {\scriptsize{}0.879$\pm$0.004}\tabularnewline
\hline 
\end{tabular}{\scriptsize \par}
\end{table*}

\subsubsection{Performance Comparison}

In this experiment, we perform different BS methods to select different
sizes of band subsets ranging from 3 to 30. We show the obtained  OAs,
AAs, and Kappas in Fig. \ref{fig:Performance-comparison-of-paviaU}
(a)-(c). When band subset size is less than $17$, both BS-Nets have
similar classification performance and are significantly superior
to the other BS methods in terms of all the three indices. When band
subset size is larger than $17$, BS-Net-FC achieves the best classification
performance, while BS-Net-Conv achieves comparable performance with
SNMF and OPBS. Using only $25$ bands, both BS-Nets show comparable
performance with all bands and no obvious Hughes phenomenon. 

In Table \ref{tab:Performance-comparison-of-PaviaU}, we compare the
detailed classification performance by setting the band subset size
to $19$. From Table \ref{tab:Performance-comparison-of-PaviaU},
both BS-Nets achieve the best OA, AA, and Kappa by comparing with
the other competitors. In addition, BS-Net-FC wins in $7$ classes
in terms of classes accuracy. For the two failed classes, i.e., No.
3 and No. 5 classes, their classification accuracy are superior to
most of the other competitors. Especially on No. 3 class, which is
relatively difficult to classify, but both BS-Nets also achieve much
more accurate results than ISSC, SpaBS, and MVPCA.

\subsubsection{Analysis of the Selected Bands}

\begin{table}[tbh]
\caption{The best 15 bands of Pavia University data set selected by different
BS methods. \label{tab:The-best-15-PaviaU}}
\centering{}{\scriptsize{}}%
\begin{tabular}{|c|c|}
\hline 
{\scriptsize{}Methods} & {\scriptsize{}Selected Bands}\tabularnewline
\hline 
\hline 
{\scriptsize{}BS-Net-FC} & {\scriptsize{}{[}38, 78, 17, 20, 85, 98, 65, 81, 79, 90, 95, 74, 66,
62, 92{]}}\tabularnewline
\hline 
{\scriptsize{}BS-Net-Conv} & {\scriptsize{}{[}90, 42, 16, 48, 71, 3, 78, 38, 80, 53, 7, 31, 4,
99, 98{]}}\tabularnewline
\hline 
{\scriptsize{}ISSC} & {\scriptsize{}{[}51, 76, 7, 64, 31, 8, 0, 24, 40, 30, 5, 3, 6, 27,
2{]}}\tabularnewline
\hline 
{\scriptsize{}SpaBS} & {\scriptsize{}{[}50, 48, 16, 22, 4, 102, 21, 25, 23, 47, 24, 20, 31,
26, 42{]}}\tabularnewline
\hline 
{\scriptsize{}MVPCA} & {\scriptsize{}{[}48, 22, 51, 16, 52, 21, 65, 17, 20, 53, 18, 54, 19,
55, 76{]}}\tabularnewline
\hline 
{\scriptsize{}SNMF} & {\scriptsize{}{[}92, 53, 43, 66, 22, 89, 82, 30, 51, 5, 83, 77, 80,
2, 48{]}}\tabularnewline
\hline 
{\scriptsize{}MOBS} & {\scriptsize{}{[}4, 15, 23, 25, 33, 35, 42, 53, 58, 61, 62, 64, 67,
73, 101{]}}\tabularnewline
\hline 
{\scriptsize{}OPBS} & {\scriptsize{}{[}90, 62, 14, 0, 2, 72, 102, 4, 33, 1, 6, 84, 45, 82,
8{]}}\tabularnewline
\hline 
\end{tabular}{\scriptsize \par}
\end{table}

\begin{figure*}[tbh]
\begin{centering}
\includegraphics[width=2\columnwidth]{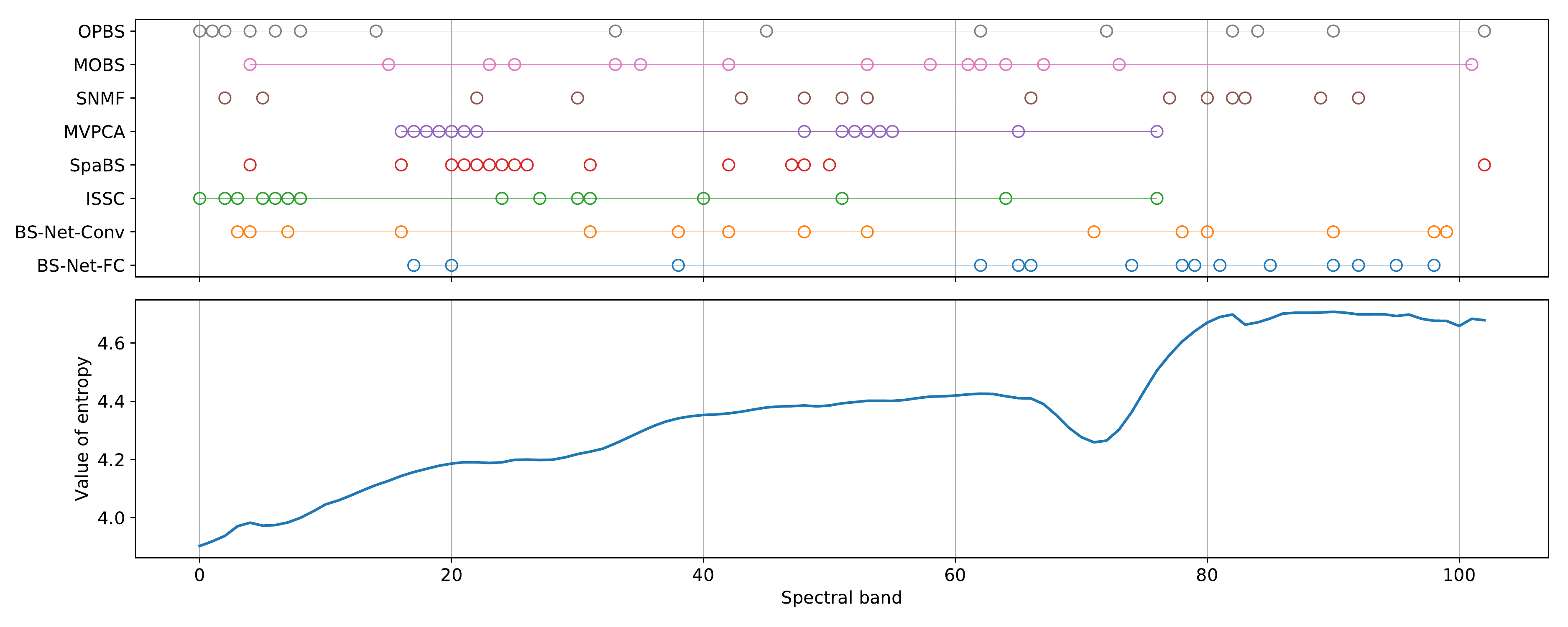}
\par\end{centering}
\caption{The best 15 bands of Pavia University data set selected by different
BS methods (above) and the entropy value of each band (below). \label{fig:The-best-15-index-PaviaU}}
\end{figure*}

\begin{figure}[tbh]
\begin{centering}
\includegraphics[width=0.8\columnwidth]{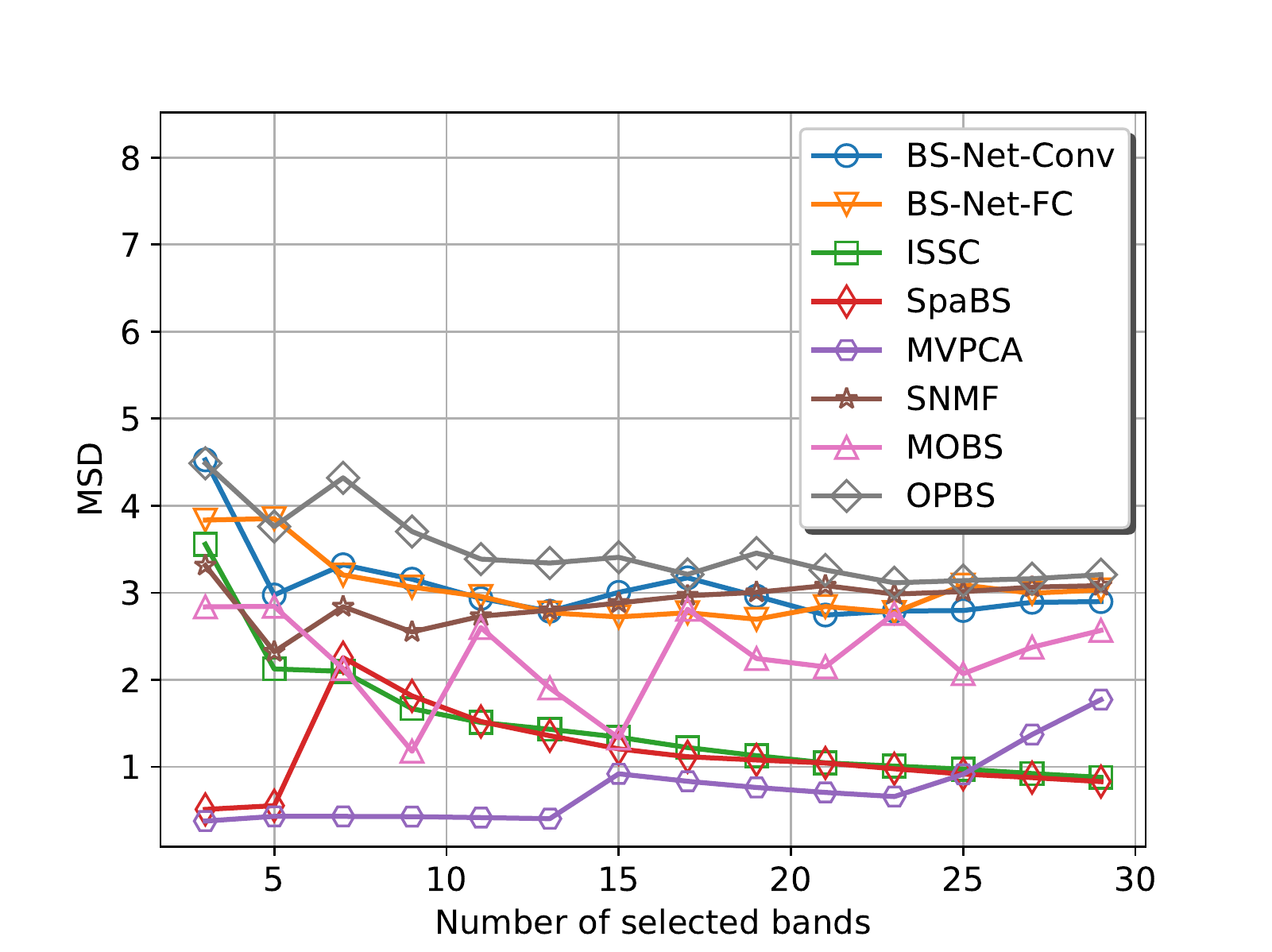}
\par\end{centering}
\caption{Mean Spectral Divergence values of different BS methods on Pavia University
data set.\label{fig:Mean-Spectral-Divergence-PaviaU}}
\end{figure}

The best $15$ bands selected by different BS methods are listed in
Table \ref{tab:The-best-15-PaviaU}. Fig. \ref{fig:The-best-15-index-PaviaU}
shows the corresponding band indices and the entropy value  of each
spectral band. The entropy curve of this data set is relatively smooth
without rapidly decreasing regions and moreover increases with spectral
bands. Compared with the competitors, the band subsets selected by
both BS-Nets contain fewer continuous bands and are more concentrated
around the positions of larger entropy. In contrast, MVPCA, SpaBS,
and ISSC include more adjacent bands at the region with low entropy.
As a result, these methods have worse classification performance than
BS-Nets. The MSD values of the different BS methods are given in Fig.
\ref{fig:Mean-Spectral-Divergence-PaviaU}. It can be seen that both
BS-Nets are comparable with OPBS and are better than ISSC, SpaBS,
MVPCA, and MOBS, showing that the selected band subsets contain less
redundant bands.

\subsection{Results on Salinas Data Set }

\subsubsection{Data set}

This scene was collected by the 224-band AVIRIS sensor over Salinas
Valley, California, and is characterized by high spatial resolution
($3.7$-meter pixels). The area covered comprises $512\times217$
samples. As with Indian Pines scene, we discarded the 20 water absorption
bands, in this case, bands: $[108-112]$, $[154-167]$, $224$. It
includes vegetables, bare soils, and vineyard fields. Salinas ground-truth
contains 16 classes. 

\subsubsection{Analysis of Convergence of BS-Nets}

\begin{figure}[tbh]
\subfloat[BS-Net-FC]{\begin{centering}
\includegraphics[width=0.48\columnwidth]{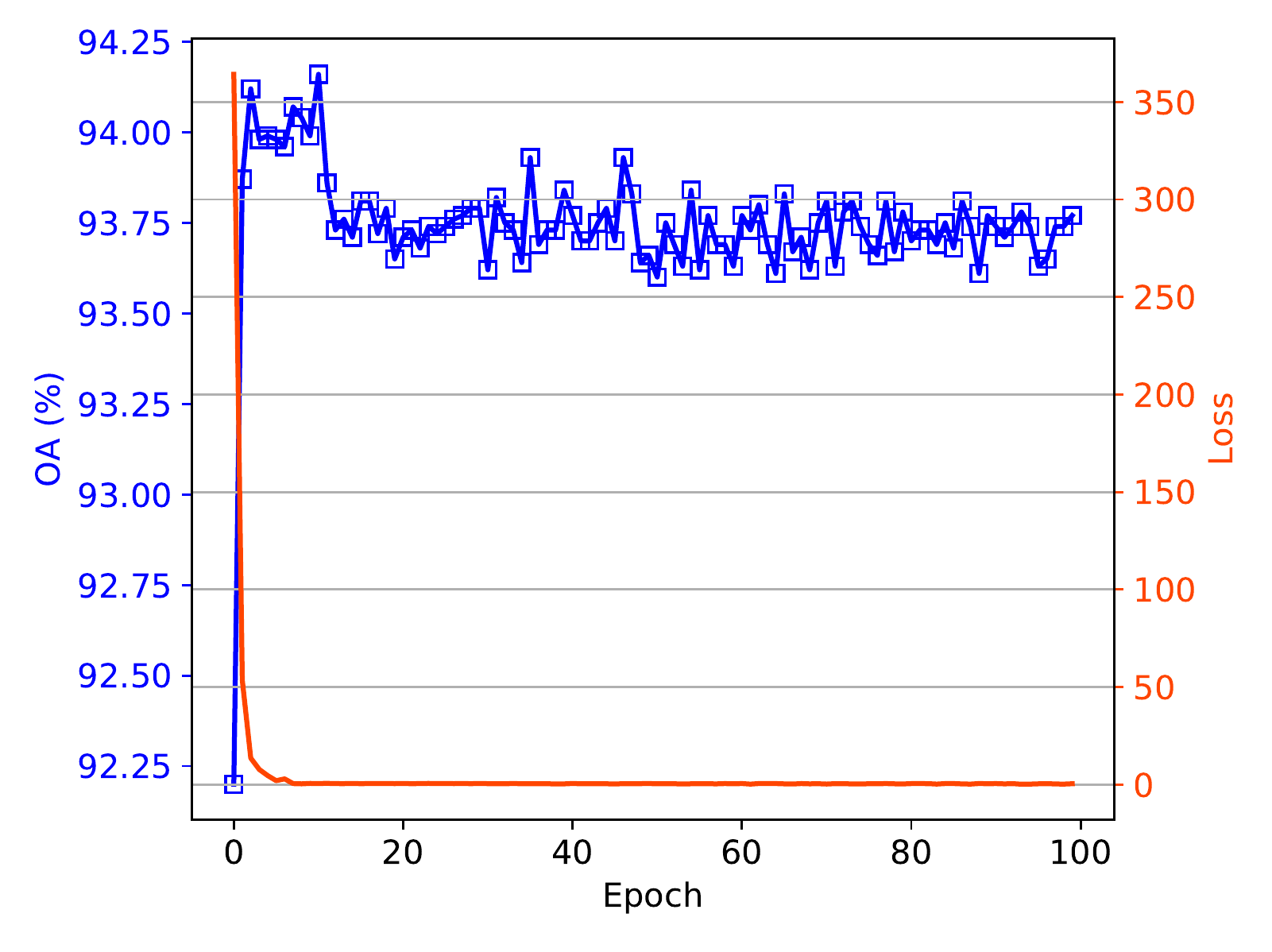}
\par\end{centering}
}\subfloat[BS-Net-Conv]{\includegraphics[width=0.48\columnwidth]{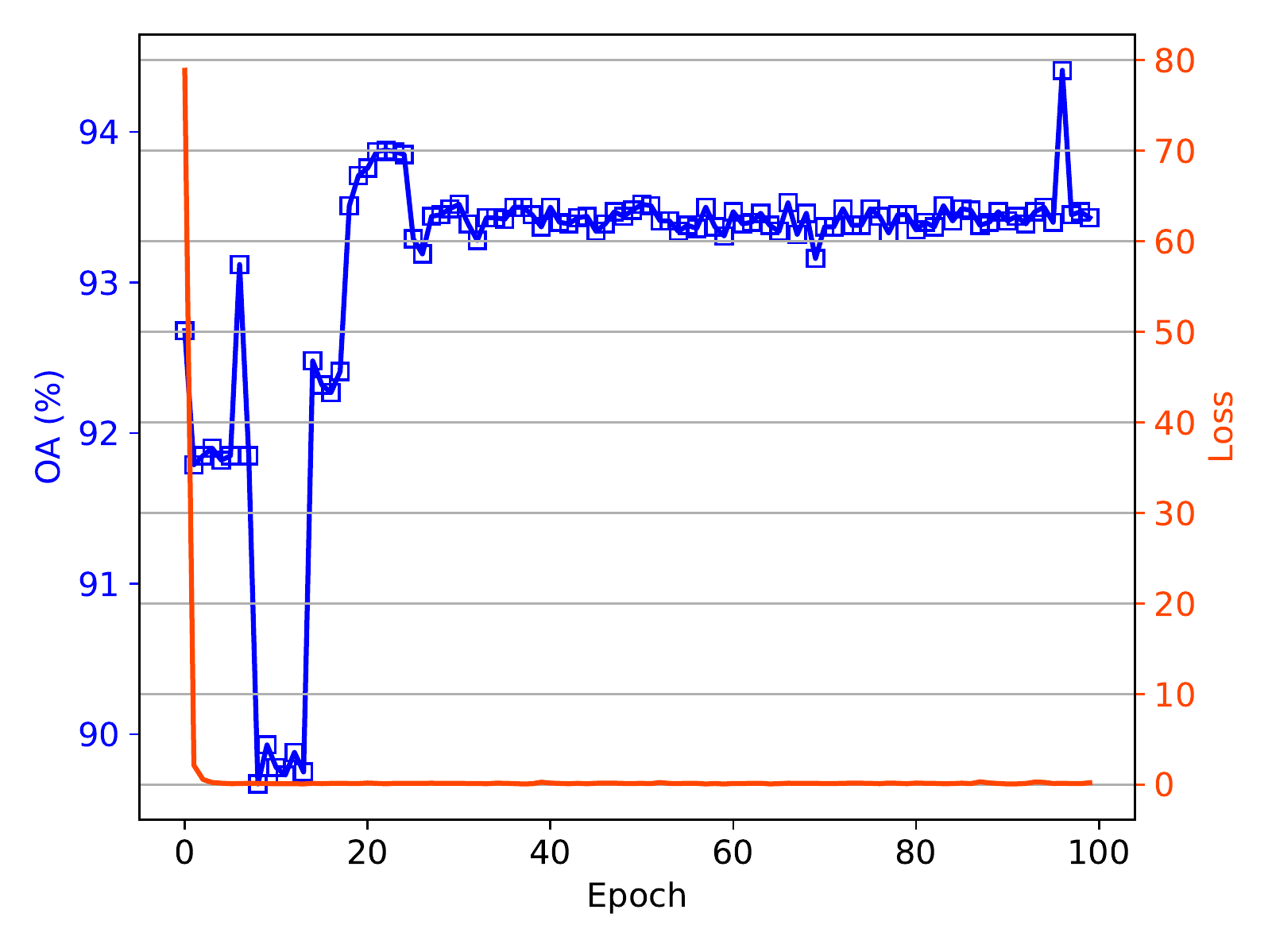}

}

\subfloat[BS-Net-FC]{\begin{centering}
\includegraphics[width=0.48\columnwidth]{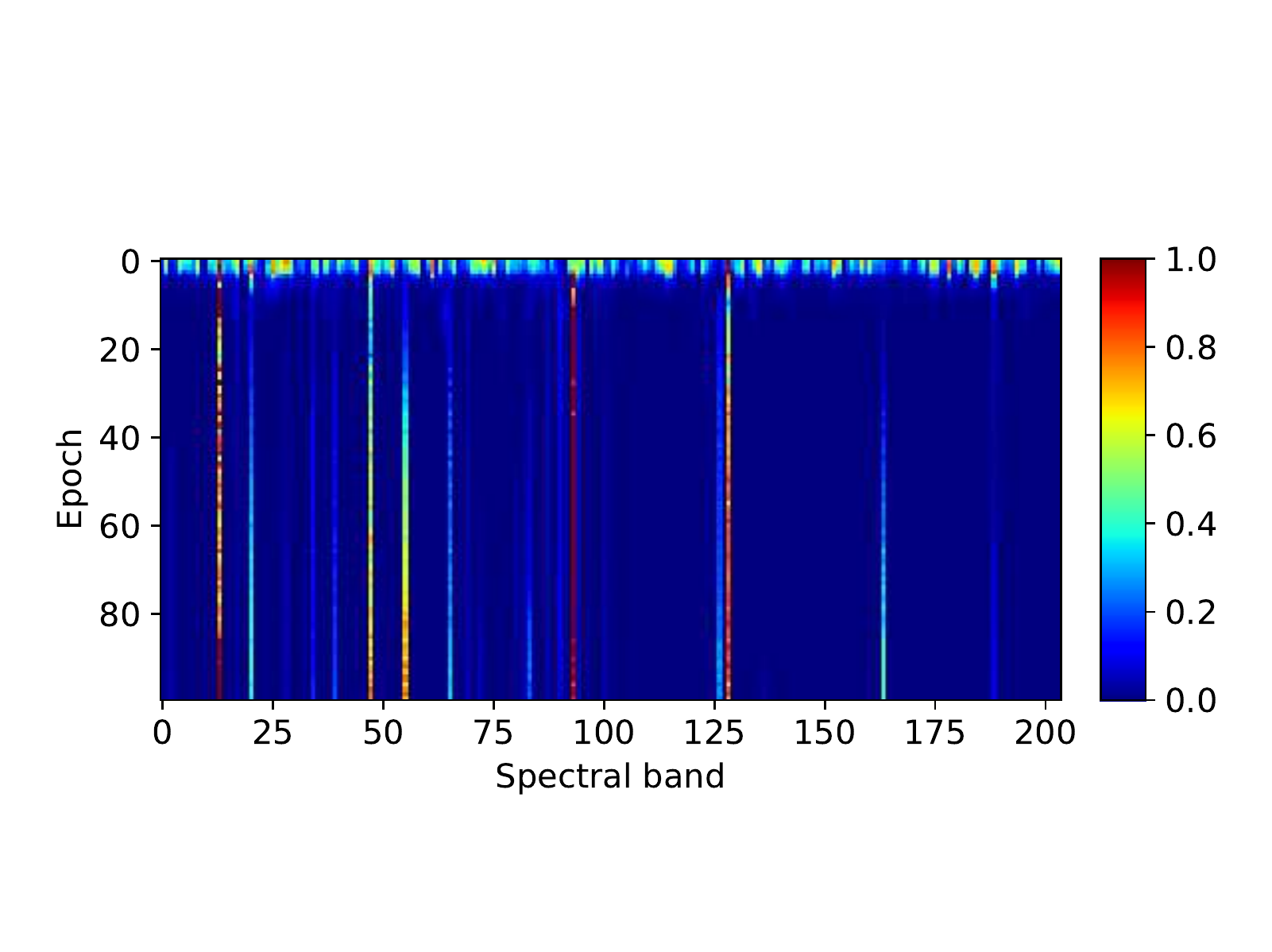}
\par\end{centering}
}\subfloat[BS-Net-Conv]{\includegraphics[width=0.48\columnwidth]{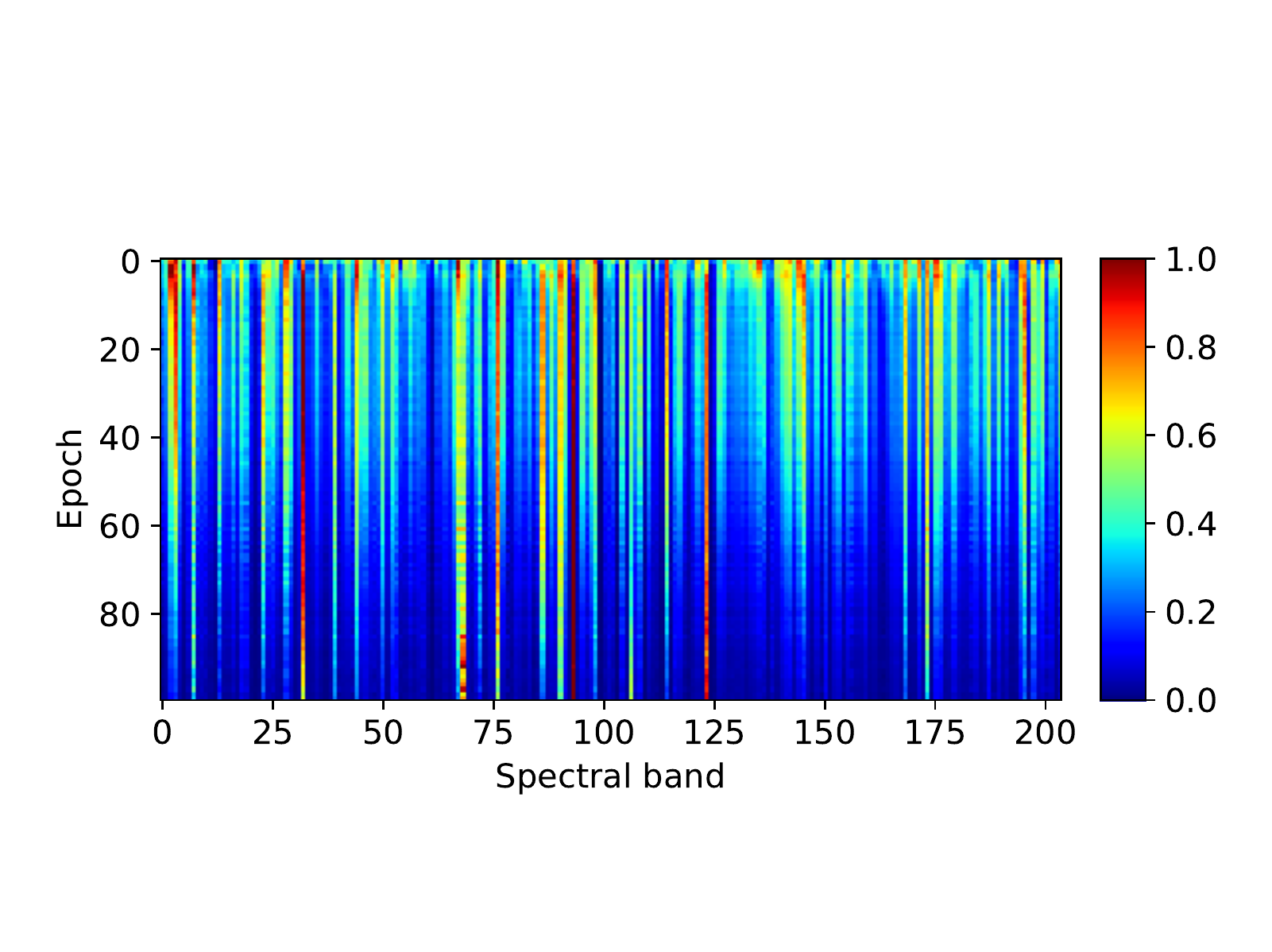}

}

\caption{Analysis of the convergence of BS-Nets on Salinas data set. Loss versus
accuracy under different iterations for (a) BS-Net-FC and (b) BS-Net-Conv.
Visualization of normalized average band weights under varying iterations
for (c) BS-Net-FC and (d) BS-Net-Conv. \label{fig:Analysis-of-band-Salinas}}
\end{figure}

Fig. \ref{fig:Analysis-of-band-Salinas} (a)-(b) show the convergence
curves of BS-Nets on Salinas data set. Training about $20$ iterations,
BS-Nets' loss and accuracy have tended to be convergent. The OA of
using $5$ bands are increased from $92\%$ to $94\%$ and from $85\%$
to $94\%$ for BS-Net-FC and BS-Net-Conv, respectively. In Fig. \ref{fig:Analysis-of-band-Salinas}
(c)-(d), we show the means of band weights under different iterations.
Similar to Indian Pines and Pavia University data sets, the learned
band weights become sparse with the increase of iterations.  

\subsubsection{Performance Comparison}

\begin{figure*}[tbh]
\begin{centering}
\subfloat[OA]{\includegraphics[width=0.65\columnwidth]{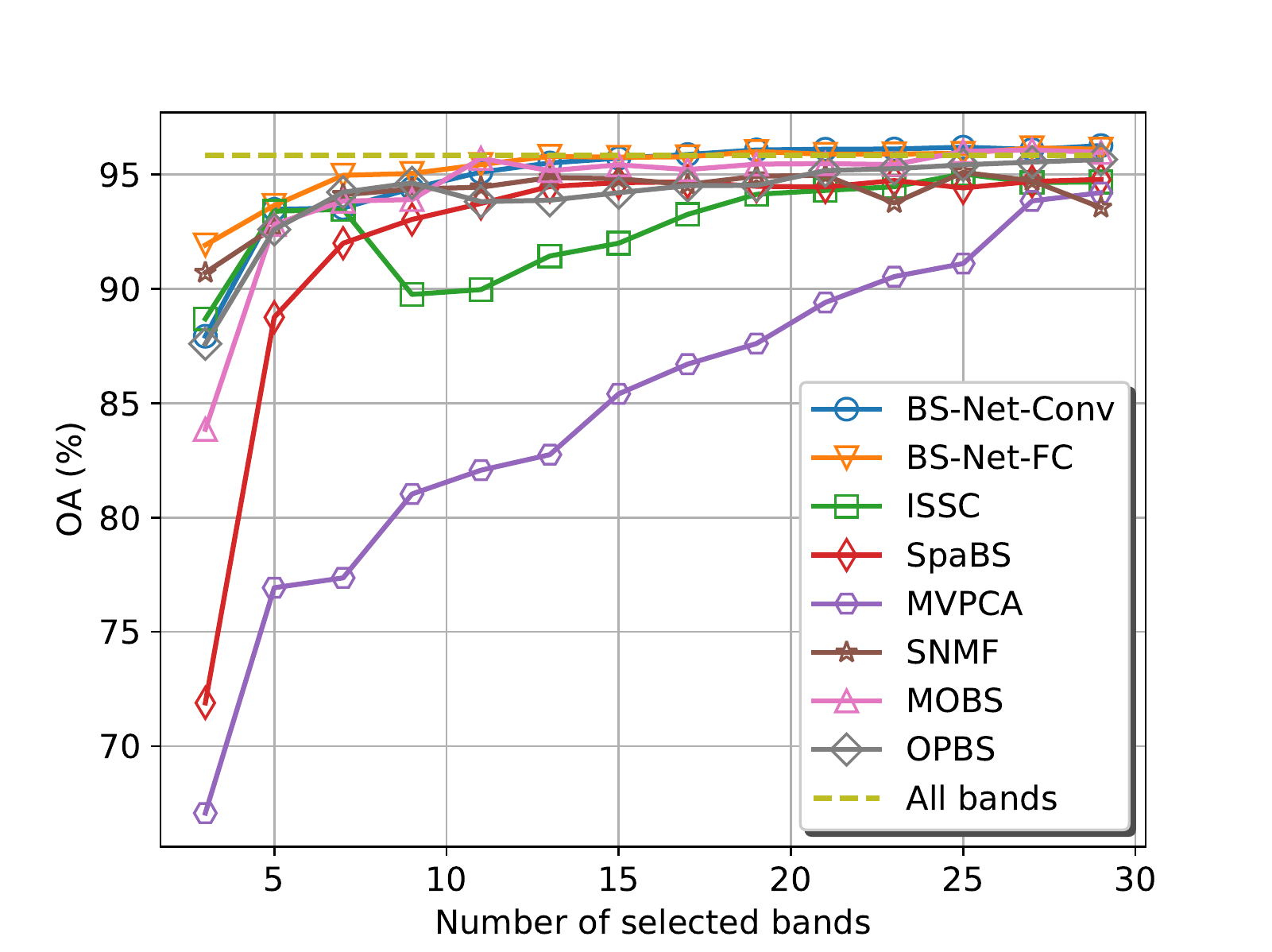}

}\subfloat[AA ]{\includegraphics[width=0.65\columnwidth]{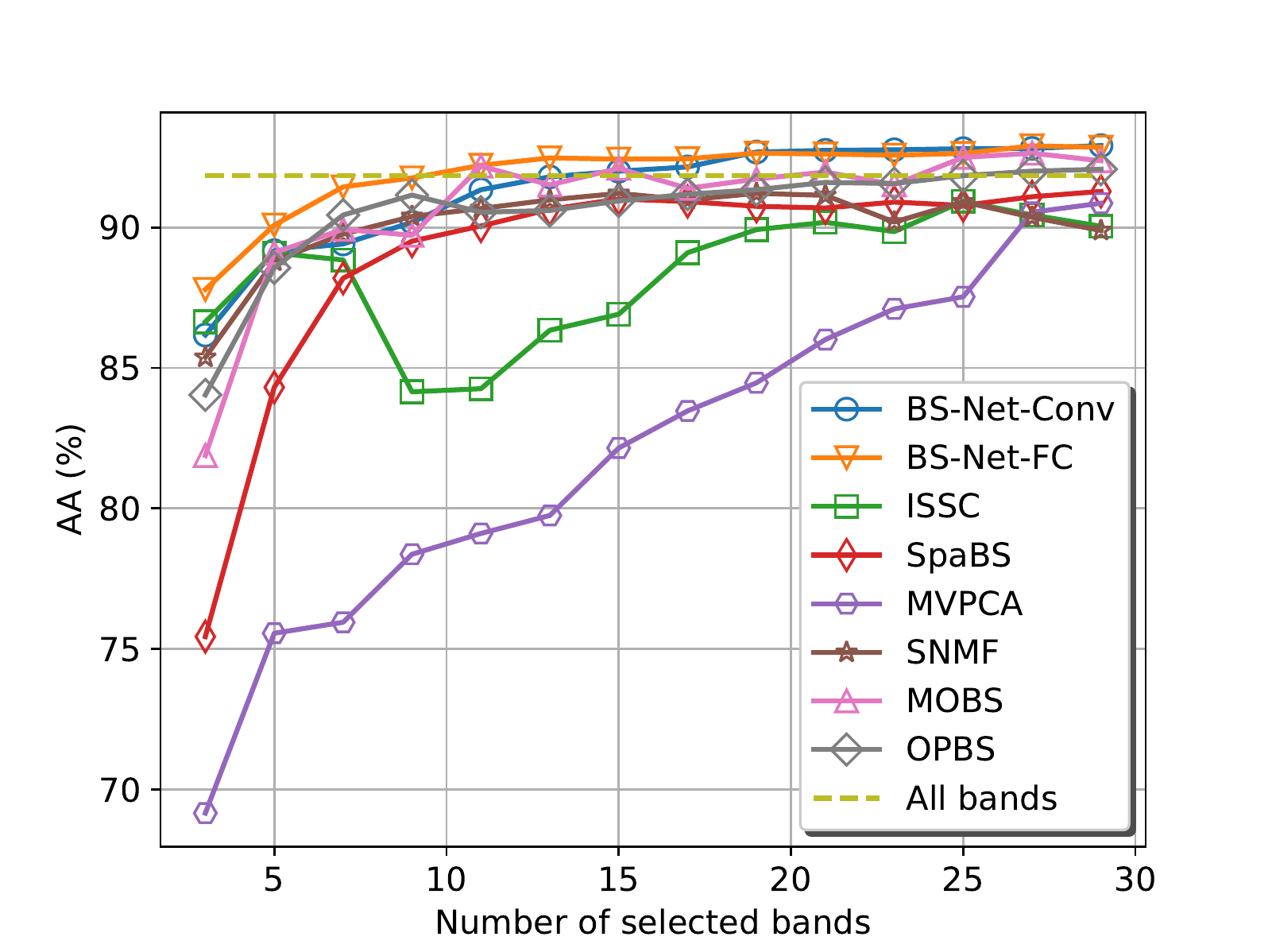}

}\subfloat[Kappa]{\includegraphics[width=0.65\columnwidth]{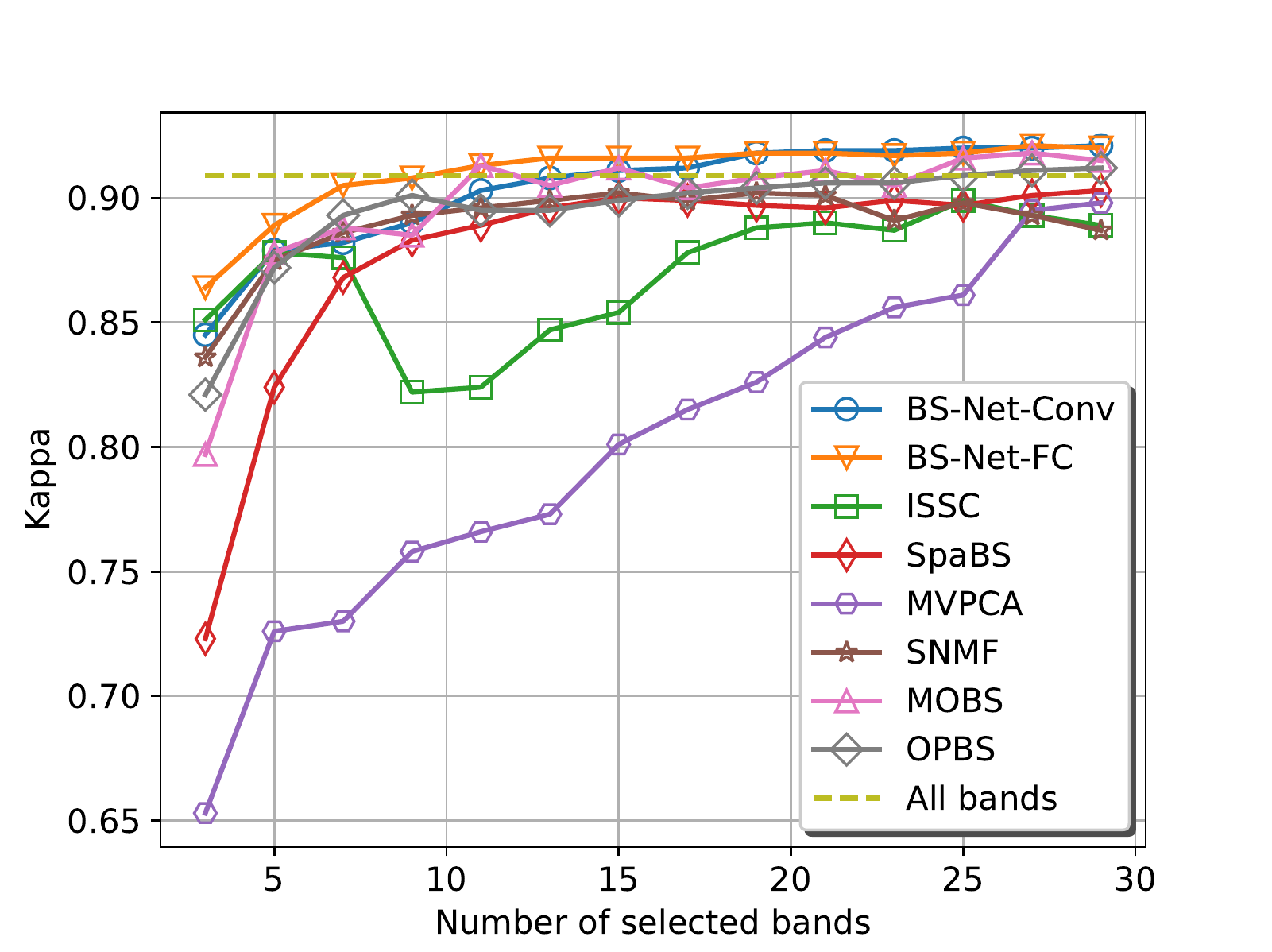}

}
\par\end{centering}
\caption{Performance comparison of different BS methods with different band
subset sizes on Salinas data set. (a) OA; (b) AA; (c) Kappa. \label{fig:Performance-comparison-of-Salinas}}
\end{figure*}

\begin{table*}[tbh]
\caption{Performance comparison of different methods using 15 bands on Salinas
data set. \label{tab:Performance-comparison-of-salinas} }
\centering{}{\scriptsize{}}%
\begin{tabular}{|c|c|c||c|c|c|c|c|c|c|c|}
\hline 
{\scriptsize{}NO.} & {\scriptsize{}\#Train} & {\scriptsize{}\#Test} & {\scriptsize{}ISSC} & {\scriptsize{}SpaBS} & {\scriptsize{}MVPCA} & {\scriptsize{}SNMF} & {\scriptsize{}MOBS} & {\scriptsize{}OPBS} & {\scriptsize{}BS-Net-FC} & {\scriptsize{}BS-Net-Conv}\tabularnewline
\hline 
\hline 
{\scriptsize{}1} & {\scriptsize{}100} & {\scriptsize{}1909} & {\scriptsize{}99.08$\pm$0.49} & {\scriptsize{}99.25$\pm$0.27} & {\scriptsize{}98.99$\pm$0.50} & {\scriptsize{}99.24$\pm$0.46} & {\scriptsize{}98.69$\pm$0.72} & {\scriptsize{}98.76$\pm$0.68} & {\scriptsize{}99.32$\pm$0.55} & \textbf{\scriptsize{}99.39$\pm$0.31}\tabularnewline
\hline 
{\scriptsize{}2} & {\scriptsize{}179} & {\scriptsize{}3547} & {\scriptsize{}99.72$\pm$0.34} & {\scriptsize{}99.55$\pm$0.23} & {\scriptsize{}98.47$\pm$0.57} & {\scriptsize{}99.64$\pm$0.22} & {\scriptsize{}99.53$\pm$0.25} & {\scriptsize{}99.69$\pm$0.26} & {\scriptsize{}99.58$\pm$0.37} & \textbf{\scriptsize{}99.72$\pm$0.25}\tabularnewline
\hline 
{\scriptsize{}3} & {\scriptsize{}97} & {\scriptsize{}1879} & {\scriptsize{}98.20$\pm$0.74} & {\scriptsize{}98.49$\pm$0.75} & {\scriptsize{}94.20$\pm$2.07} & {\scriptsize{}96.74$\pm$1.48} & {\scriptsize{}99.08$\pm$0.57} & {\scriptsize{}97.20$\pm$1.28} & \textbf{\scriptsize{}99.30$\pm$0.37} & {\scriptsize{}99.25}\textbf{\scriptsize{}$\pm$}{\scriptsize{}0.37}\tabularnewline
\hline 
{\scriptsize{}4} & {\scriptsize{}77} & {\scriptsize{}1317} & \textbf{\scriptsize{}99.21$\pm$0.62 } & {\scriptsize{}98.82$\pm$0.76} & {\scriptsize{}98.97$\pm$0.93} & {\scriptsize{}98.54$\pm$1.08} & {\scriptsize{}99.14$\pm$0.65} & {\scriptsize{}98.76$\pm$0.82} & {\scriptsize{}98.99$\pm$0.61} & {\scriptsize{}98.64}\textbf{\scriptsize{}$\pm$}{\scriptsize{}1.01}\tabularnewline
\hline 
{\scriptsize{}5} & {\scriptsize{}117} & {\scriptsize{}2561} & \textbf{\scriptsize{}98.42$\pm$0.58} & {\scriptsize{}98.19$\pm$0.49} & {\scriptsize{}94.78$\pm$0.74} & {\scriptsize{}96.29$\pm$1.19} & {\scriptsize{}97.98$\pm$0.62 } & {\scriptsize{}96.96$\pm$1.14} & {\scriptsize{}97.98$\pm$0.80} & {\scriptsize{}98.39}\textbf{\scriptsize{}$\pm$}{\scriptsize{}0.65}\tabularnewline
\hline 
{\scriptsize{}6} & {\scriptsize{}204} & {\scriptsize{}3755} & \textbf{\scriptsize{}99.88$\pm$0.06 } & {\scriptsize{}99.78$\pm$0.15} & {\scriptsize{}99.38$\pm$0.22} & {\scriptsize{}99.74$\pm$0.13} & {\scriptsize{}99.77$\pm$0.09 } & {\scriptsize{}99.79$\pm$0.09} & {\scriptsize{}99.79$\pm$0.11} & {\scriptsize{}99.79}\textbf{\scriptsize{}$\pm$}{\scriptsize{}0.12}\tabularnewline
\hline 
{\scriptsize{}7} & {\scriptsize{}180} & {\scriptsize{}3399} & \textbf{\scriptsize{}99.62$\pm$0.23} & {\scriptsize{}99.43$\pm$0.28} & {\scriptsize{}98.90$\pm$0.61} & {\scriptsize{}99.45$\pm$0.29} & {\scriptsize{}99.59$\pm$0.18} & {\scriptsize{}99.56$\pm$0.22} & {\scriptsize{}99.58$\pm$0.23} & {\scriptsize{}99.56}\textbf{\scriptsize{}$\pm$}{\scriptsize{}0.15}\tabularnewline
\hline 
{\scriptsize{}8} & {\scriptsize{}585} & {\scriptsize{}10686} & {\scriptsize{}82.18$\pm$1.76} & {\scriptsize{}85.51$\pm$1.52} & {\scriptsize{}83.51$\pm$1.32} & {\scriptsize{}85.08$\pm$1.50} & {\scriptsize{}86.24$\pm$1.23} & {\scriptsize{}85.06$\pm$1.71} & {\scriptsize{}87.27$\pm$1.60} & \textbf{\scriptsize{}88.15$\pm$1.13}\tabularnewline
\hline 
{\scriptsize{}9} & {\scriptsize{}305} & {\scriptsize{}5898} & {\scriptsize{}99.30$\pm$0.38} & {\scriptsize{}99.2$\pm$0.38} & {\scriptsize{}94.14$\pm$1.16} & {\scriptsize{}99.24$\pm$0.39} & {\scriptsize{}98.94$\pm$0.62} & {\scriptsize{}99.06$\pm$0.73} & {\scriptsize{}99.34$\pm$0.38} & \textbf{\scriptsize{}99.46$\pm$0.42}\tabularnewline
\hline 
{\scriptsize{}10} & {\scriptsize{}179} & {\scriptsize{}3099} & {\scriptsize{}93.57$\pm$1.18} & {\scriptsize{}90.29$\pm$1.01} & {\scriptsize{}85.51$\pm$2.03} & {\scriptsize{}90.97$\pm$1.02} & {\scriptsize{}93.73$\pm$1.25} & {\scriptsize{}91.75$\pm$1.75} & \textbf{\scriptsize{}94.95$\pm$1.12} & {\scriptsize{}94.82}\textbf{\scriptsize{}$\pm$}{\scriptsize{}1.01}\tabularnewline
\hline 
{\scriptsize{}11} & {\scriptsize{}47} & {\scriptsize{}1021} & {\scriptsize{}92.69$\pm$2.09} & {\scriptsize{}93.73$\pm$2.96} & {\scriptsize{}65.59$\pm$3.93} & {\scriptsize{}88.35$\pm$4.15} & {\scriptsize{}94.21$\pm$1.77} & {\scriptsize{}95.33$\pm$2.91} & {\scriptsize{}94.09$\pm$3.34} & \textbf{\scriptsize{}96.19$\pm$1.51}\tabularnewline
\hline 
{\scriptsize{}12} & {\scriptsize{}117} & {\scriptsize{}1810} & {\scriptsize{}98.89$\pm$0.92} & {\scriptsize{}99.05$\pm$0.61} & {\scriptsize{}90.51$\pm$2.81} & {\scriptsize{}97.74$\pm$1.03} & {\scriptsize{}99.58$\pm$0.33} & {\scriptsize{}99.47$\pm$0.54} & {\scriptsize{}99.49$\pm$0.88} & \textbf{\scriptsize{}99.59$\pm$0.71}\tabularnewline
\hline 
{\scriptsize{}13} & {\scriptsize{}41} & {\scriptsize{}875} & {\scriptsize{}98.56$\pm$0.80} & {\scriptsize{}96.62$\pm$2.35} & {\scriptsize{}97.78$\pm$0.93} & {\scriptsize{}96.28$\pm$1.96} & {\scriptsize{}98.09$\pm$1.34 } & {\scriptsize{}97.57$\pm$1.85} & {\scriptsize{}98.72$\pm$1.25} & \textbf{\scriptsize{}98.94$\pm$0.85}\tabularnewline
\hline 
{\scriptsize{}14} & {\scriptsize{}64} & {\scriptsize{}1006} & {\scriptsize{}94.24$\pm$1.77} & {\scriptsize{}95.51$\pm$1.28} & {\scriptsize{}95.49$\pm$1.83} & {\scriptsize{}92.78$\pm$2.13} & {\scriptsize{}95.58$\pm$1.08 } & {\scriptsize{}94.84$\pm$1.99 } & {\scriptsize{}96.18$\pm$1.55} & \textbf{\scriptsize{}96.76$\pm$1.53}\tabularnewline
\hline 
{\scriptsize{}15} & {\scriptsize{}337} & {\scriptsize{}6931} & {\scriptsize{}60.42$\pm$3.07} & {\scriptsize{}64.81$\pm$2.46} & {\scriptsize{}56.20$\pm$2.6 0} & {\scriptsize{}62.53$\pm$1.99} & {\scriptsize{}69.83$\pm$1.92 } & {\scriptsize{}70.24$\pm$2.60} & \textbf{\scriptsize{}71.60$\pm$2.34} & {\scriptsize{}70.41}\textbf{\scriptsize{}$\pm$}{\scriptsize{}2.23}\tabularnewline
\hline 
{\scriptsize{}16} & {\scriptsize{}77} & {\scriptsize{}1730} & {\scriptsize{} 97.60$\pm$0.69} & {\scriptsize{} 97.55$\pm$0.81} & {\scriptsize{}96.19$\pm$1.36} & {\scriptsize{}97.59$\pm$0.81} & {\scriptsize{}97.76$\pm$0.56} & \textbf{\scriptsize{}98.88$\pm$0.21} & {\scriptsize{}98.08$\pm$0.82} & {\scriptsize{}98.65}\textbf{\scriptsize{}$\pm$}{\scriptsize{}0.58}\tabularnewline
\hline 
\hline 
\multicolumn{3}{|c||}{{\scriptsize{}OA (\%)}} & {\scriptsize{}94.47$\pm$0.21} & {\scriptsize{}94.74$\pm$0.30} & {\scriptsize{}90.54$\pm$0.28} & {\scriptsize{}93.76$\pm$0.37} & {\scriptsize{}95.48$\pm$0.18} & {\scriptsize{}95.18$\pm$0.34} & {\scriptsize{}95.89$\pm$0.31} & \textbf{\scriptsize{}96.11$\pm$0.17}\tabularnewline
\hline 
\multicolumn{3}{|c||}{{\scriptsize{}AA (\%)}} & {\scriptsize{}89.85$\pm$0.18} & {\scriptsize{}90.90$\pm$0.22} & {\scriptsize{}87.10$\pm$0.24} & {\scriptsize{}90.19$\pm$0.22} & {\scriptsize{}91.97$\pm$0.2 } & {\scriptsize{}91.60$\pm$0.20} & {\scriptsize{}92.61$\pm$0.19} & \textbf{\scriptsize{}92.74$\pm$0.20}\tabularnewline
\hline 
\multicolumn{3}{|c||}{{\scriptsize{}Kappa}} & {\scriptsize{}0.887$\pm$0.002} & {\scriptsize{}0.899$\pm$0.002} & {\scriptsize{}0.856$\pm$0.003} & {\scriptsize{}0.891$\pm$0.003} & {\scriptsize{}0.911$\pm$0.002} & {\scriptsize{}0.906$\pm$0.002} & {\scriptsize{}0.918$\pm$0.002} & \textbf{\scriptsize{}0.919$\pm$0.002}\tabularnewline
\hline 
\end{tabular}{\scriptsize \par}
\end{table*}

The performance comparison of different BS methods on Salinas data
set is shown in Fig. \ref{fig:Performance-comparison-of-Salinas}
(a)-(c). From the results, BS-Net-FC achieves the best classification
performance when the band subset size is less than $19$. When band
subset size is larger than $19$, both BS-Nets are very comparable
in terms of OA, AA, and Kappa, and are significantly better than ISSC,
SpaBS, MVPCA, SNMF, and OPBS, as well as all bands. Table \ref{tab:Performance-comparison-of-salinas}
gives the detailed classification results of using $19$ best bands.
It can be seen that both BS-Nets are generally superior to the other
BS methods on most of the classes, and significantly outperform all
the other BS methods in terms of OA, AA, and Kappa.

\subsubsection{Analysis of the Selected Bands}

\begin{table}[tbh]
\caption{The best 15 bands of Salinas data set selected by different BS methods.\label{tab:The-best-15-Salinas}}
\centering{}{\scriptsize{}}%
\begin{tabular}{|c|c|}
\hline 
{\scriptsize{}Methods} & {\scriptsize{}Selected Bands}\tabularnewline
\hline 
\hline 
{\scriptsize{}BS-Net-FC} & {\scriptsize{}{[}53, 77, 61, 54, 16, 8, 158, 49, 176, 179, 56, 189,
197, 21, 43{]} }\tabularnewline
\hline 
{\scriptsize{}BS-Net-Conv} & {\scriptsize{}{[}116,153,19,189,97,179,171,141,95,144,142,46,104,203,91{]}}\tabularnewline
\hline 
{\scriptsize{}ISSC} & {\scriptsize{}{[}141,182,106,147,107,146,108,202,203,109,145,148,112,201,110{]} }\tabularnewline
\hline 
{\scriptsize{}SpaBS} & {\scriptsize{}{[}0, 79, 166, 80, 203, 78, 77, 76, 55, 81, 97, 5, 23,
75, 2{]} }\tabularnewline
\hline 
{\scriptsize{}MVPCA} & {\scriptsize{}{[}169,67,168,63,68,78,167,166,165,69,164,163,77,162,70{]}}\tabularnewline
\hline 
{\scriptsize{}SNMF} & {\scriptsize{}{[}24, 1, 105, 196, 203, 0, 39, 116, 38, 60, 89, 104,
198, 147{]} }\tabularnewline
\hline 
{\scriptsize{}MOBS} & {\scriptsize{}{[}20,29,35,54,60,62,75,81,93,119,129,132,141,163,201{]} }\tabularnewline
\hline 
{\scriptsize{}OPBS} & {\scriptsize{}{[}44, 31, 37, 66, 11, 1, 164, 2, 18, 0, 3, 40, 4, 54,
33{]}}\tabularnewline
\hline 
\end{tabular}{\scriptsize \par}
\end{table}

\begin{figure*}[tbh]
\begin{centering}
\includegraphics[width=2\columnwidth]{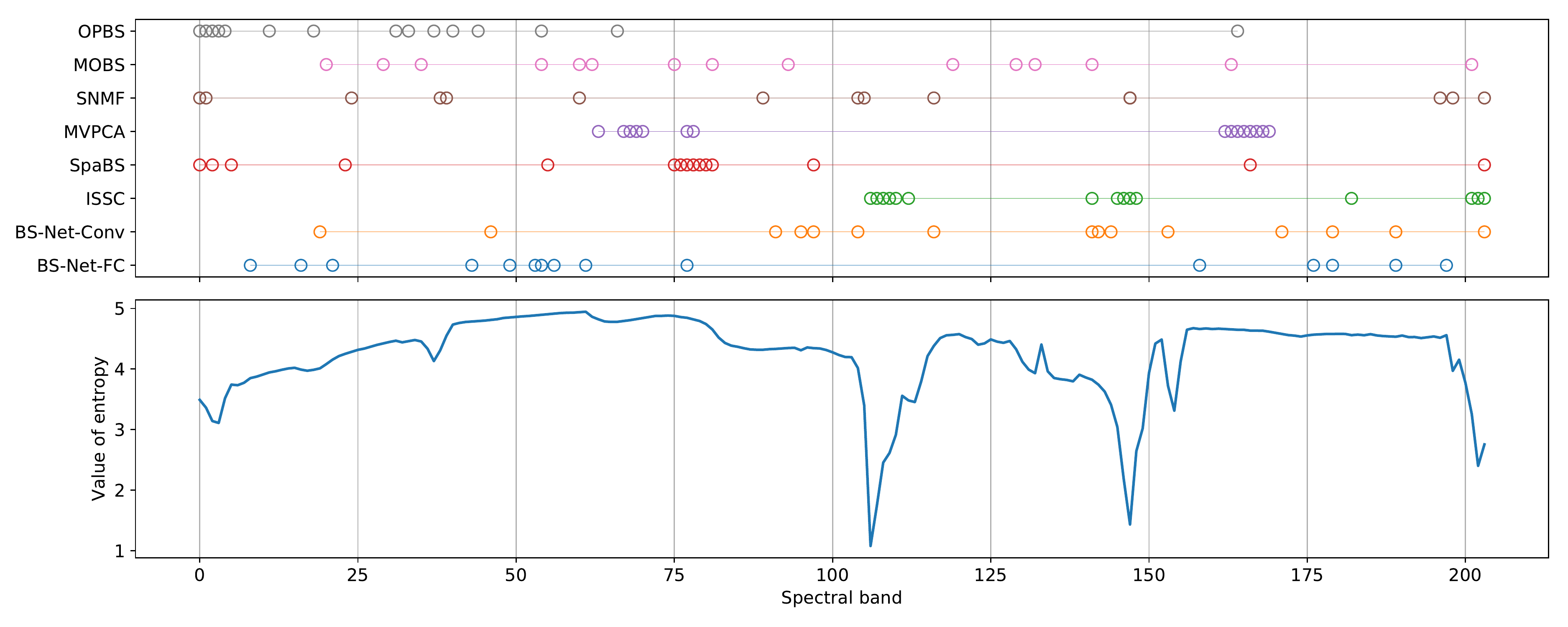}
\par\end{centering}
\caption{The best 15 bands of Salinas data set selected by different BS methods
(above) and the entropy value of each band (below). \label{fig:Entropy-of-indx-Salinas}}
\end{figure*}

\begin{figure}[tbh]
\begin{centering}
\includegraphics[width=0.8\columnwidth]{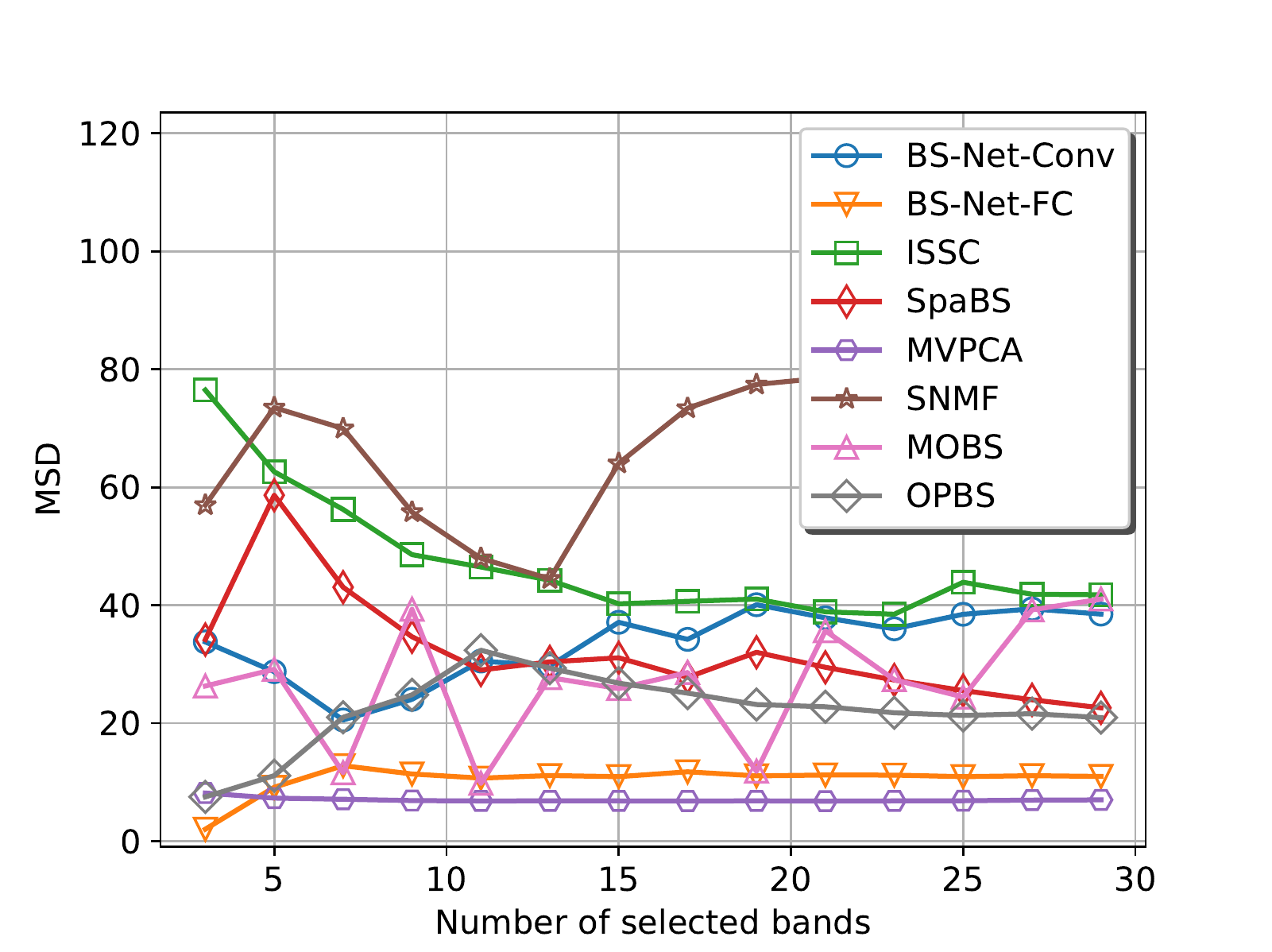}
\par\end{centering}
\caption{Mean Spectral Divergence values of different BS methods on Salinas
data set.\label{fig:MSD-values-of-Salinas}}
\end{figure}

The best $15$ bands selected by different BS methods are shown in
Table \ref{tab:The-best-15-Salinas}. Their distribution and entropy
are shown in Fig. \ref{fig:Entropy-of-indx-Salinas}. As we can see,
BS-Nets contain less adjacent bands and distribute relatively uniformly.
Observing the entropy curve, both BS-Nets can avoid the sharply decreasing
regions with low entropy, i.e., $\left[106,107\right]$ and $\left[146,147\right]$.
From Fig. \ref{fig:Entropy-of-indx-Salinas}, some BS methods include
a few continuous bands, i.e.,  OPBS, MVPCA, SpaBS, and ISSC, which
means higher correlation is included in their selected band subsets.
Fig. \ref{fig:MSD-values-of-Salinas} shows the MSD values of different
BS methods, showing that BS-Net-Conv has comparable MSD with ISSC
when the band subset size is larger than $15$. Although SNMF and
ISSC achieve the better MSDs, they can not obtain the best classification
performance since few of their selected bands locate at noisy regions.
Similar to the analysis for Indian Pines data set, these noise bands
will increase MSD but reduce the classification performance. In contrast,
BS-Net-FC has relative lower MSD, but it completely avoids noisy bands
and achieves better classification performance than other BS methods. 

\subsection{Computational Time Complexity Analysis}

To analyze the running time, we conduct all the BS methods on the
same computer and collect their absolute running time. MOBS and OPBS
are implemented in Matlab and the other methods are implemented in
Python. Instead of executing on CPU platform, we train BS-Nets on
a GPU platform due to its friendly GPU support. Fig. \ref{fig:Training-time-of-indianP}
illustrates the training time of BS-Net-FC and BS-Net-Conv trained
on Indian Pines data set with different iterations. According to the
implementation details shown in Table \ref{tab:BS-Net-architecture-details-FC}-\ref{tab:BS-Net-architecture-details-Conv},
BS-Net-FC and BS-Net-Conv include about $152,592$ and $590,288$
trainable parameters, respectively. However, the number of training
samples used in BS-Net-FC is $21025$ while that in BS-Net-Conv is
$4489$. It can be seen from Fig. \ref{fig:Training-time-of-indianP},
BS-Net-Conv saves about half of the time cost by comparing with BS-Net-FC.
It is interesting to notice that the training time is approximatively
linear to the iterations. 

Table \ref{tab:Computational-time-all-BS-method} shows the computational
time of selecting $19$ bands using different BS methods on the three
data sets. It can be seen that the running times of BS-Nets are comparable
with MOBS which bases on heuristic searching and significantly faster
than SpaBS and SNMF. Since ISSC and MVPCA can be solved using the
algebraic method, they show faster computation speed but cannot achieve
better classification performance than BS-Nets. In summary, the proposed
BS-Nets are able to balance classification performance and running
time.
\begin{center}
\begin{figure}[tbh]
\begin{centering}
\includegraphics[width=0.8\columnwidth]{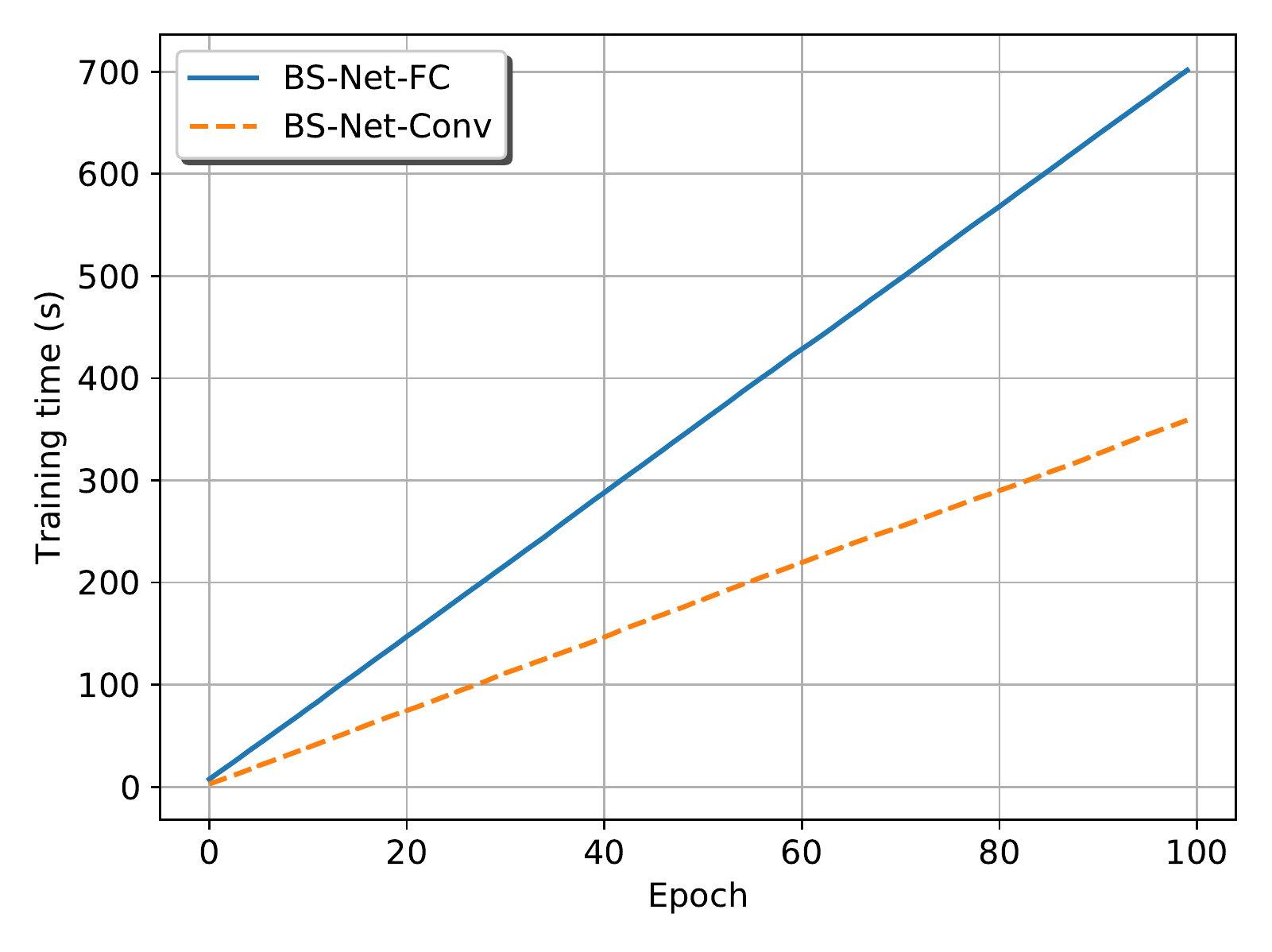}
\par\end{centering}
\caption{Training time of BS-Nets on Indian Pines data set. \label{fig:Training-time-of-indianP}}
\end{figure}
\par\end{center}

\begin{center}
\begin{table}[tbh]
\caption{Computational time (in seconds) for selecting $20$ bands using different
BS methods. \label{tab:Computational-time-all-BS-method}}
\centering{}%
\begin{tabular}{|c|c|c|c|}
\hline 
Method & Indian Pines & Pavia University & Salinas\tabularnewline
\hline 
\hline 
ISSC & 0.43 & 14.71 & 17.44\tabularnewline
\hline 
SpaBS & 332.54 & 2026.80 & 3224.57\tabularnewline
\hline 
MVPCA & 0.44 & 4.24 & 7.86\tabularnewline
\hline 
SNMF & $>1h$ & $>1h$ & $>1h$\tabularnewline
\hline 
MOBS & 275.76 & 289.18 & 330.30\tabularnewline
\hline 
OPBS & 2.00 & 9.62 & 9.65\tabularnewline
\hline 
BS-Net-FC & 652.08 & 2015.55 & 1116.99\tabularnewline
\hline 
 BS-Net-Conv & 238.04  & 493.15 & 444.37\tabularnewline
\hline 
\end{tabular}
\end{table}
\par\end{center}

\section{Conclusions \label{sec:Conclusions}}

This paper presents a novel end-to-end band selection network framework
for HSI band selection. The main idea behind the framework is to treat
HSI band selection as a sparse spectral reconstruction task and to
explicitly learn the spectral band's significance using deep neural
networks by considering the nonlinear correlation between spectral
bands. The resulting framework allows to learn band weights from full
spectral bands, resulting in more efficient use of the global spectral
relationship, and consists of two flexible sub-networks, band attention
module (BAM) and reconstruction network (RecNet), making it easy to
train and apply in practice. The experimental results show that the
implemented BS-Net-FC and BS-Net-Conv can not only adaptively produce
sparse band weights, but also can significantly better classification
performance than many existing BS methods with an acceptable time
cost. 

We notice that the proposed framework has the capacity of combining
with many deep learning based classification methods to reduce computational
complexity and enhance the classification performance. That will also
be further explored in our future works. 

\section*{Acknowledgment}

The authors would like to thank the anonymous reviewers for their
constructive suggestions and criticisms. We would also like to thank
Dr. W. Zhang who provided the source codes of the OPBS method, and
Prof. M. Gong who provided the source codes of the MOBS method. 

\bibliographystyle{IEEEtran}
\bibliography{Ref-A}

\begin{thebibliography}{10}
\providecommand{\url}[1]{#1}
\csname url@samestyle\endcsname
\providecommand{\newblock}{\relax}
\providecommand{\bibinfo}[2]{#2}
\providecommand{\BIBentrySTDinterwordspacing}{\spaceskip=0pt\relax}
\providecommand{\BIBentryALTinterwordstretchfactor}{4}
\providecommand{\BIBentryALTinterwordspacing}{\spaceskip=\fontdimen2\font plus
\BIBentryALTinterwordstretchfactor\fontdimen3\font minus
  \fontdimen4\font\relax}
\providecommand{\BIBforeignlanguage}[2]{{%
\expandafter\ifx\csname l@#1\endcsname\relax
\typeout{** WARNING: IEEEtran.bst: No hyphenation pattern has been}%
\typeout{** loaded for the language `#1'. Using the pattern for}%
\typeout{** the default language instead.}%
\else
\language=\csname l@#1\endcsname
\fi
#2}}
\providecommand{\BIBdecl}{\relax}
\BIBdecl

\bibitem{HSI-Agriculture-Applications-2015-JSTARS-2015}
C.~M. Gevaert, J.~Suomalainen, J.~Tang, and L.~Kooistra, ``Generation of
  spectral-temporal response surfaces by combining multispectral satellite and
  hyperspectral uav imagery for precision agriculture applications,''
  \emph{IEEE Journal of Selected Topics in Applied Earth Observations and
  Remote Sensing}, vol.~8, no.~6, pp. 3140--3146, June 2015.

\bibitem{HSI-spectral-unmixing-urban-environment-ReSensEnv-2017}
J.~Pontius, R.~P. Hanavan, R.~A. Hallett, B.~D. Cook, and L.~A. Corp, ``High
  spatial resolution spectral unmixing for mapping ash species across a complex
  urban environment,'' \emph{Remote Sensing of Environment}, vol. 199, pp. 360
  -- 369, 2017.

\bibitem{HSI-Medical-JBO-Lu-2014}
B.~F. Guolan~Lu, ``Medical hyperspectral imaging: a review,'' \emph{Journal of
  Biomedical Optics}, vol.~19, no.~1, pp. 1 -- 24 -- 24, 2014.

\bibitem{HSI-forensic-traces-FCI-2012}
G.~Edelman, E.~Gaston, T.~van Leeuwen, P.~Cullen, and M.~Aalders,
  ``Hyperspectral imaging for non-contact analysis of forensic traces,''
  \emph{Forensic Science International}, vol. 223, no.~1, pp. 28 -- 39, 2012.

\bibitem{HSI_BS-EvoMultiObj-GongMG-TGRS-2016}
M.~Gong, M.~Zhang, and Y.~Yuan, ``Unsupervised band selection based on
  evolutionary multiobjective optimization for hyperspectral images,''
  \emph{IEEE Transactions on Geoscience and Remote Sensing}, vol.~54, no.~1,
  pp. 544--557, Jan 2016.

\bibitem{Graph-Regu-Fast-RPCA-HSI_BS-SunW-TGRS-2018}
W.~Sun and Q.~Du, ``Graph-regularized fast and robust principal component
  analysis for hyperspectral band selection,'' \emph{IEEE Transactions on
  Geoscience and Remote Sensing}, vol.~56, no.~6, pp. 3185--3195, June 2018.

\bibitem{HSI_BS-Laplacian-Regularized-LRSC-ZhaiH-TGRS-2018}
H.~Zhai, H.~Zhang, L.~Zhang, and P.~Li, ``Laplacian-regularized low-rank
  subspace clustering for hyperspectral image band selection,'' \emph{IEEE
  Transactions on Geoscience and Remote Sensing}, pp. 1--18, 2018.

\bibitem{HSIC_SVM-MelganiF-TGRS-2004}
F.~Melgani and L.~Bruzzone, ``Classification of hyperspectral remote sensing
  images with support vector machines,'' \emph{IEEE Transactions on Geoscience
  and Remote Sensing}, vol.~42, no.~8, pp. 1778--1790, Aug 2004.

\bibitem{HSI_BS-Mutual-Information-Semi-Supervised-FengJ-TGRS-2015}
J.~Feng, L.~Jiao, F.~Liu, T.~Sun, and X.~Zhang, ``Mutual-information-based
  semi-supervised hyperspectral band selection with high discrimination, high
  information, and low redundancy,'' \emph{IEEE Transactions on Geoscience and
  Remote Sensing}, vol.~53, no.~5, pp. 2956--2969, May 2015.

\bibitem{HSI_BS-ISSC-SunWW-JSTARS-2015}
W.~Sun, L.~Zhang, B.~Du, W.~Li, and Y.~M. Lai, ``Band selection using improved
  sparse subspace clustering for hyperspectral imagery classification,''
  \emph{IEEE Journal of Selected Topics in Applied Earth Observations and
  Remote Sensing}, vol.~8, no.~6, pp. 2784--2797, June 2015.

\bibitem{HSI_BS-Dual-Clustering-YuanY-TGRS-2016}
Y.~Yuan, J.~Lin, and Q.~Wang, ``Dual-clustering-based hyperspectral band
  selection by contextual analysis,'' \emph{IEEE Transactions on Geoscience and
  Remote Sensing}, vol.~54, no.~3, pp. 1431--1445, March 2016.

\bibitem{HSI-Lap-Collaborative-Discriminant-Jiangxw-RS-2019}
X.~Jiang, X.~Song, Y.~Zhang, J.~Jiang, J.~Gao, and Z.~Cai, ``Laplacian
  regularized spatial-aware collaborative graph for discriminant analysis of
  hyperspectral imagery,'' \emph{Remote Sensing}, vol.~11, no.~1, p.~29, 2019.

\bibitem{HSI_BS-multi-feature-info-maxi-clustering-ICIP-2017}
M.~Bevilacqua and Y.~Berthoumieu, ``Unsupervised hyperspectral band selection
  via multi-feature information-maximization clustering,'' in \emph{2017 IEEE
  International Conference on Image Processing (ICIP)}, Sep. 2017, pp.
  540--544.

\bibitem{HSI_BS-Self-Representation-SunWW-JSTARS-2017}
W.~Sun, L.~Tian, Y.~Xu, D.~Zhang, and Q.~Du, ``Fast and robust
  self-representation method for hyperspectral band selection,'' \emph{IEEE
  Journal of Selected Topics in Applied Earth Observations and Remote Sensing},
  vol.~10, no.~11, pp. 5087--5098, Nov 2017.

\bibitem{HSI_BS-Multi-objectiveOpt-ZhangMY-AppSoftCom-2018}
M.~Zhang, M.~Gong, and Y.~Chan, ``Hyperspectral band selection based on
  multi-objective optimization with high information and low redundancy,''
  \emph{Applied Soft Computing}, vol.~70, pp. 604 -- 621, 2018.

\bibitem{HSI_BS-EMO-SparRepres-HuP-GRSL-2018}
P.~Hu, X.~Liu, Y.~Cai, and Z.~Cai, ``Band selection of hyperspectral images
  using multiobjective optimization-based sparse self-representation,''
  \emph{IEEE Geoscience and Remote Sensing Letters}, pp. 1--5, 2018.

\bibitem{HSI_BS-Optimal-Clustering-Framework-WangQ-TGRS-2018}
Q.~Wang, F.~Zhang, and X.~Li, ``Optimal clustering framework for hyperspectral
  band selection,'' \emph{IEEE Transactions on Geoscience and Remote Sensing},
  vol.~56, no.~10, pp. 5910--5922, Oct 2018.

\bibitem{HSI_BS-SNMF-JiM-FITEE}
L.~I. Ji-Ming and Y.~T. Qian, ``Clustering-based hyperspectral band selection
  using sparse nonnegative matrix factorization,'' \emph{Frontiers of
  Information Technology and Electronic Engineering}, vol.~12, no.~7, pp.
  542--549, 2011.

\bibitem{HSI_BS-MVPCA-ChangCI-TGRS-1999}
C.-I. Chang, Q.~Du, T.-L. Sun, and M.~L.~G. Althouse, ``A joint band
  prioritization and band-decorrelation approach to band selection for
  hyperspectral image classification,'' \emph{IEEE Transactions on Geoscience
  and Remote Sensing}, vol.~37, no.~6, pp. 2631--2641, Nov 1999.

\bibitem{HSI_BS-SpaBS-TargetDectection-SunK-GRSL-2015}
K.~Sun, X.~Geng, and L.~Ji, ``A new sparsity-based band selection method for
  target detection of hyperspectral image,'' \emph{IEEE Geoscience and Remote
  Sensing Letters}, vol.~12, no.~2, pp. 329--333, Feb 2015.

\bibitem{HSI_BS-Multitask-Sparsity-Pursuit-YuanY-TGRS-2015}
Y.~Yuan, G.~Zhu, and Q.~Wang, ``Hyperspectral band selection by multitask
  sparsity pursuit,'' \emph{IEEE Transactions on Geoscience and Remote
  Sensing}, vol.~53, no.~2, pp. 631--644, Feb 2015.

\bibitem{HSI-Band_Selection-OPBS-Geometry-Based-ZhangW-TGRS-2018}
W.~{Zhang}, X.~{Li}, Y.~{Dou}, and L.~{Zhao}, ``A geometry-based band selection
  approach for hyperspectral image analysis,'' \emph{IEEE Transactions on
  Geoscience and Remote Sensing}, vol.~56, no.~8, pp. 4318--4333, Aug 2018.

\bibitem{HSIC-Gaussian-Process-JiangXW-GRSL-2017}
X.~{Jiang}, X.~{Fang}, Z.~{Chen}, J.~{Gao}, J.~{Jiang}, and Z.~{Cai},
  ``Supervised gaussian process latent variable model for hyperspectral image
  classification,'' \emph{IEEE Geoscience and Remote Sensing Letters}, vol.~14,
  no.~10, pp. 1760--1764, Oct 2017.

\bibitem{Deep-learning-LeCun-Nature-2015}
Y.~LeCun, Y.~Bengio, and G.~Hinton, ``Deep learning,'' \emph{Nature}, vol. 521,
  no. 7553, pp. 436--444, 2015.

\bibitem{CNN-overview-GuJX-PR-2018}
J.~Gu, Z.~Wang, J.~Kuen, L.~Ma, A.~Shahroudy, B.~Shuai, T.~Liu, X.~Wang,
  G.~Wang, J.~Cai, and T.~Chen, ``Recent advances in convolutional neural
  networks,'' \emph{Pattern Recognition}, vol.~77, pp. 354--377, 2018.

\bibitem{CNN-HSI-FeaExtr-Clas-Chenyushi-TGRS-2016}
Y.~Chen, H.~Jiang, C.~Li, X.~Jia, and P.~Ghamisi, ``Deep feature extraction and
  classification of hyperspectral images based on convolutional neural
  networks,'' \emph{IEEE Transactions on Geoscience and Remote Sensing},
  vol.~54, no.~10, pp. 6232--6251, 2016.

\bibitem{Bi-Conv-LSTM-HSI-Liu-RS-2017}
Q.~Liu, F.~Zhou, R.~Hang, and X.~Yuan, ``Bidirectional-convolutional lstm based
  spectral-spatial feature learning for hyperspectral image classification,''
  \emph{Remote Sensing}, vol.~9, no.~12, p. 1330, 2017.

\bibitem{ResidualNN-HSI-Zhong-TGRS-2018}
Z.~Zhong, J.~Li, Z.~Luo, and M.~Chapman, ``Spectral-spatial residual network
  for hyperspectral image classification: A 3-d deep learning framework,''
  \emph{IEEE Transactions on Geoscience and Remote Sensing}, vol.~56, no.~2,
  pp. 847--858, 2018.

\bibitem{BLDE-CNN-HSI-Zhao-TGRS-2016}
W.~Zhao and S.~Du, ``Spectral-spatial feature extraction for hyperspectral
  image classification: A dimension reduction and deep learning approach,''
  \emph{IEEE Transactions on Geoscience and Remote Sensing}, vol.~54, no.~8,
  pp. 4544--4554, 2016.

\bibitem{Deep-FullyCNN-HSI-Jiao-TGRS-2017}
L.~Jiao, M.~Liang, H.~Chen, S.~Yang, H.~Liu, and X.~Cao, ``Deep fully
  convolutional network-based spatial distribution prediction for hyperspectral
  image classification,'' \emph{IEEE Transactions on Geoscience and Remote
  Sensing}, vol.~55, no.~10, pp. 5585--5599, 2017.

\bibitem{DL-HSI-overview-ZhangLP-GRSM-2016}
L.~Zhang, L.~Zhang, and B.~Du, ``Deep learning for remote sensing data: A
  technical tutorial on the state of the art,'' \emph{IEEE Geoscience and
  Remote Sensing Magazine}, vol.~4, no.~2, pp. 22--40, 2016.

\bibitem{HE-ELM-CYM-PRL-2018}
Y.~Cai, X.~Liu, Y.~Zhang, and Z.~Cai, ``Hierarchical ensemble of extreme
  learning machine,'' \emph{Pattern Recognition Letters}, 2018.

\bibitem{Mixed-dense-CNN-PeltMD-PNAS-2018}
D.~M. Pelt and J.~A. Sethian, ``A mixed-scale dense convolutional neural
  network for image analysis,'' \emph{Proc Natl Acad Sci U S A}, vol. 115,
  no.~2, pp. 254--259, 2018.

\bibitem{GAN-Goodfellow-NIPS-2014}
I.~Goodfellow, J.~Pouget-Abadie, M.~Mirza, B.~Xu, D.~Warde-Farley, S.~Ozair,
  A.~Courville, and Y.~Bengio, ``Generative adversarial nets,'' in
  \emph{Advances in Neural Information Processing Systems 27}, Z.~Ghahramani,
  M.~Welling, C.~Cortes, N.~D. Lawrence, and K.~Q. Weinberger, Eds.\hskip 1em
  plus 0.5em minus 0.4em\relax Curran Associates, Inc., 2014, pp. 2672--2680.

\bibitem{GAN-Overview-Creswell-IEEEIPM-2018}
A.~{Creswell}, T.~{White}, V.~{Dumoulin}, K.~{Arulkumaran}, B.~{Sengupta}, and
  A.~A. {Bharath}, ``Generative adversarial networks: An overview,'' \emph{IEEE
  Signal Processing Magazine}, vol.~35, no.~1, pp. 53--65, Jan 2018.

\bibitem{HSIC-RNN-Mou-TGRS-2017}
L.~{Mou}, P.~{Ghamisi}, and X.~X. {Zhu}, ``Deep recurrent neural networks for
  hyperspectral image classification,'' \emph{IEEE Transactions on Geoscience
  and Remote Sensing}, vol.~55, no.~7, pp. 3639--3655, July 2017.

\bibitem{Squeeze-and-Excitation-Hu-CVPR-2018}
J.~Hu, L.~Shen, and G.~Sun, ``Squeeze-and-excitation networks,'' in \emph{The
  IEEE Conference on Computer Vision and Pattern Recognition (CVPR)}, June
  2018.

\bibitem{Spatial-Transformer-Networks-attention-NIPS-2015}
M.~Jaderberg, K.~Simonyan, A.~Zisserman, and k.~kavukcuoglu, ``Spatial
  transformer networks,'' in \emph{Advances in Neural Information Processing
  Systems 28}, C.~Cortes, N.~D. Lawrence, D.~D. Lee, M.~Sugiyama, and
  R.~Garnett, Eds.\hskip 1em plus 0.5em minus 0.4em\relax Curran Associates,
  Inc., 2015, pp. 2017--2025.

\bibitem{Residual-Attention-Wang-CVPR-2017}
F.~Wang, M.~Jiang, C.~Qian, S.~Yang, C.~Li, H.~Zhang, X.~Wang, and X.~Tang,
  ``Residual attention network for image classification,'' in \emph{The IEEE
  Conference on Computer Vision and Pattern Recognition (CVPR)}, July 2017.

\bibitem{ICML-Xu_2015-img-caption-attention-Xu-2015}
K.~Xu, J.~Ba, R.~Kiros, K.~Cho, A.~Courville, R.~Salakhudinov, R.~Zemel, and
  Y.~Bengio, ``Show, attend and tell: Neural image caption generation with
  visual attention,'' in \emph{International conference on machine learning},
  2015, pp. 2048--2057.

\bibitem{Attention-Machine-Translation-Luong-EMNLP-2015}
T.~Luong, H.~Pham, and C.~D. Manning, ``Effective approaches to attention-based
  neural machine translation,'' in \emph{Proceedings of the 2015 Conference on
  Empirical Methods in Natural Language Processing}, 2015, pp. 1412--1421.

\bibitem{Attention-is-All-you-Need-NIPS-2017}
A.~Vaswani, N.~Shazeer, N.~Parmar, J.~Uszkoreit, L.~Jones, A.~N. Gomez, L.~u.
  Kaiser, and I.~Polosukhin, ``Attention is all you need,'' in \emph{Advances
  in Neural Information Processing Systems 30}, I.~Guyon, U.~V. Luxburg,
  S.~Bengio, H.~Wallach, R.~Fergus, S.~Vishwanathan, and R.~Garnett, Eds.\hskip
  1em plus 0.5em minus 0.4em\relax Curran Associates, Inc., 2017, pp.
  5998--6008.

\bibitem{SBAM-channel-spatial-attention-ECCV-2018}
S.~Woo, J.~Park, J.-Y. Lee, and I.~S. Kweon, ``Cbam: Convolutional block
  attention module,'' in \emph{Computer Vision -- ECCV 2018}, V.~Ferrari,
  M.~Hebert, C.~Sminchisescu, and Y.~Weiss, Eds.\hskip 1em plus 0.5em minus
  0.4em\relax Cham: Springer International Publishing, 2018, pp. 3--19.

\bibitem{Deep-Learning-information-bottleneck-Tishby-ITM-2015}
N.~Tishby and N.~Zaslavsky, ``Deep learning and the information bottleneck
  principle,'' in \emph{2015 IEEE Information Theory Workshop (ITW)}, April
  2015, pp. 1--5.

\bibitem{HSI-Band_Selection-Volume-Gradient-Based-GengX-TGRS-2014}
X.~{Geng}, K.~{Sun}, L.~{Ji}, and Y.~{Zhao}, ``A fast volume-gradient-based
  band selection method for hyperspectral image,'' \emph{IEEE Transactions on
  Geoscience and Remote Sensing}, vol.~52, no.~11, pp. 7111--7119, Nov 2014.

\bibitem{Hughes-Phenomenon-Explanation-TPAMI-1979}
G.~V. Trunk, ``A problem of dimensionality: A simple example,'' \emph{IEEE
  Transactions on Pattern Analysis and Machine Intelligence}, vol. PAMI-1,
  no.~3, pp. 306--307, July 1979.

\end{thebibliography}

\end{document}